\documentclass{article}
\PassOptionsToPackage{numbers,sort&compress}{natbib}

    \usepackage[final]{neurips_2021}

\usepackage[utf8]{inputenc} 
\usepackage[T1]{fontenc}    
\usepackage{hyperref}       
\usepackage{url}            
\usepackage{booktabs}       
\usepackage{amsfonts}       
\usepackage{nicefrac}       
\usepackage{microtype}      
\usepackage{xcolor}         
\usepackage{tabularx}
\usepackage{tabulary}
\usepackage{multirow}
\newcolumntype{Y}{>{\centering\arraybackslash}X}

\usepackage{physics}
\usepackage{amsmath}
\usepackage{tikz}
\usepackage{mathdots}
\usepackage{yhmath}
\usepackage{cancel}
\usepackage{color}
\usepackage{siunitx}
\usepackage{array}
\usepackage{multirow}
\usepackage{amssymb}
\usepackage{tabularx}
\usepackage{extarrows}
\usepackage{booktabs}
\usetikzlibrary{fadings}
\usetikzlibrary{patterns}
\usetikzlibrary{shadows.blur}
\usetikzlibrary{shapes}

\usepackage[capitalise,noabbrev]{cleveref}
\usepackage{bm}

\usepackage{array}
\newcolumntype{C}[1]{>{\centering\arraybackslash}p{#1}}

\usepackage{lscape}
\usepackage{rotating}
\usepackage{appendix}

\usepackage{minitoc}

\DeclareMathOperator*{\mindisttolineseg}{\mathbf{D}^2_{seg}}

\DeclareMathOperator{\sketch}{\mathbf{S}}
\DeclareMathOperator{\ReLU}{ReLU}
\DeclareMathOperator{\gameloss}{l_{game}}
\DeclareMathOperator{\perceploss}{l_{perceptual}}

\title{Learning to Draw: Emergent Communication through Sketching}

\author{%
  Daniela Mihai\\
  Electronics and Computer Science\\
  The University of Southampton\\
  Southampton, UK \\
  \texttt{adm1g15@soton.ac.uk} \\
   \And
   Jonathon Hare\\
  Electronics and Computer Science\\
  The University of Southampton\\
  Southampton, UK \\
  \texttt{jsh2@soton.ac.uk}\\
}

\begin{document}

\doparttoc 
\faketableofcontents 

\maketitle

\begin{abstract}
  Evidence that visual communication preceded written language and provided a basis for it goes back to prehistory, in forms such as cave and rock paintings depicting traces of our distant ancestors. Emergent communication research has sought to explore how agents can learn to communicate in order to collaboratively solve tasks. Existing research has focused on language, with a learned communication channel transmitting sequences of discrete tokens between the agents. In this work, we explore a visual communication channel between agents that are allowed to draw with simple strokes. Our agents are parameterised by deep neural networks, and the drawing procedure is differentiable, allowing for end-to-end training. In the framework of a referential communication game, we demonstrate that agents can not only successfully learn to communicate by drawing, but with appropriate inductive biases, can do so in a fashion that humans can interpret. We hope to encourage future research to consider visual communication as a more flexible and directly interpretable alternative of training collaborative agents.
\end{abstract}

\section{Introduction}
Imagine you and a friend are playing a game where you have to get your friend to guess an object in the room by you sketching the object. No other communication is allowed beyond the sketched image. This is an example of a \textit{referential communication game}. To play this game you need to have learned how to draw in a way that your friend can understand. This paper explores how artificial agents parameterised by neural networks can learn to play similar drawing games. More specifically, we reformulate the traditional referential game such that one agent draws a sketch of a given photo and the second agent guesses, based on the drawing, the corresponding photo from a set of images.

Spurred by innovations in artificial neural networks, deep and reinforcement learning techniques, recent work in multi-agent emergent communication \citep{chaabouni2020compositionality, guo2019emergence,ren2020compositional, lazaridou2018emergence, Havrylov2017} pursues interactions in the form of gameplay between agents to induce human-like communication. Artificial communicating agents can collaborate to solve various tasks: image referential games with realistic visual input \citep{Havrylov2017, lazaridou2017multi, lazaridou2018emergence}, negotiation \citep{cao2018emergent}, navigation of virtual environments \citep{das2019tarmac, jaques2019social}, reconstruction of missing input \citep{chaabouni2020compositionality, guo2019emergence} and, more recently, drawing games~\citep{fernando2020language}. The key to achieving the shared goal in many of these games is collaboration, and implicitly, communication. To date, studies on communication emergence in multi-agent games have focused on exploring a language-based communication channel, with messages represented by discrete tokens or token sequences \citep{Havrylov2017, DBLP:journals/corr/MordatchA17, das2017learning, lazaridou2018emergence, lazaridou2017multi, kharitonov2020entropy, guo2019emergence}. However, these communication protocols can be difficult for a human to interpret \citep{KotturMLB17,lowe2019pitfalls,chaab2019antieff}. In this work we propose a direct and potentially self-explainable means of transmitting knowledge: \textit{sketching}.

Evidence suggests pre- and early-humans were able to communicate by drawing long before developing the various stages of written language \citep{henshilwood2009reading, robinson2002lost}. Drawings such as petrograms and petroglyphs exist from the oldest palaeolithic times and may have been used to record past experiences, events, beliefs or simply the relation with other beings \citep{hoffmann2018u, fox1937prehistoric}. These pictorial characters which are merely impressions of real objects or beings stand at the basis of all writing \citep{gelb1963study}. This leads us to question if drawing is a more natural way of \textit{starting} to study emergent communication and if it could lead to better written communication later on.

This idea has recently gained interest in several domains. In the cognitive science literature, neural models of sketching have been developed to study the factors which enable contextual flexibility in visual communication \citep{fan2020pragmatic}. Likewise, works such as \citet{fernando2020language} attempt to automate the artistic process of drawing by training agents in a reinforcement learning framework, to play a variety of drawing games. However, the focus of this paper is to open the doorway to exploring different types of communication between artificial agents and humans. The novelty of our work is also evident in the model framework which can be easily extended well beyond aspects previously studied.

Concretely, we propose a visual communication channel in the context of image-based referential games. We leverage recent advances in differentiable sketching that enables us to construct an agent that \textit{can learn to communicate intent through drawing}. Through a range of experiments we show that:
\begin{itemize}
    \item Agents can successfully communicate about real-world images through a sketching game. However, training with a loss that tries to maximise gameplay alone does not lead to human decipherable sketches, irrespective of any visual system preconditioning;
    \item Introducing a perceptual loss improves human interpretability of the communication protocol, at little to no cost in the gameplay success;
    \item Changes to the game objective, such as playing an object-oriented game, can steer the emergent communication protocol towards a more pictographic or symbolic form of expression;
    \item Inducing a shape-bias into the agents' visual system leads to more explainable drawings;
    \item A drawing agent trained with a perceptual loss can successfully communicate and play the game with a human.
\end{itemize}

\section{Communication between agents}
Communication emerges when two or more participants are involved, share a goal, task or incentive which can be achieved only by transfer of information and so is beneficial for all parties involved. Studies on language origins  \citep{nowak1999evolution, steels1997synthetic} consider cooperation to be a key prerequisite to language evolution as it implies multiple agents having to self-organise and adapt to the same convention. Studies on the emergence of communication in cooperative multi-agent environments from recent years have focused on (natural) language learning~\citep{lazaridou2018emergence, lazaridou2017multi} and its inherent properties such as compositionality and expressivity~\citep{guo2019emergence, ren2020compositional, guo2020inductive}.

A number of works specifically relate to the overarching ideas of gameplay and learning in this paper. For example, \citet{FoersterAFW16a} proposed a framework for differentiable training of communicating agents which was later used by \citet{JorgeKG16} to solve image search tasks with two interacting agents communicating with atomic symbols. \citet{lazaridou2017multi} proposed an image-based referential game in which the agents again communicated using atomic symbols, and were trained using policy gradients. \citet{Havrylov2017} and \citet{DBLP:journals/corr/MordatchA17} both demonstrated that it was possible to use differentiable relaxations to train agents that communicated with sequences of symbols. In the former case, the agents played the referential game that we adopt for our experiments. 

One of the long-term goals of this research in language emergence is to develop interactive machines that can productively communicate with humans. As such we should ensure that whatever language artificial agents develop, it is one that human agents can understand. In our work, we take inspiration from the process and evolution of writing. Written language has undergone many transitions from early times to reach the forms we now know: from pictures and drawings to word-syllabic, syllabic and, finally, alphabetic systems. In the beginning, our early ancestors did not know how to communicate in writing. Instead, they began drawing and painting pictures of their life, representing people and things they knew about \citep{gelb1963study}. Studies on the communication systems developed in primitive societies compare ancient drawings to the very early sketches drawn by children and talk about their tendency of concretely identifying certain things or events in their surrounding world \citep{gelb1963study, kellogg1969analyzing}. Psychological and behavioural studies have shown that children try to communicate to the world through the images they create even when they cannot associate them with words \citep{farokhi2011analysis}.



\section{A model for learning to communicate by drawing}\label{sec:setup}
\begin{figure}[ht!]
    \centering
    \resizebox{0.945\textwidth}{!}{\input{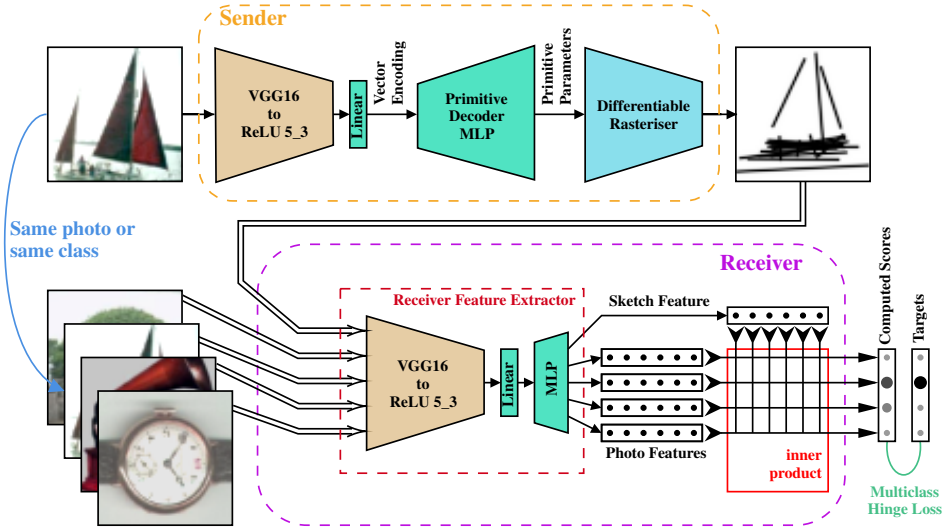}}
    \caption{\textbf{Overview of the agent architecture and game setup.} The `sender' agent is presented with an image and sketches its content through a learnable drawing procedure. The `receiver' agent is presented with the sketch and a collection of photographs, and has to learn to correctly associate the sketch with the corresponding photograph by predicting scores which are compared to a one-hot target. Both agents are parameterised by neural networks trained end-to-end using gradient methods.}
    \label{fig:overview}
\end{figure}

We present a model consisting of two agents, the sender and the receiver in which the sender learns to draw by playing a game with the receiver. The overall architecture of the agents in the context of the game they are learning to play is shown in \cref{fig:overview}. Full code for the model and all experiments can be found at \url{https://github.com/Ddaniela13/LearningToDraw}.

\subsection{The Game Environment}\label{subsec:games}
Our experimental setup builds upon the image referential game previously explored in studies of emergent communication \citep{Havrylov2017, lazaridou2017multi,lazaridou2018emergence} that derives from Lewis's signalling game \citep{Lewis1969-LEWCAP-4}. We implemented several variants of \citet{Havrylov2017}'s image guessing game. The overall setting of these games is formulated as follows:
\begin{enumerate}
    \item Two target photographs, $\mathbf{P}_s$ and $\mathbf{P}_r$, and set of $K$ distractor photographs, $\{\mathbf{P}_d^{(k)}\}_{k=1}^K$, are selected. 
    \item There are two agents: a sender and a receiver.
    \item After being presented the $\mathbf{P}_s$ target image, the sender has to formulate a message conveying information about that image.
    \item Given the message and the set of photographs, $\{\mathbf{P}_d^{(k)}\}_{k=1}^K \cup \{\mathbf{P}_r\}$, consisting of all the distractors and the target $\mathbf{P}_r$, the receiver has to identify the target correctly.
\end{enumerate}
The specifics of how the photographs are selected (step 1 above) depend on the game variant as described below. Success in these games is measured by the binary ability of the receiver to correctly guess the correct image or not; as such, the measure of \textit{communication rate} is used to assess averaged performance over many games using independent images to those used during training. Unlike \citet{Havrylov2017}'s game in which the sender helps the receiver identify the correct image by sending a message constructed as a sequence of tokens drawn from a predefined vocabulary, we propose using a directly interpretable means of communication: \textit{sketching the target photograph}. 

\paragraph{Original game variant.} In \citet{Havrylov2017}'s variant of the game there is a pool of photos from which the distractors and target $\mathbf{P}_s$ are drawn randomly without replacement. The target $\mathbf{P}_r$ is set to be equal to $\mathbf{P}_s$. In our \textit{original} variant experiments the number of distractors, $K$, is set to $99$.

\paragraph{Object-oriented game variants.} In addition to the original setup, we explored two slightly different and potentially harder game configurations which were intended to induce the agents to draw sketches that would be more representative to the object class they belong to rather than to the specific instance of the class. These setups use labelled datasets where each image belongs to a class based on its contents. In the first of these variants (we refer to this as \textit{OO-game same}), the target $\mathbf{P}_r$ is set to be equal to $\mathbf{P}_s$, and the distractors and target are sampled such that their class labels are disjoint (that is every photo provided to the receiver has a different class). The second setup (\textit{OO-game different}) is similar to the first, but the target $\mathbf{P}_r$ is chosen to be a different photograph with the same class label as target $\mathbf{P}_s$. The intention behind these games is to explore a universally interpretable depiction of the different object classes, which does not focus on individual details but rather conveys the concept. To some extent, this task is an example of multiple instance classification within a weakly supervised setting \citep{amores2013multiple}, which has been previously explored in the emergent communication literature \citep{lazaridou2017multi}.

\subsection{Agents' Architectures}\label{subsec:archi}
Both agents act on visual inputs.
The agents are parameterised by deep neural networks and are trained using standard gradient techniques (\cref{sec:training}).

\paragraph{The agent's early visual system.}
We choose to model the early visual systems of both agents with the head part of the VGG16 CNN architecture~\citep{Simonyan15} through to the $\ReLU$ activation at the end of the last convolutional layer (commonly referred to as the \verb|ReLU5_3| layer) before the final max-pooling and fully connected layers. In all experiments, we utilise pretrained weights and freeze this part of the model during training. We justify this choice on the basis that it provides the agents with an initial grounding in understanding the statistics of the visual world, and ensures that the visual system cannot collapse and remains universal. The weights are the standard \texttt{torchvision} ImageNet weights, except in the cases where we explore the effect of shape bias (see \cref{sec:shapebias}). As these pretrained weights were learned with images that were normalised according to the ImageNet statistics, all inputs to the VGG16 backbone (including sketches) are normalised accordingly. The output feature maps of this convolutional backbone are flattened and are linearly projected to a fixed dimensional vector encoding (64-dimensions unless otherwise specified). Because the datasets used in gameplay have different resolutions, the number of weights in the learned projection varies.

\paragraph{Sender Agent.}\label{sec:sender}
The goal of the sender is to produce a sketch from the input photograph. For experiments in \cref{sec:experiments}, we restrict the production of sketches to be a drawing composed of 20 black, constant width, straight lines on a white canvas of the same size as the input images. Experiments with fewer lines can be found in \cref{app-complexity}. It is of course possible to have a much more flexible definition of a sketch and incorporate many different modelling assumptions. We choose to leave such exploration for future work and focus on the key question of whether we can actually achieve successful (and potentially interpretable) communication with our simplified but not unrealistic setup.

Given an input image, the agent's early visual system produces a vector encoding which is then processed by a three-layer multilayer perceptron (MLP) that learns to decode the primitive parameters used to draw the sketch. This MLP has $\ReLU$ activations on the first two layers and $\tanh$ activation on the final layer. Unless otherwise specified, the first two layers have 64 and 256 neurons respectively. The output layer produces four values for each line that will be drawn; the values are the start and end coordinates of each line stroke in an image canvas with the origin at the centre and edges at $\pm 1$.

To produce a sketch image from the line parameters output by the MLP, we utilise the differentiable rasterisation approach introduced by \citet{DBLP:journals/corr/abs-2103-16194}. At a high level, this approach works by computing the distance transform on a pixel grid for each primitive being rendered. A relaxed approximation of a rasterisation function is applied to the distance transform to compute a raster image of the specific primitive. Finally, a differentiable composition function is applied to compose the individual rasters into a single image. More specifically, the squared Euclidean Distance Transform is computed, $\mindisttolineseg(\bm{s}, \bm{e})$ over all pixels in the image, for each line segment starting at coordinate $\bm{s}$ and ending at $\bm{e}$. These squared distance transforms are simply images in which the value of each pixel is replaced with the closest squared distance to the line (computed when the pixels are mapped to the same coordinate system as the line --- so the top left of the image is $(-1,-1)$ and bottom-right is $(1,1)$). Using the subscript $i$ to refer to the $i$-th line in the sketch, each $\mindisttolineseg(\bm{s_i}, \bm{e_i})$ is rasterised as
\begin{equation}
    \mathbf{R}_i = \exp(-\frac{\mindisttolineseg(\bm{s_i}, \bm{e_i}))}{ \sigma^2}) \;,
\end{equation}
where $\sigma^2$ is a hyperparameter that controls how far gradients flow in the image, as well as the visible thickness of the line ($\sigma^2=\num{5e-4}$ for all experiments in this paper). We adopt the soft-or composition function~\citep{DBLP:journals/corr/abs-2103-16194} to compose the individual line rasters into a single image, but incorporate an inversion so that a sketch image, $\sketch$, with a white canvas and black lines is produced,
\begin{equation}
    \sketch = \prod_{i=1}^n (\bm{1} - \mathbf{R}_i) \;,
\end{equation}
where $n$ is the number of lines. Finally, because the backbone CNNs work with three-band colour images, we replicate the greyscale sketch image three times across the channel dimension.

\paragraph{Receiver Agent.}\label{sec:receiver}

The receiver agent is given a set of photographs and a sketch image, and is responsible for predicting which photograph matches the sketch under the rules of the specific game being played. The receiver's visual system is coupled with a two-layer MLP with a $\ReLU$ nonlinearity on the first layer (the latter layer has no activation function). Unless otherwise specified, all experiments use 64 neurons in the first layer and 64 in the final layer. The sketch image and each photograph are passed through the visual system and MLP independently to produce a feature vector representation of the respective input. A score vector $\bm{x}$ is produced for the photographs by computing the scalar product of the sketch feature with the feature of each respective photograph. This score vector is un-normalised but could be viewed as a probability distribution by passing it through a softmax. The photograph with the highest score is the one predicted.

\subsection{Training details}\label{sec:training}
By incorporating a loss between the predicted scores of the receiver agent and the known correct target photograph, it is possible to propagate gradients back through both the receiver and sender agents. As such, we can train the agents to play the different game settings. For the loss function, we follow \citet{Havrylov2017} and choose to use \citet{Weston1999}'s multi-class generalisation of hinge loss (aka multi margin loss),
\begin{equation}
    \gameloss(\bm{x}, y) = \sum_{j \neq y} \max(0, 1 - \bm{x}_y + \bm{x}_j)\;,
\end{equation}
where $\bm{x}$ is the score vector produced by the receiver, and $y$ is the true index of the target, and the subscripts indicate indexing into the vector.
The rationale for this choice is that the (soft) margin constraint should help force the distractor photographs' features to be more dissimilar to the sketch feature. Tests using cross-entropy also indicated that it could work well as an alternative, however. 

Optimisation of the parameters of both agents is performed using the Adam optimiser with an initial learning rate of $\num{1e-4}$ for all experiments. For efficiency, we train the model with batches of games where the sender is given multiple images which are converted to sketches and passed to the receiver which reuses the same set of photographs for each sketch in the batch (with each sketch targeting a different receiver photograph). The order of the targets with respect to the input image's sketches is shuffled every batch. Batch size is $K+1$, where $K$ is the number of distractors, for all experiments. Unless otherwise stated, training was performed for 250 epochs. A mixture of Nvidia GTX1080s, RTX2080s, Quadro RTX8000s, and an RTX-Titan was used for training the models. Higher resolution images required more memory. Training time varied from around 488 games/second (10 secs/epoch) for games using STL10 to around 175 games/second (around 5 mins/epoch) for Caltech-101 experiments with 128px images.

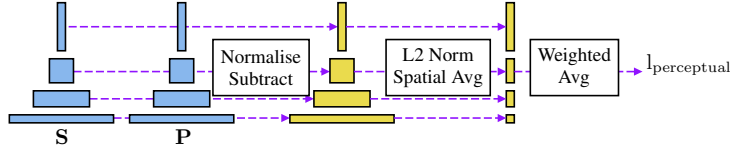
\begin{figure}[t]
    \centering
    \resizebox{0.7\textwidth}{!}{\tikzset{every picture/.style={line width=0.75pt}} 

\begin{tikzpicture}[x=0.75pt,y=0.75pt,yscale=-1,xscale=1]

\draw [color={rgb, 255:red, 144; green, 19; blue, 254 }  ,draw opacity=1 ] [dash pattern={on 3.75pt off 1.5pt}]  (65,72) -- (172,72) ;
\draw [shift={(175,72)}, rotate = 180] [fill={rgb, 255:red, 144; green, 19; blue, 254 }  ,fill opacity=1 ][line width=0.08]  [draw opacity=0] (5.36,-2.57) -- (0,0) -- (5.36,2.57) -- cycle    ;
\draw [color={rgb, 255:red, 144; green, 19; blue, 254 }  ,draw opacity=1 ] [dash pattern={on 3.75pt off 1.5pt}]  (50,60) -- (187,60) ;
\draw [shift={(190,60)}, rotate = 180] [fill={rgb, 255:red, 144; green, 19; blue, 254 }  ,fill opacity=1 ][line width=0.08]  [draw opacity=0] (5.36,-2.57) -- (0,0) -- (5.36,2.57) -- cycle    ;
\draw [color={rgb, 255:red, 144; green, 19; blue, 254 }  ,draw opacity=1 ] [dash pattern={on 3.75pt off 1.5pt}]  (40,43) -- (197,43) ;
\draw [shift={(200,43)}, rotate = 180] [fill={rgb, 255:red, 144; green, 19; blue, 254 }  ,fill opacity=1 ][line width=0.08]  [draw opacity=0] (5.36,-2.57) -- (0,0) -- (5.36,2.57) -- cycle    ;
\draw [color={rgb, 255:red, 144; green, 19; blue, 254 }  ,draw opacity=1 ] [dash pattern={on 3.75pt off 1.5pt}]  (30,15) -- (202,15) ;
\draw [shift={(205,15)}, rotate = 180] [fill={rgb, 255:red, 144; green, 19; blue, 254 }  ,fill opacity=1 ][line width=0.08]  [draw opacity=0] (5.36,-2.57) -- (0,0) -- (5.36,2.57) -- cycle    ;
\draw  [fill={rgb, 255:red, 136; green, 183; blue, 237 }  ,fill opacity=1 ] (30,0) -- (35,0) -- (35,30) -- (30,30) -- cycle ;
\draw  [fill={rgb, 255:red, 136; green, 183; blue, 237 }  ,fill opacity=1 ] (25,35) -- (40,35) -- (40,50) -- (25,50) -- cycle ;
\draw  [fill={rgb, 255:red, 136; green, 183; blue, 237 }  ,fill opacity=1 ] (15,55) -- (50,55) -- (50,65) -- (15,65) -- cycle ;
\draw  [fill={rgb, 255:red, 136; green, 183; blue, 237 }  ,fill opacity=1 ] (0,70) -- (65,70) -- (65,75) -- (0,75) -- cycle ;
\draw  [fill={rgb, 255:red, 136; green, 183; blue, 237 }  ,fill opacity=1 ] (105,0) -- (110,0) -- (110,30) -- (105,30) -- cycle ;
\draw  [fill={rgb, 255:red, 136; green, 183; blue, 237 }  ,fill opacity=1 ] (100,35) -- (115,35) -- (115,50) -- (100,50) -- cycle ;
\draw  [fill={rgb, 255:red, 136; green, 183; blue, 237 }  ,fill opacity=1 ] (90,55) -- (125,55) -- (125,65) -- (90,65) -- cycle ;
\draw  [fill={rgb, 255:red, 136; green, 183; blue, 237 }  ,fill opacity=1 ] (75,70) -- (140,70) -- (140,75) -- (75,75) -- cycle ;
\draw  [fill={rgb, 255:red, 234; green, 219; blue, 77 }  ,fill opacity=1 ] (205,0) -- (210,0) -- (210,30) -- (205,30) -- cycle ;
\draw  [fill={rgb, 255:red, 234; green, 219; blue, 77 }  ,fill opacity=1 ] (200,35) -- (215,35) -- (215,50) -- (200,50) -- cycle ;
\draw  [fill={rgb, 255:red, 234; green, 219; blue, 77 }  ,fill opacity=1 ] (190,55) -- (225,55) -- (225,65) -- (190,65) -- cycle ;
\draw  [fill={rgb, 255:red, 234; green, 219; blue, 77 }  ,fill opacity=1 ] (175,70) -- (240,70) -- (240,75) -- (175,75) -- cycle ;
\draw  [fill={rgb, 255:red, 234; green, 219; blue, 77 }  ,fill opacity=1 ] (310,0) -- (315,0) -- (315,30) -- (310,30) -- cycle ;
\draw  [fill={rgb, 255:red, 234; green, 219; blue, 77 }  ,fill opacity=1 ] (310,35) -- (315,35) -- (315,50) -- (310,50) -- cycle ;
\draw  [fill={rgb, 255:red, 234; green, 219; blue, 77 }  ,fill opacity=1 ] (310,55) -- (315,55) -- (315,65) -- (310,65) -- cycle ;
\draw  [fill={rgb, 255:red, 234; green, 219; blue, 77 }  ,fill opacity=1 ] (310,70) -- (315,70) -- (315,75) -- (310,75) -- cycle ;
\draw [color={rgb, 255:red, 144; green, 19; blue, 254 }  ,draw opacity=1 ] [dash pattern={on 3.75pt off 1.5pt}]  (240,72) -- (307,72) ;
\draw [shift={(310,72)}, rotate = 180] [fill={rgb, 255:red, 144; green, 19; blue, 254 }  ,fill opacity=1 ][line width=0.08]  [draw opacity=0] (5.36,-2.57) -- (0,0) -- (5.36,2.57) -- cycle    ;
\draw [color={rgb, 255:red, 144; green, 19; blue, 254 }  ,draw opacity=1 ] [dash pattern={on 3.75pt off 1.5pt}]  (225,60) -- (307,60) ;
\draw [shift={(310,60)}, rotate = 180] [fill={rgb, 255:red, 144; green, 19; blue, 254 }  ,fill opacity=1 ][line width=0.08]  [draw opacity=0] (5.36,-2.57) -- (0,0) -- (5.36,2.57) -- cycle    ;
\draw [color={rgb, 255:red, 144; green, 19; blue, 254 }  ,draw opacity=1 ] [dash pattern={on 3.75pt off 1.5pt}]  (215,43) -- (307,43) ;
\draw [shift={(310,43)}, rotate = 180] [fill={rgb, 255:red, 144; green, 19; blue, 254 }  ,fill opacity=1 ][line width=0.08]  [draw opacity=0] (5.36,-2.57) -- (0,0) -- (5.36,2.57) -- cycle    ;
\draw [color={rgb, 255:red, 144; green, 19; blue, 254 }  ,draw opacity=1 ] [dash pattern={on 3.75pt off 1.5pt}]  (210,15) -- (307,15) ;
\draw [shift={(310,15)}, rotate = 180] [fill={rgb, 255:red, 144; green, 19; blue, 254 }  ,fill opacity=1 ][line width=0.08]  [draw opacity=0] (5.36,-2.57) -- (0,0) -- (5.36,2.57) -- cycle    ;
\draw  [fill={rgb, 255:red, 255; green, 255; blue, 255 }  ,fill opacity=1 ] (127,23) -- (187,23) -- (187,58) -- (127,58) -- cycle ;
\draw  [fill={rgb, 255:red, 255; green, 255; blue, 255 }  ,fill opacity=1 ] (235,23) -- (300,23) -- (300,58) -- (235,58) -- cycle ;
\draw [color={rgb, 255:red, 144; green, 19; blue, 254 }  ,draw opacity=1 ] [dash pattern={on 3.75pt off 1.5pt}]  (315,43) -- (392,43) ;
\draw [shift={(395,43)}, rotate = 180] [fill={rgb, 255:red, 144; green, 19; blue, 254 }  ,fill opacity=1 ][line width=0.08]  [draw opacity=0] (5.36,-2.57) -- (0,0) -- (5.36,2.57) -- cycle    ;
\draw  [fill={rgb, 255:red, 255; green, 255; blue, 255 }  ,fill opacity=1 ] (325,23) -- (380,23) -- (380,58) -- (325,58) -- cycle ;

\draw (157,40.5) node  [font=\footnotesize] [align=left] {\begin{minipage}[lt]{40.8pt}\setlength\topsep{0pt}
\begin{center}
Normalise\\Subtract
\end{center}

\end{minipage}};
\draw (267.5,40.5) node  [font=\footnotesize] [align=left] {\begin{minipage}[lt]{44.2pt}\setlength\topsep{0pt}
\begin{center}
L2 Norm\\Spatial Avg
\end{center}

\end{minipage}};
\draw (352.5,40.5) node  [font=\footnotesize] [align=left] {\begin{minipage}[lt]{37.4pt}\setlength\topsep{0pt}
\begin{center}
Weighted\\Avg
\end{center}

\end{minipage}};
\draw (396,30.4) node [anchor=north west][inner sep=0.75pt]    {$\perceploss$};
\draw (27,77.4) node [anchor=north west][inner sep=0.75pt]    {$\mathbf{S}$};
\draw (102,77.4) node [anchor=north west][inner sep=0.75pt]    {$\mathbf{P}$};

\end{tikzpicture}}
    \caption{\textbf{Computing a `perceptual' loss with the early visual system.} Features are extracted from the sketch $\mathbf{S}$ and corresponding photograph $\mathbf{P}$ from different layers of the backbone. The features are normalised over channels and subtracted. We take the sum of the squared differences over channels and average spatially. Finally, we compute a weighted average across layers.}
    \label{fig:perceptualloss}
\end{figure}

\subsection{Making the sender agent's sketches more perceptually relevant}\label{subsec:peceploss}
Perception of drawings has a long history of study in neuroscience \citep[see e.g.][for an overview]{10.3389/fnhum.2011.00118}. In order to induce the sender to produce sketches that are more interpretable, we explore the idea of using an additional loss function between the differences in feature maps of the backbone CNN from the produced sketch and the input image. Such a loss has a direct grounding in biology, where it has been observed through human brain imaging studies that sketches and photographs of the same scene result in similar activations of neuron populations in area V4 of the visual cortex, as well as other areas related to higher-order visual cognition~\citep{Walther9661}. At the same time, it has also been demonstrated that differences in feature maps from pre-trained CNN architectures can be good proxies for approximating human notions of perceptual similarity between pairs of images~\citep{zhang2018perceptual}. 

Inspired by \citet{zhang2018perceptual} we formulate a loss based on the normalised differences between feature maps of the backbone network from the application of the network to both the input photograph and the corresponding sketch. Unlike \citeauthor{zhang2018perceptual} we choose not to learn weightings for each feature map channel individually, but rather we consider all feature maps produced by a layer of the backbone to be weighted equally. Learning individual channel weighting would be an interesting direction for future research, but is challenging because we would want to avoid the network learning zero weights for each channel, where the perceptual loss is basically ignored. 

\Cref{fig:perceptualloss} illustrates our perceptual loss formulation; note also that unlike \citet{zhang2018perceptual} the final averaging operation does incorporate a (per-layer) weighting, $\bm{w}_l$, which we explore the effect of in \cref{sec:weighting}.  More formally, denoting the sketch as $\mathbf{S}$ and corresponding photo as $\mathbf{P}$, we extract $L=5$ feature maps, ${\hat{\mathbf{S}}^{(l)}},{\hat{\mathbf{P}}^{(l)}} \in \mathbb{R}^{H_l \times W_l \times C_l}$, for the $l$-th layer from the backbone VGG16 network and unit normalise each across the channel dimension. The loss is thus defined as,
\begin{equation}
    \perceploss(\mathbf{S}, \mathbf{P}, \bm{w}) = \sum_l \frac{\bm{w}_l}{H_l W_l} \sum_{h,w} \big\| \hat{\mathbf{S}}_{hw}^{(l)} - \hat{\mathbf{P}}_{hw}^{(l)} \big\|^2_2 \; .
\end{equation}
To extract the feature maps we choose to use the outputs of the VGG16 layers immediately before the max-pooling layers (\verb|relu1_2|, \verb|relu2_2|, \verb|relu3_3|, \verb|relu4_3| and \verb|relu5_3|). During training, this perceptual loss is added to the game loss ($\gameloss$).
We note that the perceptual loss formulation is basically equivalent to the \emph{content loss} in neural style transfer~\cite{7780634}. Neural style transfer combines this content loss with a \emph{style loss} which encourages the texture statistics of a generated raster image to match a target image (which could be a sketch). Our model is different because instead of a loss \emph{encouraging} a sketch-like style we directly \emph{impose} production of sketches by drawing strokes.

\section{Experiments}\label{sec:experiments}
We next present a series of experiments where we explore if it is possible for the two agents to learn to successfully communicate, and what factors affect human interpretation of the drawings. We report numerical results averaged across 10 seeds for models evaluated on test sets isolated from training. Sample sketches from one seed are shown, but an overlay of 10 seeds can be found in \cref{app-sec:overlaid-sketches}.

\subsection{Can agents communicate by learning to draw?}\label{subsec:baselineexpts}
We explore the game setups described in \cref{subsec:games} and train our agents to play the games using $96\times96$ photographs from the STL-10 dataset~\citep{coates2011analysis}. For the \textit{original} game we use 99 distractors. For the object-oriented games, due to the dataset only having 10 classes, we are limited to 9 distractors.

In \cref{tab:baseline-expts}, we show quantitative and qualitative results of the visual communication game played under the three different configurations. The results demonstrate that it is possible for agents to successfully play this type of image referential game by learning to draw. One can observe that although agents achieve a high communication success rate, using only the $\gameloss$ loss leads to the emergence of a communication protocol that is indecipherable to a human. However, the addition of the perceptual loss, motivated in \cref{subsec:peceploss}, significantly improves the interpretability of the communication channel at almost no cost in the actual communication success rate.

One interesting observation is that although the sketches for some of the classes have greatly improved when incorporating the perceptual loss, for photographs of animals or birds, the sketches are not particularly representative of the class instance or distinguishable for the human eye. In the following sections we explore the model to try to better understand what factors affect drawing production.

\begin{table}[t]
\caption{\textbf{Communication success rate and example sketches produced by the agents in order to achieve the game objective in various setups and with different losses.} Sample input images seen by the sender (the left column) are described as the sketches in the second and third column. Although successful communication seems to be achieved in all setups, the addition of the perceptual loss significantly improves human interpretability of the drawings. Examples are from STL-10.}
    \label{tab:baseline-expts}
    \centering
    \begin{tabular}{ccc}
        \toprule
         & $\gameloss$ & $\gameloss + \perceploss$\\
        \midrule
        Original game & $71.8\%\ (\pm 6.1)$ & $69.57\%\ (\pm 2.6)$ \\
        \includegraphics[width=0.28\textwidth]{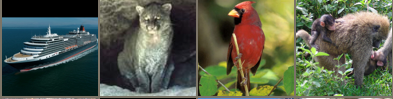} &
        \includegraphics[width=0.28\textwidth]{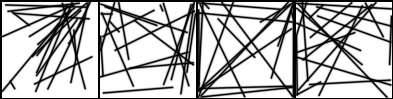} &
        \includegraphics[width=0.28\textwidth]{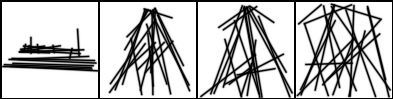} \\
        \midrule
        OO-game same & $95.46\%\ (\pm 0.6)$ & $96.04\%\ (\pm 0.5)$ \\ 
        \includegraphics[width=0.28\textwidth]{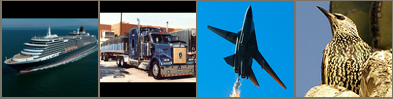} &
        \includegraphics[width=0.28\textwidth]{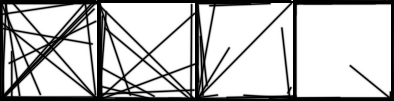}&
        \includegraphics[width=0.28\textwidth]{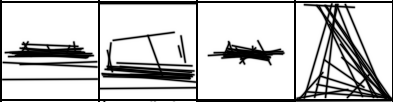}\\
        \midrule
        OO-game different & $82.72\%\ (\pm 0.8)$ & $81.09\%\ (\pm 0.6)$ \\
        \includegraphics[width=0.28\textwidth]{images/STL-10-games/oo1-samples.png} &
        \includegraphics[width=0.28\textwidth]{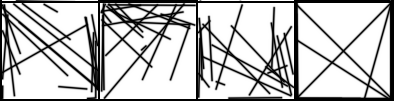}&
        \includegraphics[width=0.28\textwidth]{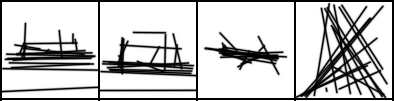}\\
        \bottomrule
    \end{tabular}
\end{table}

\subsection{What effect does weighting the perceptual loss have on the sketches?} \label{sec:weighting}
Next, we explore the effect of manually weighting the perceptual loss. More precisely, we look at what happens when the perceptual loss is applied to the features maps from just one layer of the backbone network. As previously mentioned in \cref{subsec:archi}, the feature maps are extracted using a VGG16 CNN up to \verb|ReLU5_3| layer. For example, we can discard all feature maps except those from the first layer by weighting the perceptual loss by $[1,0,0,0,0]$. The effect of the different weights, which allow only one block of feature maps to be used for drawing the sketch, is illustrated in \cref{tab:sop-weights}.  We apply these constraints in two setups, the \textit{original} and the \textit{OO-game different}. In both cases, the drawings are unrecognisable if the perceptual loss takes into account only the first or the second block of feature maps. Blocks 3 through 5 seem to provide increasing structure under both game setups. It is worth noticing that, similar to the results shown in \cref{subsec:baselineexpts}, the communication success rate in the original setup is always lower than that from the \textit{OO-game different} setup. Overall, the information provided by individual layers in the visual extractor network is enough for the agents to develop a visual communication strategy that can be used to play the game. For humans, however, the later layers contribute the most to the emergence of a communication protocol that we can understand.


\begin{table}[t]
\caption{\textbf{The effect of weighting the perceptual loss such that only the feature maps from one backbone layer are used.} The features extracted in the last three layers of the visual system seem to capture information that leads to sketches which resemble to an extent the corresponding photograph.}
    \label{tab:sop-weights}
    \centering
    \begin{tabular}{lccccc}
         \toprule
         Loss weights & $[1,0,0,0,0]$ & $[0,1,0,0,0]$ & $[0,0,1,0,0]$ & $[0,0,0,1,0]$ & $[0,0,0,0,1]$  \\
         \midrule
         Orig. game & \small$68.4\%\ (\pm 3.6)$ & \small$69.6\%\ (\pm 2.2)$ & \small$71.1\%\ (\pm 2.4)$ & \small$76.4\%\ (\pm 2.1)$ & \small$60.5\%\ (\pm 4.8)$ \\
         \includegraphics[width=0.13\textwidth]{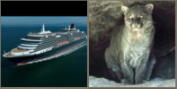} & \includegraphics[width=0.13\textwidth]{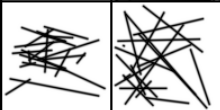} &
         \includegraphics[width=0.13\textwidth]{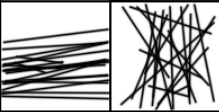} & \includegraphics[width=0.13\textwidth]{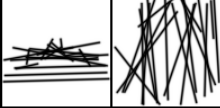} & 
         \includegraphics[width=0.13\textwidth]{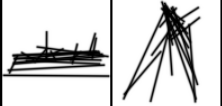} & 
         \includegraphics[width=0.13\textwidth]{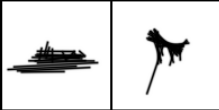} \\
         \midrule
         OO-game diff & \small$81.9\%\ (\pm 1.2)$ & \small$81.5\%\ (\pm 0.9)$ & \small$82.3\%\ (\pm 0.9)$ & \small$82.5\%\ (\pm 0.5)$ & \small$81.4\%\ (\pm 0.8)$ \\
         \includegraphics[width=0.13\textwidth]{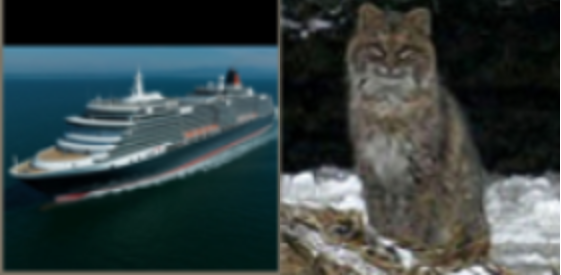} &
         \includegraphics[width=0.13\textwidth]{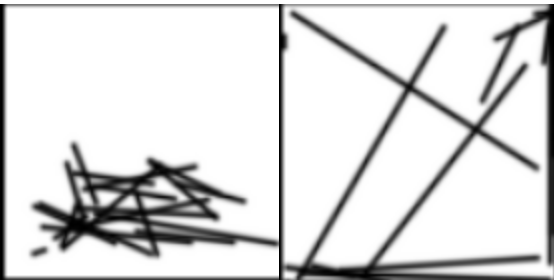}&
         \includegraphics[width=0.13\textwidth]{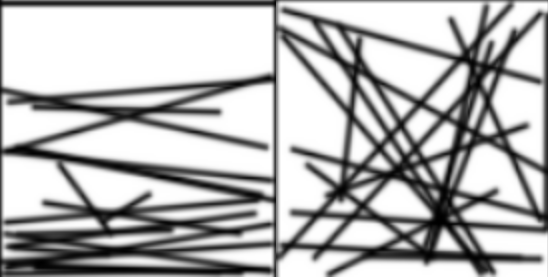}&
         \includegraphics[width=0.13\textwidth]{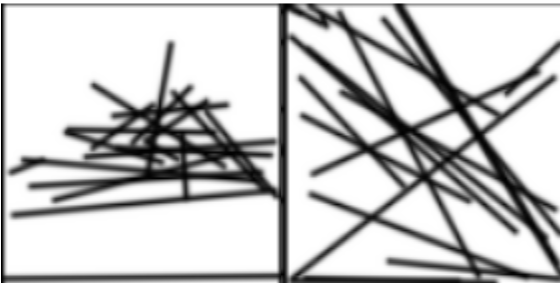}&
         \includegraphics[width=0.13\textwidth]{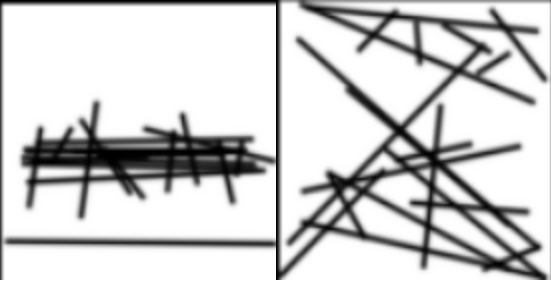}&
         \includegraphics[width=0.13\textwidth]{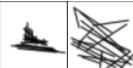}\\
         \bottomrule
    \end{tabular}
\end{table}

\subsection{Does the \textit{OO-game} influence the sketches to be more recognisable as the type of object?}\label{subsec:oo-games}

Comparing the qualitative results of different game formats from \cref{tab:baseline-expts}, we notice that agents develop distinct strategies for representing the target photograph under different conditions. If there is more variability in the sketches that correspond to photographs from the same class in the original game setup, and a bit less in the \textit{OO-game same}, the sketches become more like symbols representing all the photographs from one class when playing \textit{OO-game different}. In other words, the object-oriented games influence the sketches to be more recognisable as the type of object, than the specific instance of the class. Further examples are shown in \cref{app-games}.

Finally, it is worth noting how our results connect to how humans communicate through sketching when constrained under similar settings. The far/close contexts used in \citep{fan2020pragmatic} are somewhat equivalent to our original/object-oriented settings. As \citet{fan2020pragmatic} observe when humans play a similar drawing game, our agents achieve a higher recognition accuracy in settings that involve targets from different classes and develop different communication behaviours based on the context of the receiver. 

\subsection{How does the model's capacity influence the visual communication channel?}
Regarding the model's architecture, we look into how drawings are influenced by the width of the model. In this experiment (results shown in \cref{tab:modelcapacity}), we compare the baseline model architecture detailed in \cref{subsec:archi} with a wider variant that has the following changes: the sender encodes the target photograph to a 1024-dimensional vector (baseline model encodes to 64-dimensional vector), the receiver's MLP capacity is also increased from 64 to 1024 in both layers. We present results for the \textit{OO-game different} setup played with $128\times128$ Caltech-101 images \citep{fei2004learning}. The increased number of classes in Caltech-101 may explain the drop in the communication rate in this particular game setting, which compared to the same model played under the original game setup (see the ImageNet-pretrained model in \cref{tab:shape-texture-bias}), is with almost $ 30\% $ lower. As one might expect, the wider model allows for more details to be captured, and, hence, conveyed in the sketches. Unlike the baseline model which, in this object-oriented setup, develops a communication system that is more representative to the class than to the instance (as discussed in \cref{subsec:oo-games}), the wider model starts to draw distinctive representations for objects of the same type. More sketches can be found in \cref{app-modelcapacity} where one can observe the difference between all images with chairs, for example.

\begin{table}[bt]
\caption{\textbf{The effect of the model's capacity on its sketches.} The wide model's sender encodes the photo into a 1024-dimensional vector (baseline 64), and the receiver's MLP linear layers have 1024 neurons each versus 64. Examples from training on Caltech-101 in the \textit{OO-game different} setting.}
    \label{tab:modelcapacity}
    \centering
    \begin{tabular}{ccc}
        \toprule
         & Baseline & Wide \\
        \midrule
        & $50.46\%\ (\pm 1.5)$ & $64.99\%\ (\pm 1.5)$ \\
        \includegraphics[width=0.28\textwidth]{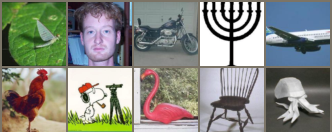} &
        \includegraphics[width=0.28\textwidth]{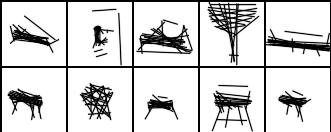} &
        \includegraphics[width=0.28\textwidth]{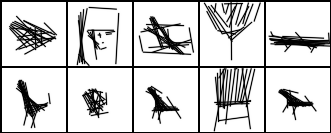} \\
        \bottomrule
    \end{tabular}
\end{table}

\subsection{How does the texture/shape bias of the visual system alter communication?} \label{sec:shapebias}
Next, we show that a texture or shape bias of the visual system influences visual communication. This experiment was run under the original game setup with $128\times128$ Caltech-101 images \citep{fei2004learning}. The results shown in \cref{tab:shape-texture-bias} suggest that inducing a ``shape bias" into the model does not significantly improve the agent's performance in playing the game, but produces more meaningful drawings. By using the VGG16 weights pretrained on Stylized-ImageNet \citep{geirhos2018imagenetTexture}, the communication protocol also becomes more faithful to the actual shape of the objects. A shape-based sketch is much more interpretable to humans, as it has been known for a long time that shape is the most important cue for human object recognition \citep{landau1988importance}. Further results from this experiment can be found in \cref{app-shape-texture}.

\begin{table}[bt]
    \caption{\textbf{The effect on the communication protocol of using a VGG16 feature extractor network pretrained on datasets that have texture (ImageNet) or shape (Stylized-ImageNet \citep{geirhos2018imagenetTexture}) bias.} Examples are from agents trained using the \textit{original} game with Caltech-101 data. The shape-biased sketches are better at capturing the overall object form, particularly for things like faces.}
    \label{tab:shape-texture-bias}
    \centering
    \begin{tabular}{ccc}
        \toprule
         & ImageNet weights & Stylized-ImageNet weights\\
        \midrule
        & $78.46\%\ (\pm 2.0)$ & $77.09\%\ (\pm 1.9)$ \\
        \includegraphics[width=0.28\textwidth]{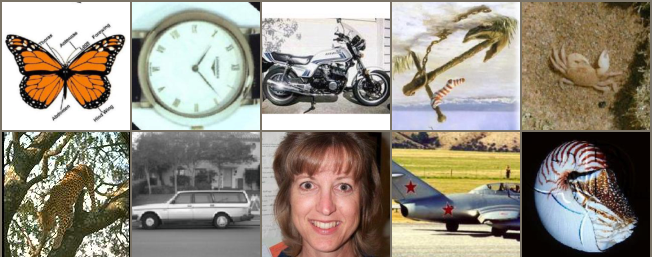} &
        \includegraphics[width=0.28\textwidth]{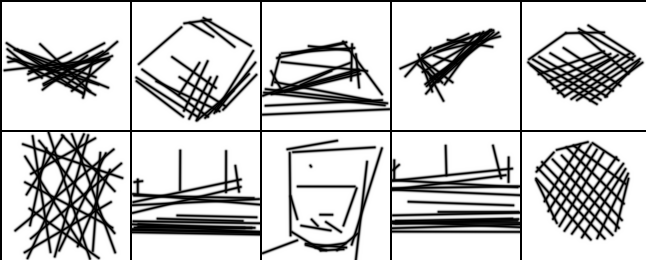} &
        \includegraphics[width=0.28\textwidth]{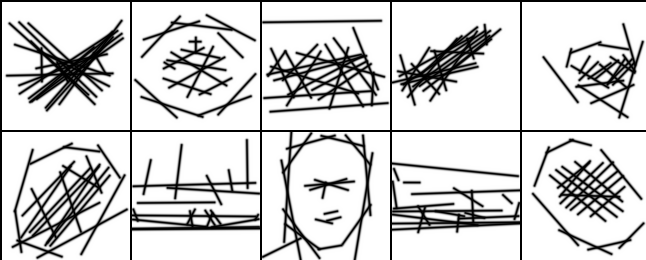} \\
        \bottomrule
    \end{tabular}
\end{table}

\subsection{Do the models learn to pick out salient features?}\label{subsec:salient-features}
\begin{figure}[tb]
    \centering
    \begingroup
    \setlength{\tabcolsep}{2pt} 
    \renewcommand{\arraystretch}{0} 
    \setlength\extrarowheight{0pt}
    \begin{tabular}{m{0.1\linewidth}m{0.83\linewidth}}
         &  \includegraphics[width=\linewidth]{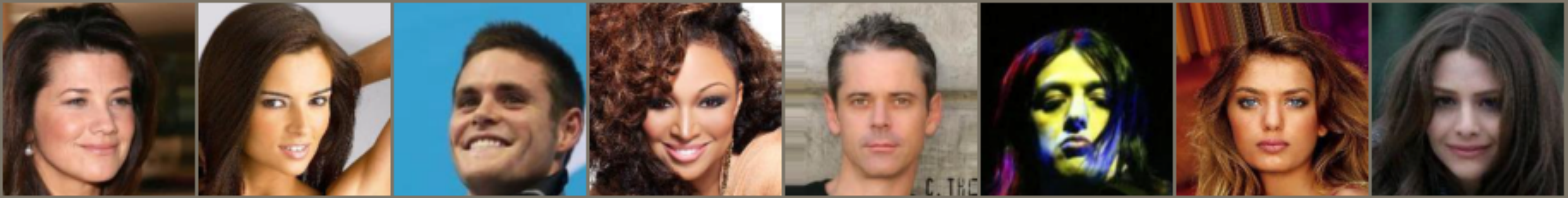}\\
         \begin{flushright} ImageNet weights (98.9\%) \end{flushright} &  \includegraphics[width=\linewidth]{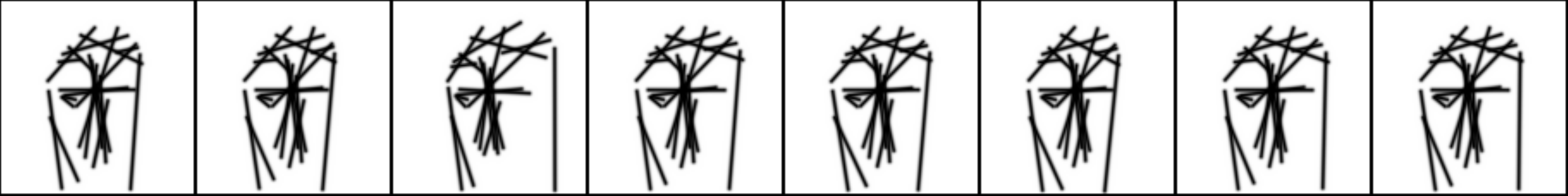}\\
         \begin{flushright} Stylized weights (99.6\%) \end{flushright} &  \includegraphics[width=\linewidth]{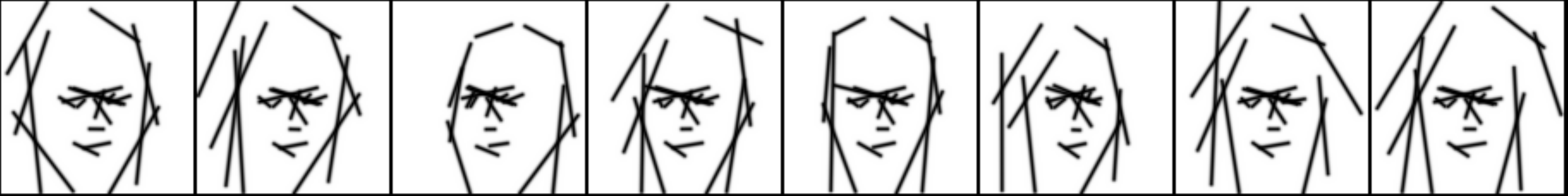}
    \end{tabular}
    \endgroup
    \caption{\textbf{Sketches from \textit{original} variant games using the CelebA dataset with perceptual loss and different biases from backbone weights.} Both the texture-biased (ImageNet) and shape-biased (Stylized-ImageNet) settings exhibit near-perfect communication success, but the shape-biased sketches are considerably more interpretable and show visual variations correlated with the photos.}
    \label{fig:celeba}
\end{figure}

From the results we have presented so far, it is evident that, particularly with the perceptual loss, the sender agent is able to broadly draw pictures of particular classes of object. The high communication rates in the \textit{original} game setting would also suggest that the drawings can capture something \textit{specific} about the target images that allow them to be identified amongst the distractors. To further analyse what is being captured by the models we train the agents in the original game setting (using both normal and stylized backbone weights) with images from the CelebA dataset~\citep{liu2015faceattributes}, which we take the maximal square centre-crop and resize to 112px. As this dataset contains only images of faces, messages between the agents will have to capture much more subtle information to distinguish the target from the distractors. Results are shown in \cref{fig:celeba}; the communication rate is near perfect for both models, but the difference between the texture-biased and shape-biased models is striking. There is subtle variation in the texture biased model's sketches which broadly seems to capture head pose, but the overall sketch structure is similar. In the shape-biased model head pose is evident, but so are other salient features like hairstyle and (see \cref{app-saliency}) head-wear and glasses.

\subsection{Do agents learn to draw in a fashion that humans can interpret?} \label{sec:humaneval}

In order to assess the interpretability of sketches drawn by artificial agents, we set up a pilot study in which a `sender' agent, pretrained in five different game configurations on STL10, is paired up with a human `receiver' to play the visual communication game. For this pilot study, we collect results from 6 human participants. Each participant played a total of 150 games, i.e. had to select the target image for each of the 150 sketches drawn by a pretrained sender. Depending on the game setting, the list of options differs, but it is composed of distractors and the true target image. The experimental setup is detailed in \cref{app-sec:humaneval-details}. \Cref{tab:humaneval-nofeedback} compares the averaged human gameplay success to that of a trained `receiver' agent. The results show that the addition of the perceptual loss leads to statistically significant improvement of humans' ability to recognise the identity of sketches. For the original game setting, played in this study with $K=9$ distractors which might be of the same category as the target, we also assess the ability of participants to recognise the class of the sketch. The human class communication rate shows that humans are better at determining the class of the sketch rather than the specific instance, even in the case of sketches generated with the game loss only. In the appendices, we extend the discussion of these results and look into whether communication with an agent can be improved if the human participants are allowed to learn via feedback.

\begin{table}
    \caption{\textbf{Human Evaluation results, no learning allowed.} Trained agents communicate successfully between themselves in all settings. Addition of the perceptual loss allows humans to achieve significantly better than random performance (images from STL-10, original games have 9 distractors/game for these experiments \& random chance is 10\%). In addition, humans are better at guessing the correct image class when the models are trained with the additional perceptual loss.}
    \label{tab:humaneval-nofeedback}
    \centering
    \begin{tabular}{llllll}
    \toprule
         &  &  & Agent & Human & Human \\ 
        Game & Loss & Lines & comm. rate & comm. rate & class comm. rate \\ 
        \midrule
        original & $l=l_{game}$ & 20 & $100\%$ & $8.3\%\ (\pm5.4)$ & $15.0\%\ (\pm2.5)$ \\ 
        original & $l=l_{game} + l_{perceptual}$ & 20 & $93.3\%$ & $38.3\%\ (\pm2.5)$ & $55.6\%\ (\pm7.1)$ \\ 
        original & $l=l_{game} + l_{perceptual}$ & 50 & $100\%$ & $37.2\%\ (\pm5.9)$ & $47.8\%\ (\pm7.4)$ \\ 
        oo diff & $l=l_{game} + l_{perceptual}$ & 20 & $83.3\%$ & $23.9\%\ (\pm6.2)$ & $23.9\%\ (\pm6.2)$ \\ 
        oo diff & $l=l_{game} + l_{perceptual}$ & 50 & $90.0\%$ & $ 38.9\%\ (\pm9.9)$ & $38.9\%\ (\pm9.9)$ \\ \bottomrule
    \end{tabular}
\end{table}

\section{Conclusions and Future Work}\label{sec:conclusion}

We have demonstrated that it is possible to develop and study an emergent communication system between agents where the communication channel is visual. Further, we have shown that a simple addition to the loss function (that is motivated by biological observations) can be used to produce messages between the agents that are directly interpretable by humans. 

The immediate next steps in this line of work are quite clear. It is evident from our experiments that the incorporation of the perceptual loss dramatically helps produce more interpretable images. One big question to explore in the future is to what extent this is influenced by the original training biases of the backbone network --- are these drawings produced as a result of the original labels of the ImageNet training data, or are they in some way more generic than that? We plan to address this by exploring what happens if the weights of the backbone are replaced with ones learned through a self-supervised learning approach like Barlow twins~\citep{zbontar2021barlow}. We would also like to explore what happens if the agents' visual systems had independent weights.

Going further, as previously mentioned, learning a perceptual loss would be a good direction to explore, but perhaps this should also be coupled with a top-down attention mechanism based on the latent representation of the input. An open question from doing this would be to ask if this allows for a richer variation in drawing, and for features to be exaggerated as in the case of a  caricature. Such an extension could also be coupled with a much richer approach to drawing, with variable numbers of strokes, which are not necessarily constrained to being straight lines. Coupling feedback or attention into the drawing mechanism itself could also prove to be a worthy endeavour.

We hope that this work lays the groundwork for more study in this space. Fundamentally our desire is that it provides the foundations for exploring how different types of drawing and communication --- from primitive drawings through to pictograms, to ideograms and ultimately to writing --- emerges between artificial agents under differing environmental and internal constraints and pressures. Unlike other work that `generates' images, we explicitly focus on learning to capture \textit{intent} in our drawings. We recognise however that our work may have broader implications beyond just understanding how communication evolves. Could for example in the future we see a sketching agent replace a trained illustrator? The creation of messages for communication inherently involves elements of individual creative expression and adaption to the emotive environment of both the sender and receiver of the message. Our current models are clearly incapable of this, but such innovations will happen in the future. When they do we need to be prepared for the surrounding ethical debate and discussions about what constitutes `art'. This has already been seen in the domain of robot art in which Pix18~\citep{lipsonPix18} is a trailblazer as it is not only a robot that paints oil on canvas but can also conceive its own art subject with minimal human intervention.

\begin{ack}
D.M. is supported by the EPSRC Doctoral Training Partnership (EP/R513325/1). J.H. received funding from the EPSRC Centre for Spatial Computational Learning (EP/S030069/1).  The authors acknowledge the use of the IRIDIS High-Performance Computing Facility, the ECS Alpha Cluster, and associated support services at the University of Southampton in the completion of this work.
\end{ack}

\bibliographystyle{plainnat}
\bibliography{ms}

\newpage
\part{Appendices}
\parttoc 
\newpage

\appendix
\renewcommand{\thetable}{\Roman{table}}   
\renewcommand{\thefigure}{\Roman{figure}}

\setcounter{table}{0}
\setcounter{figure}{0}

\section{How does the sketch complexity influence communication?}
\label{app-complexity}

An interesting question one might ask about a model that learns to communicate by drawing is how complex the sketch image needs to be so that its meaning can be conveyed successfully and communication can be established. We attempt to answer this question by varying the number of lines that our model is allowed to draw to represent the input photograph. In \cref{tab:drawing-complexity}, we show results for experiments run with 5, 10 and 20 lines allowed for sketching. As with previous experiments, we provide the communication success rate with standard deviation over 10 seeds and qualitative results under two game setups. Under the original game format, contrary to what one might expect, the communication rate decreases as the number of lines is increased (see also \cref{tab:drawing-complexity2}). From a visual point of view, using more lines results in sketches that are more interpretable to a human observer, although that does not seem to correlate with the agent's communication strategy. Varying the complexity of drawings in the object-oriented game does not significantly influence the communication rate. The sketches, however, show once more that such a setup can induce a more interpretable communication channel. It is clear that even when drawing 5 lines, the model is trying to capture the overall shape of the object. 

\begin{table}[h!]
    \caption{\textbf{The effect of the drawing complexity (5, 10 or 20 line strokes) on the emergent visual communication channel.} The communication success rate (i.e. receiver agent correctly guessing the target image) and standard deviation across 10 runs are shown next to sample sketches.}
    \label{tab:drawing-complexity}
    \centering
    \begin{tabular}{cccc}
        \toprule
        & 5 & 10 & 20\\
        \midrule
        Original game & $73.41\%\ (\pm 1.6)$ & $69.48\%\ (\pm 3.3)$ & $69.57\% \ (\pm 2.6)$  \\
        \includegraphics[trim=196 882 294 294,clip,width=0.22\textwidth]{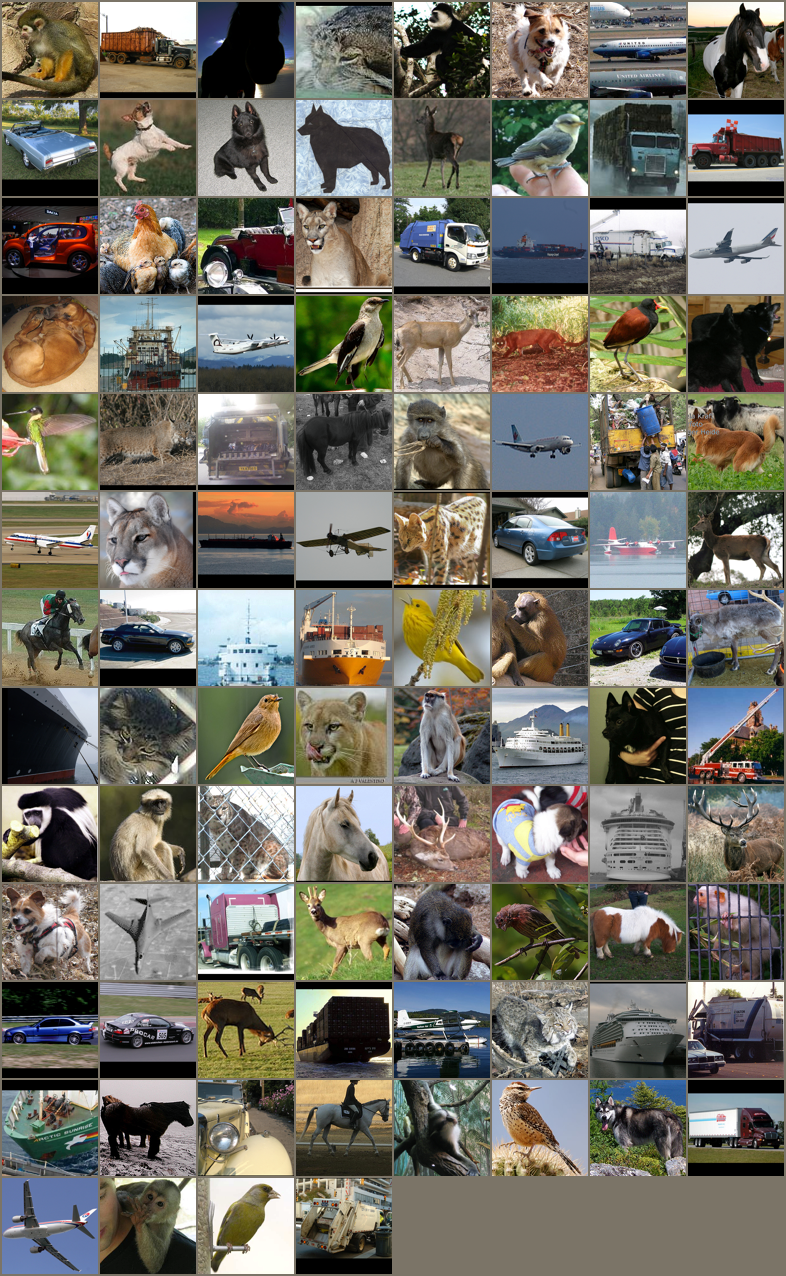} &
        \includegraphics[trim=196 882 294 294,clip,width=0.22\textwidth]{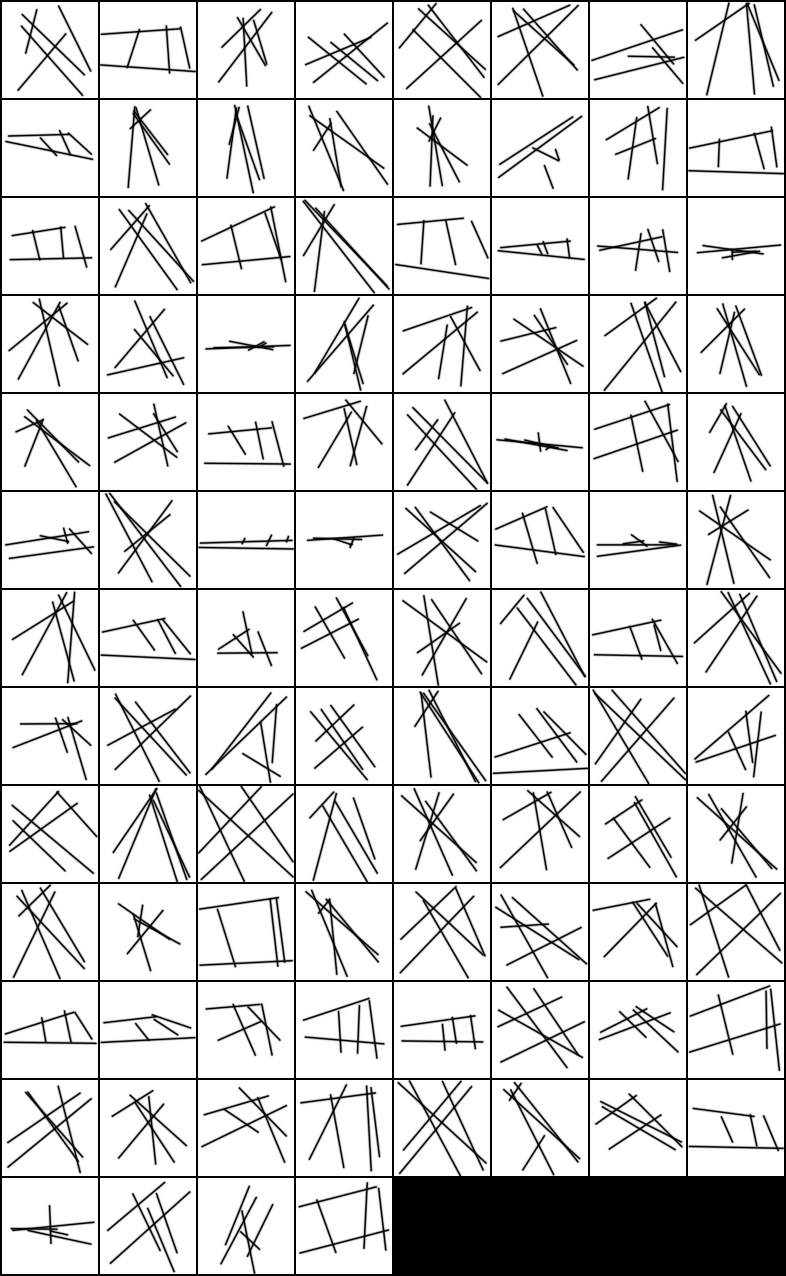} &
        \includegraphics[trim=196 882 294 294,clip,width=0.22\textwidth]{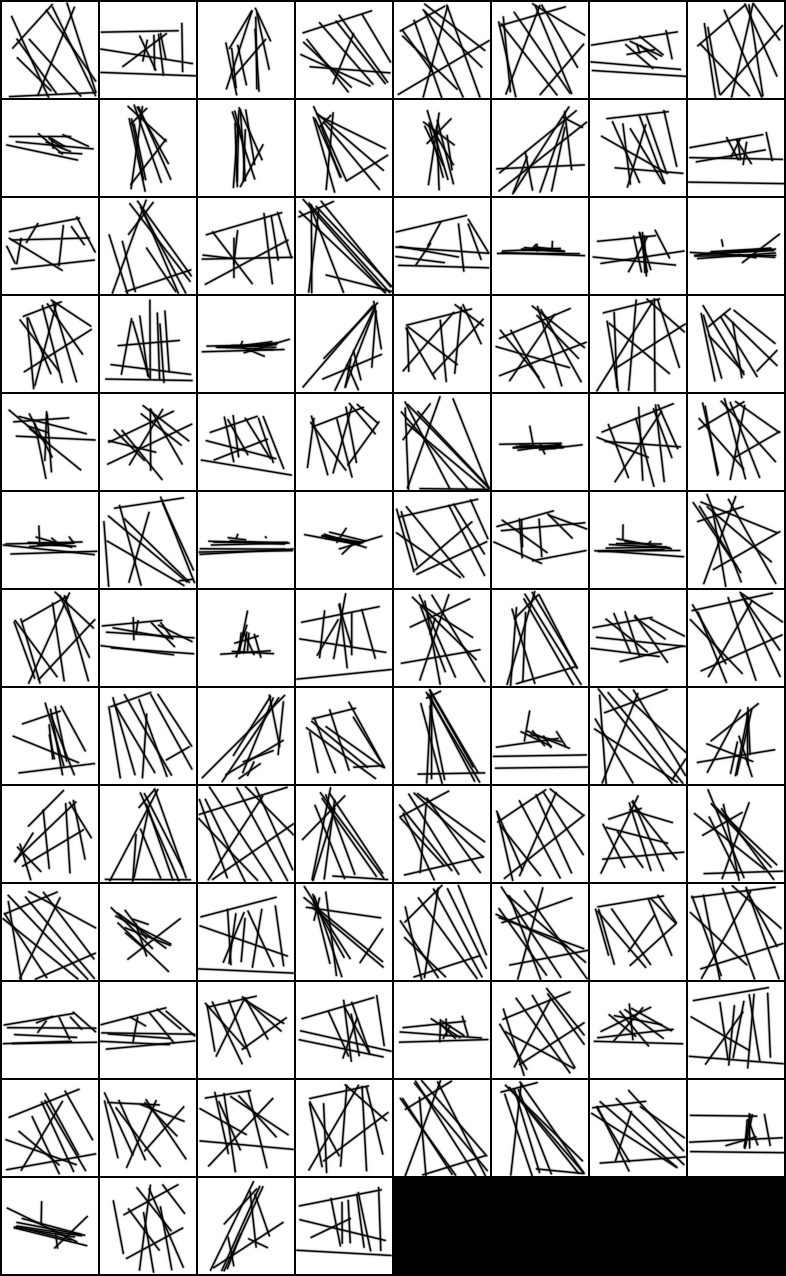} &
        \includegraphics[trim=196 882 294 294,clip,width=0.22\textwidth]{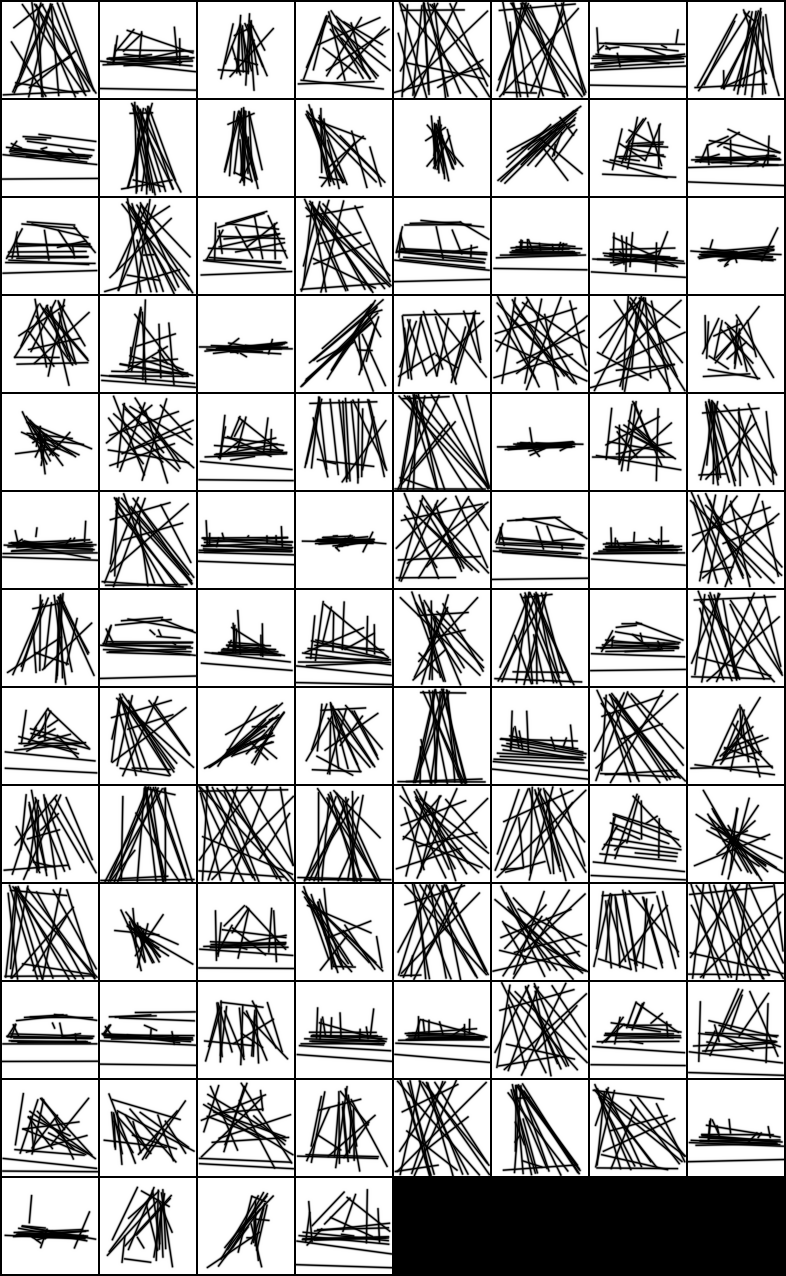} \\
        \midrule
        OO-game different & $80.69\%\ (\pm 1.1)$ & $80.9\%\ (\pm 0.6)$ & $81.09\% \ (\pm 0.6)$\\
        \includegraphics[width=0.22\textwidth]{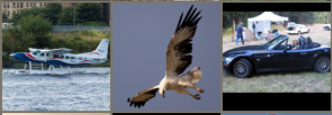} &
        \includegraphics[width=0.22\textwidth]{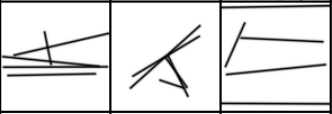} &
        \includegraphics[width=0.22\textwidth]{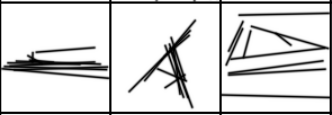} &
        \includegraphics[width=0.22\textwidth]{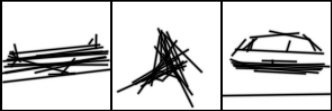} \\
        \bottomrule
    \end{tabular}
\end{table}

Further, we show results with increased number of strokes, in the original game setting, in \cref{tab:drawing-complexity2}. Compared to the model trained to draw with 20 lines in the original game setting (see \cref{tab:drawing-complexity})  which tries to cover the overall space occupied by the photograph's main object, the models trained with more strokes start to draw different lengths, and thus, the object becomes visually more recognisable.

\begin{table}[h!]
    \caption{\textbf{The effect of increasing drawing complexity (30, 40 or 50 lines) in the original game setting.} Sketches become visibly more correlated with the input photographs as the increase in the number of line allows for shorter strokes to be used which help with the overall interpretability.}
    \label{tab:drawing-complexity2}
    \centering
    \begin{tabular}{cccc}
        \toprule
        & 30 & 40 & 50\\
        \midrule
        Original game & $71.13\%\ (\pm 1.9)$ & $70.01\%\ (\pm 2.1)$ & $69.21\% \ (\pm 1.4)$  \\
        \includegraphics[trim=196 882 294 294,clip,width=0.22\textwidth]{images/drawing-complexity/test_samples_seed0.png} &
        \includegraphics[trim=196 882 294 294,clip,width=0.22\textwidth]{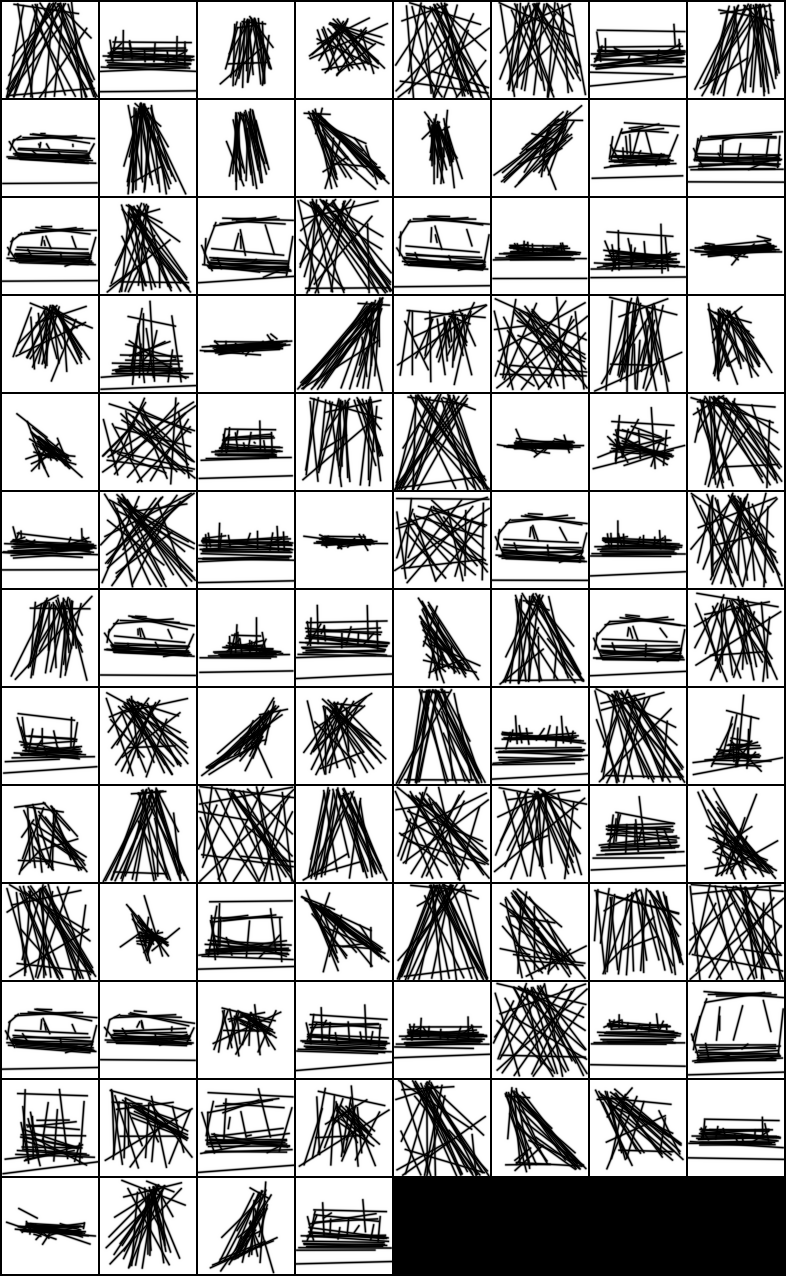} 
        &
        \includegraphics[trim=196 882 294 294,clip,width=0.22\textwidth]{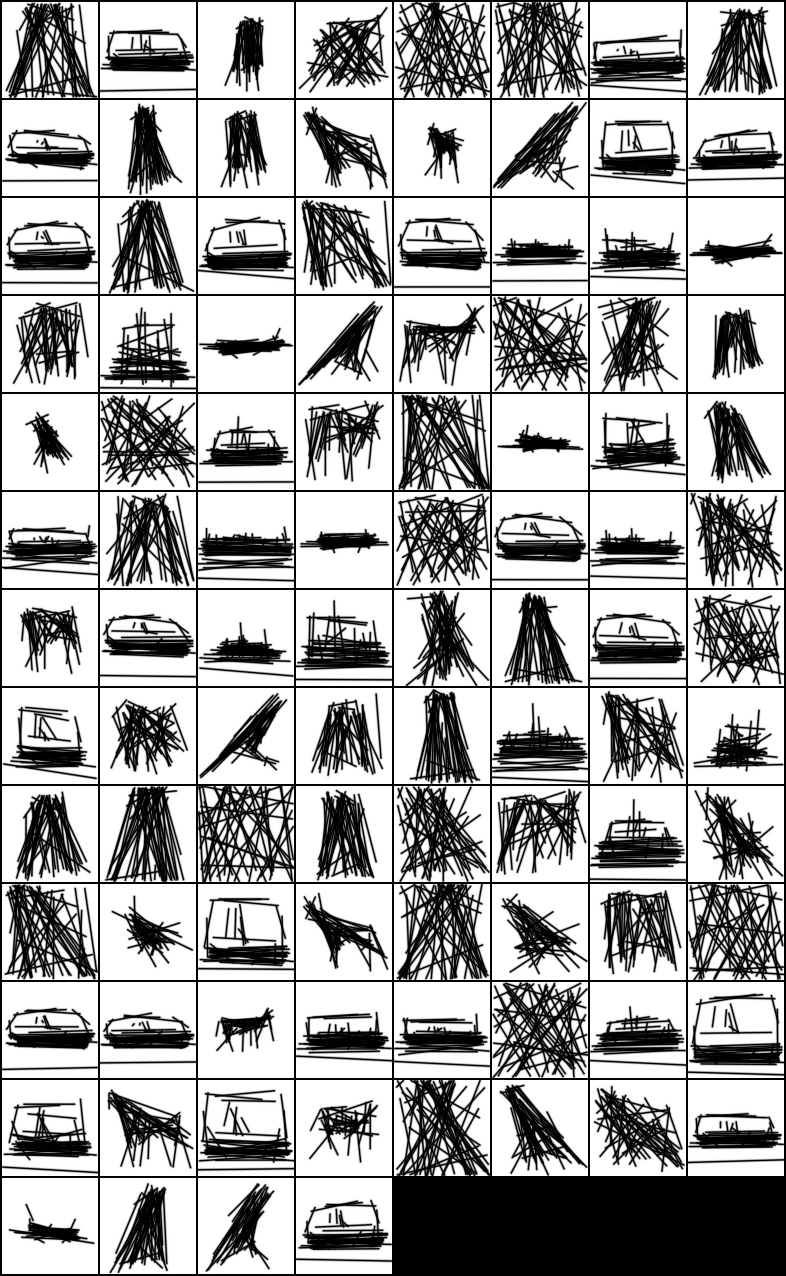} 
        &
        \includegraphics[trim=196 882 294 294,clip,width=0.22\textwidth]{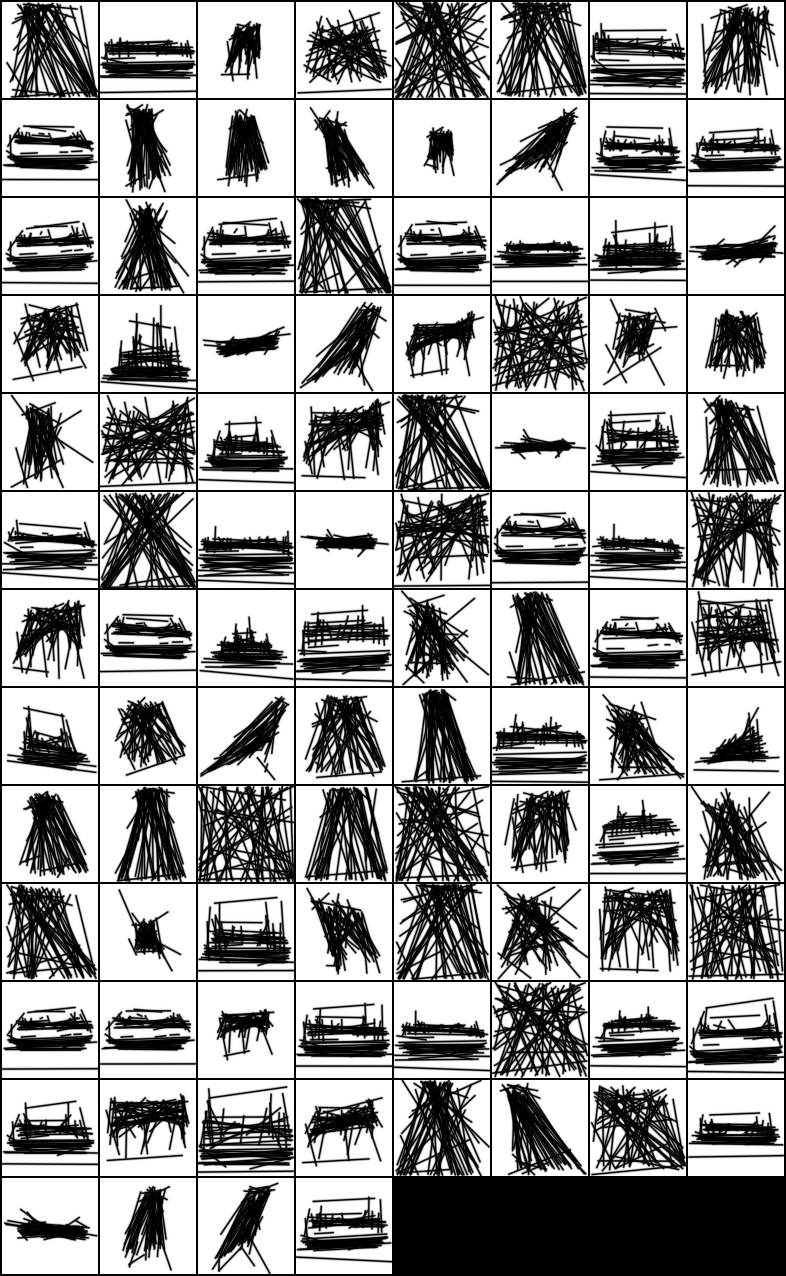} 
        \\
        \bottomrule
    \end{tabular}
\end{table}


\section{How important is the rasteriser?}
To further challenge our hypothesis about visual communication being possible between fully self-supervised agents, we ask the question of how important the rasteriser, and hence the sketch, is for the emergent communication protocol. Instead of line strokes, we constrain the agents to encode images into a cloud of points. We observe that communication between agents is definitely possible even when extracting as little as 10 points from an image, but the resulting sketch does not have any meaning to a human observer. When increasing the number of points to 50, or better 100, the communication success slightly drops to 0.71, 0.66 respectively, but object contours/shapes start to emerge in the sketches as shown in \cref{tab:drawing-points}. Encoding to a cloud of points is possible but less efficient, as it requires more coordinates to be learned to create sketches that are interpretable (to some extent) for humans.

\begin{table}[h!]
    \caption{\textbf{The effect of encoding the images into a cloud of points (10, 50, 100) in the original game setting.} Communication is possible with a points rasteriser, but more inefficient. More interpretable sketches require a larger number of points and, hence, more parameters to be learned.}
    \label{tab:drawing-points}
    \centering
    \begin{tabular}{cccc}
        \toprule
        Points & 10 & 50 & 100\\
        \midrule
        Original game & $75\% $ & $71\%$ & $66\%$  \\
        \includegraphics[trim=196 882 294 294,clip,width=0.22\textwidth]{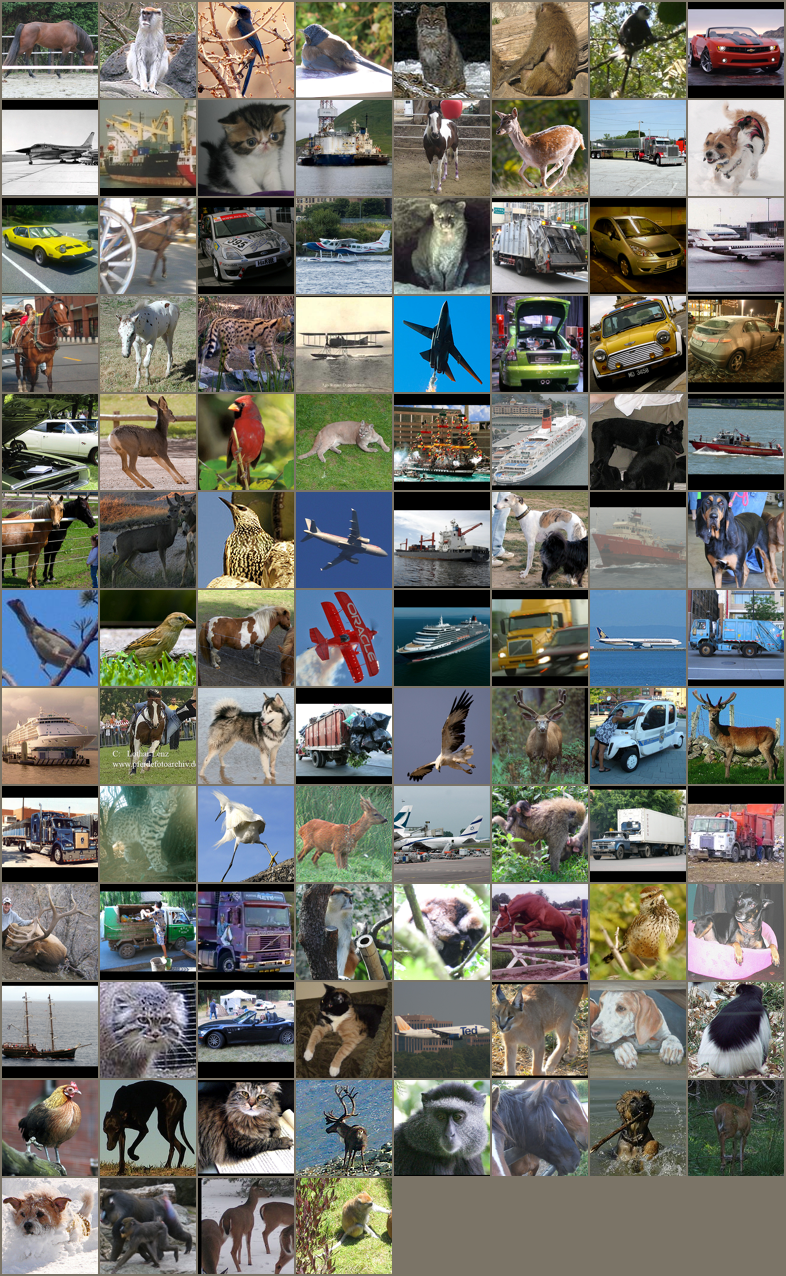} &
        \includegraphics[trim=196 882 294 294,clip,width=0.22\textwidth]{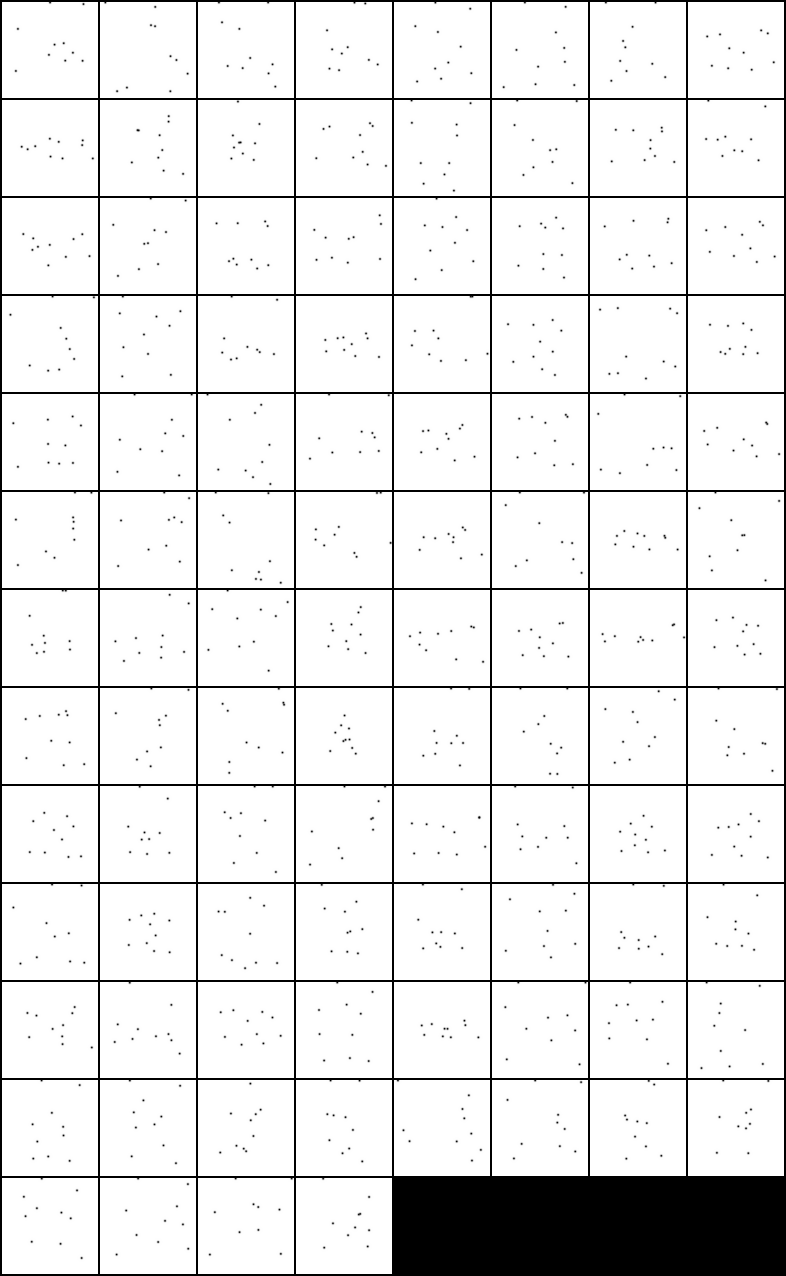} 
        &
        \includegraphics[trim=196 882 294 294,clip,width=0.22\textwidth]{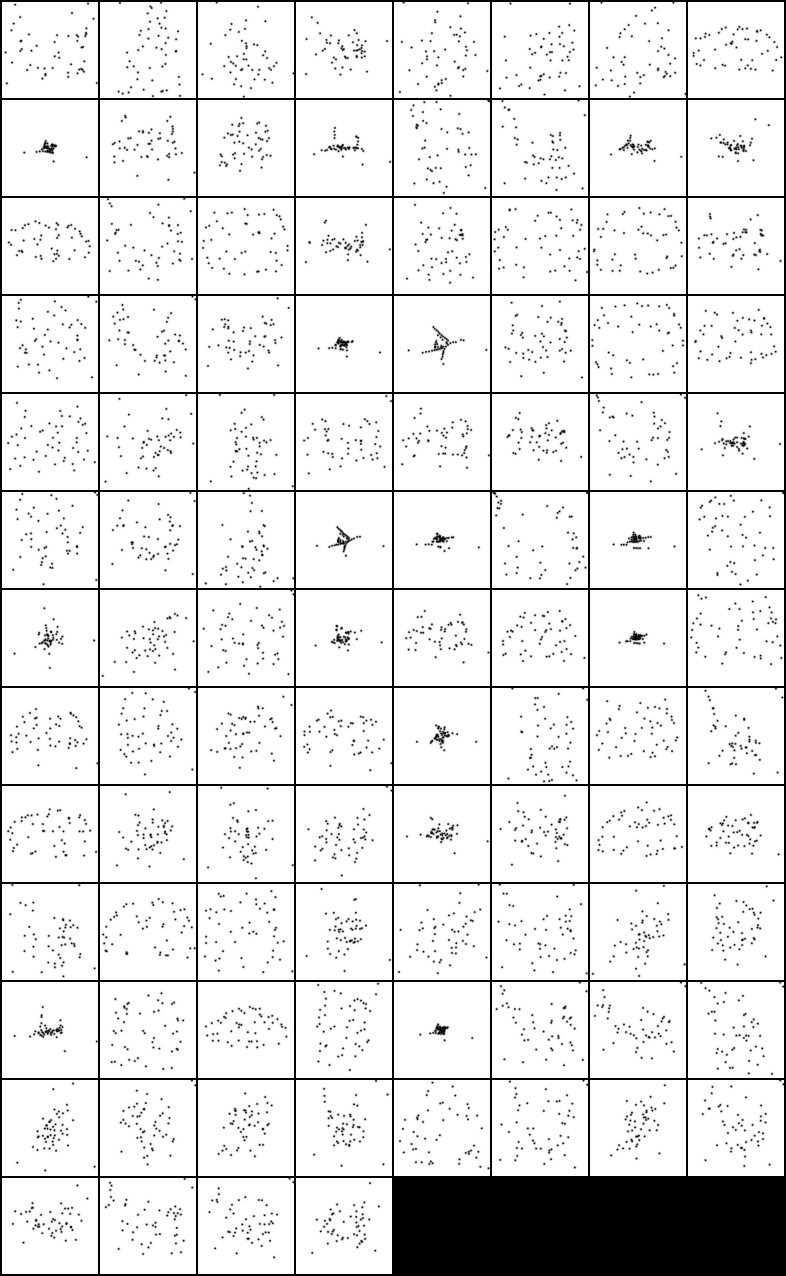} 
        &
        \includegraphics[trim=196 882 294 294,clip,width=0.22\textwidth]{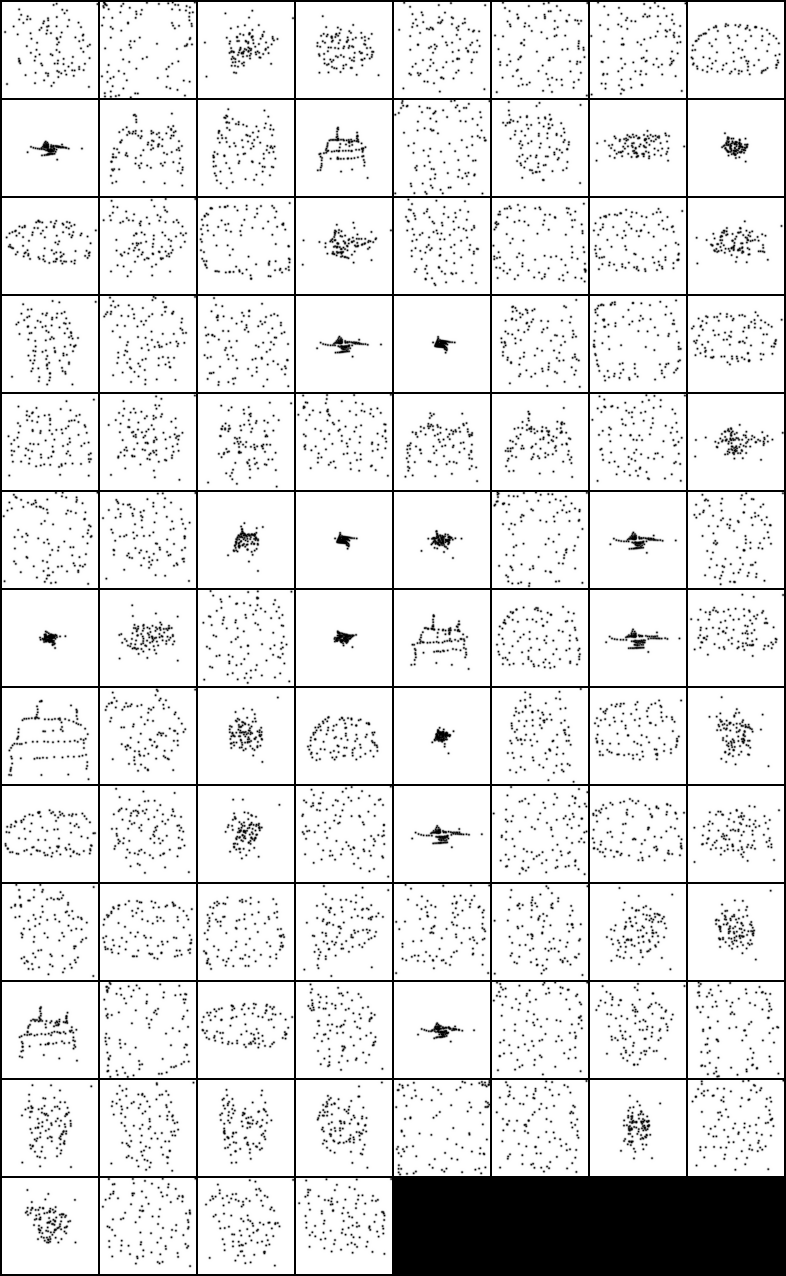} 
        \\
        \bottomrule
    \end{tabular}
\end{table}

\section{What is the impact of L in the computation of the perceptual loss on the emergent sketches?}

When computing the additional perceptual loss to induce sketches to become visually more similar to the target photographs, we use the outputs of $L=5$ feature maps, extracted from the VGG16 layers immediately before the max-pooling layers (\verb|relu1_2|, \verb|relu2_2|, \verb|relu3_3|, \verb|relu4_3| and \verb|relu5_3|), we will refer to this set of feature maps as $fmaps$. \Cref{tab:feature-maps-ablation} shows the effect of decreasing $L$ and using feature maps only up to the specified layer. More concretely, in the table results for \verb|relu4_3| show how the sketches look like when the perceptual loss is computed over features extracted after \verb|relu1_2|, \verb|relu2_2|, \verb|relu3_3|, \verb|relu4_3| only. We perform this ablation study in the original game setting with 20 line sketches, and show qualitative examples, the communication success rates averaged over 10 seeds and standard deviations. Note that these results are from when $L$ is changed for both sender and receiver agents. We observe that there is a drastic drop in the communication success rate as $L$ decreases from 5 to 4. Even more, if the perceptual loss is computed over the features extracted up to the third block of the VGG16 extraction network (i.e. anything up to \verb|relu3_3|), the model no longer converges and the communication completely fails.

Similarly, \cref{tab:feature-maps-ablation2} shows the effect of increasing $L$. To the original set of feature maps ($fmaps$) used in the computation of the perceptual loss,  the output of the other convolutional layers in a certain block (5, 4 or 3) of the VGG16 feature extraction network are added. The results show that increasing the number of feature maps neither impacts the communication success rate nor makes the sketch visually more similar to the corresponding image.

\begin{table}[h!]
    \caption{\textbf{Ablation study on the number of feature maps extracted from the visual system.} Studying the effect of decreasing the number of feature maps (L) extracted from the backbone VGG16 network. We present results by using features extracted from layers in $fmaps$ up to the specified layer.}
    \label{tab:feature-maps-ablation}
    \centering
    \begin{tabular}{cccc}
        \toprule
        & \verb|relu5_3| & \verb|relu4_3| & \verb|relu3_3|\\
        \midrule
        Original game & $69.57\%\ (\pm 2.6)$ & $19.09\%\ (\pm 10.1)$ & $1.0\% \ (\pm 0)$\\
        \includegraphics[trim=490 490 0 686,clip,width=0.22\textwidth]{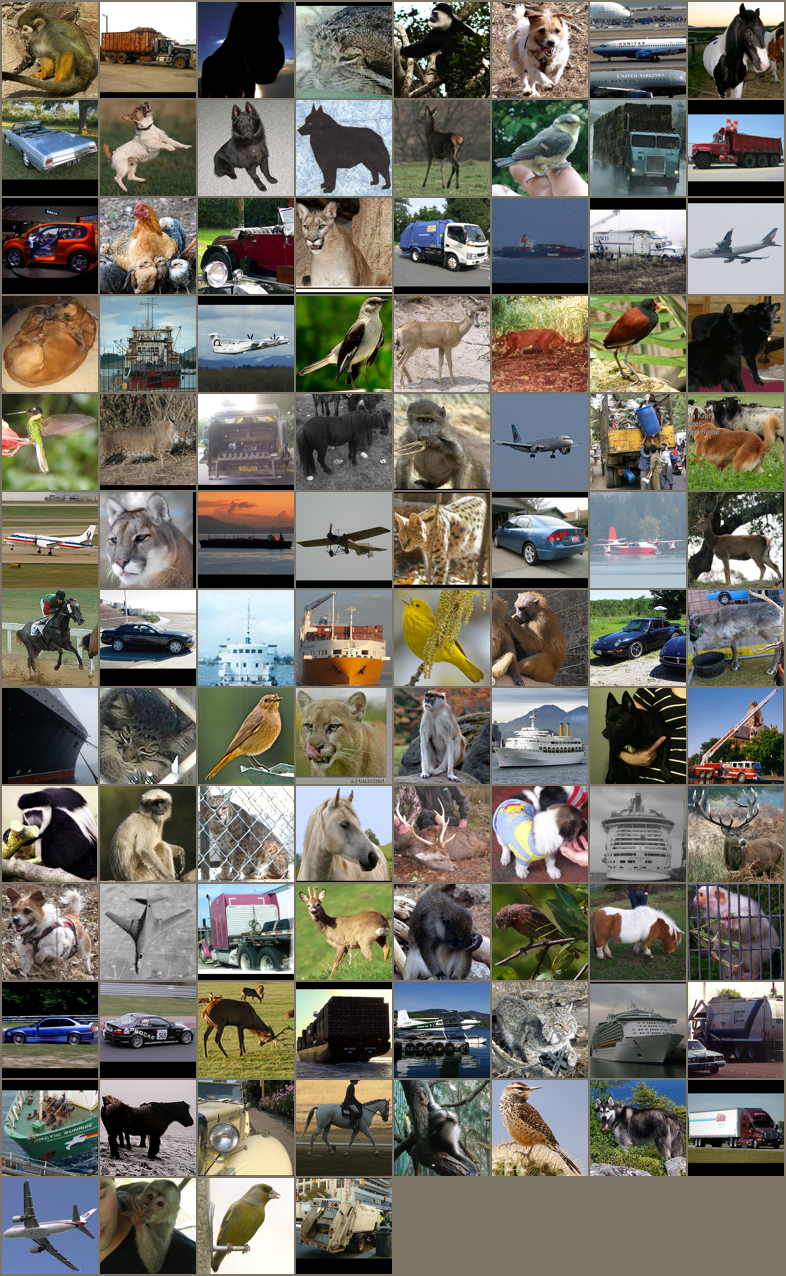} &
        \includegraphics[trim=490 490 0 686,clip,width=0.22\textwidth]{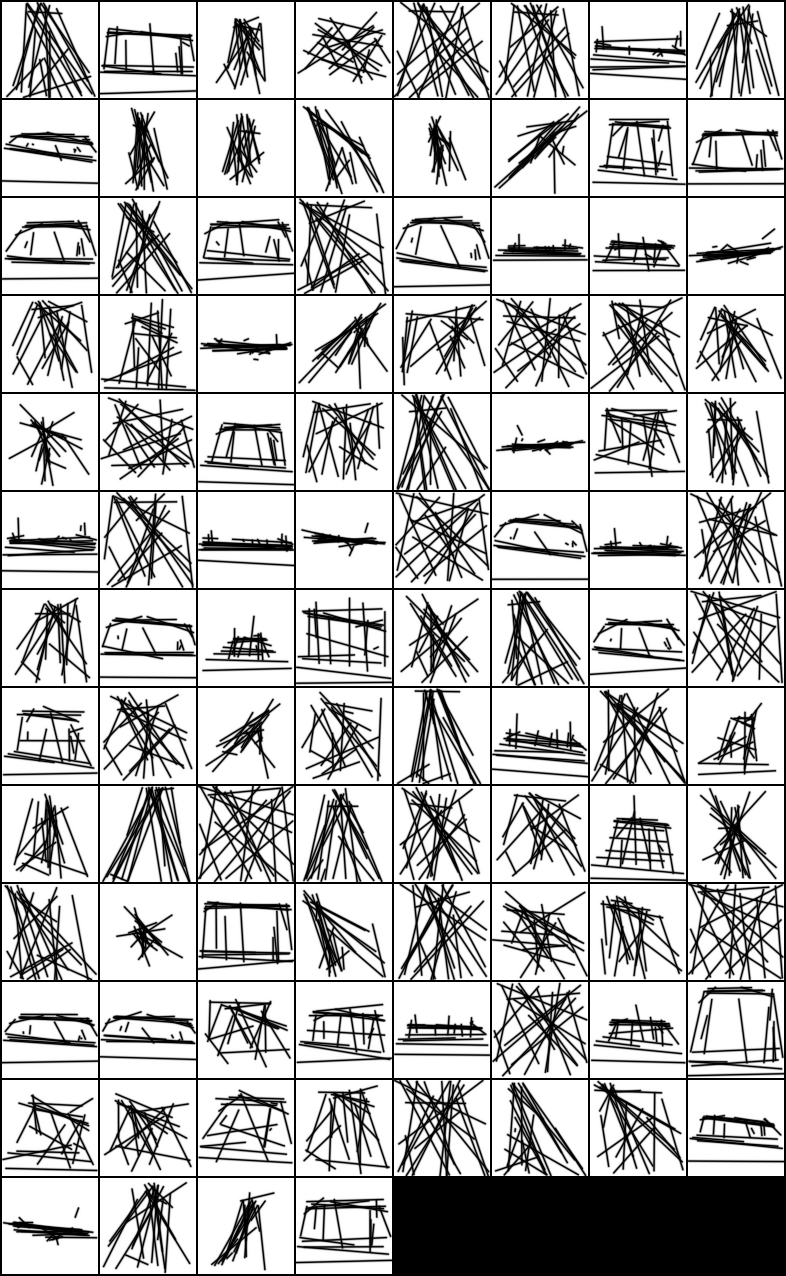} 
        &
        \includegraphics[trim=490 490 0 686,clip,width=0.22\textwidth]{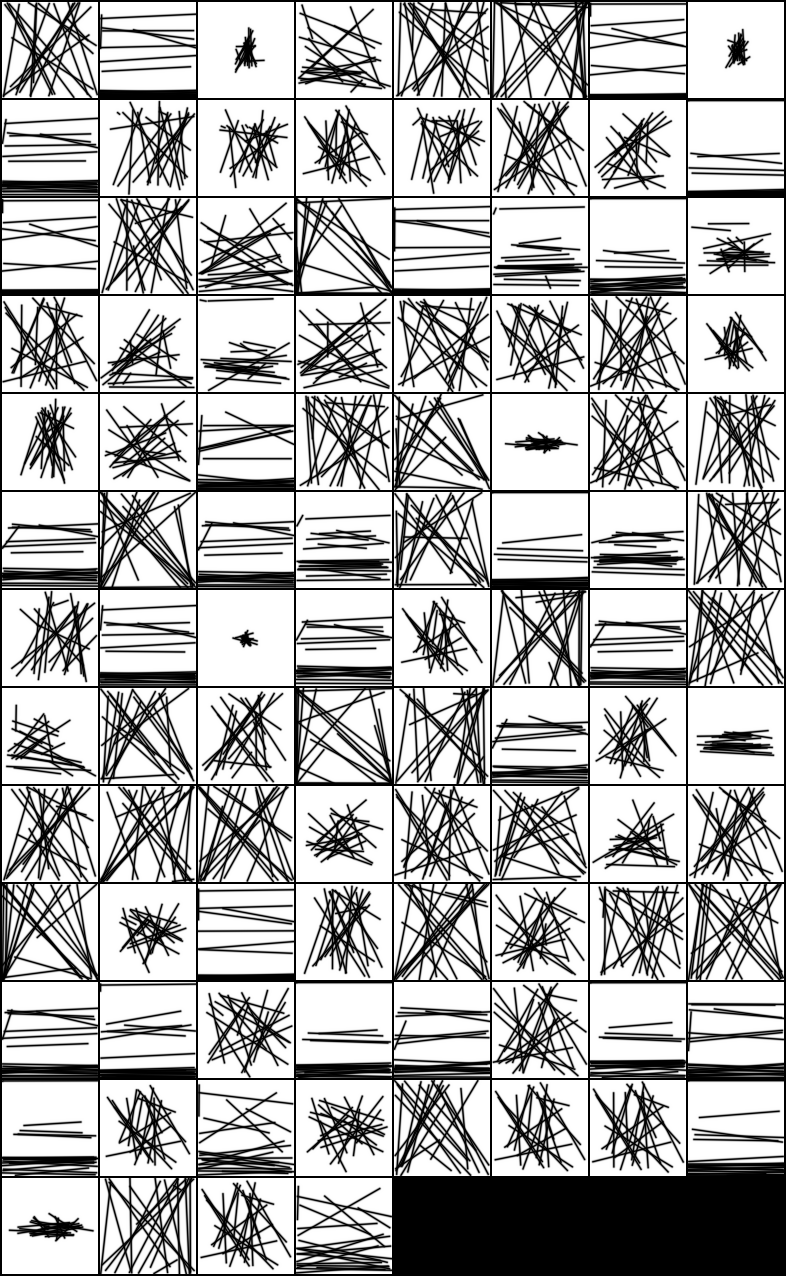} 
        &
        \includegraphics[trim=490 490 0 686,clip,width=0.22\textwidth]{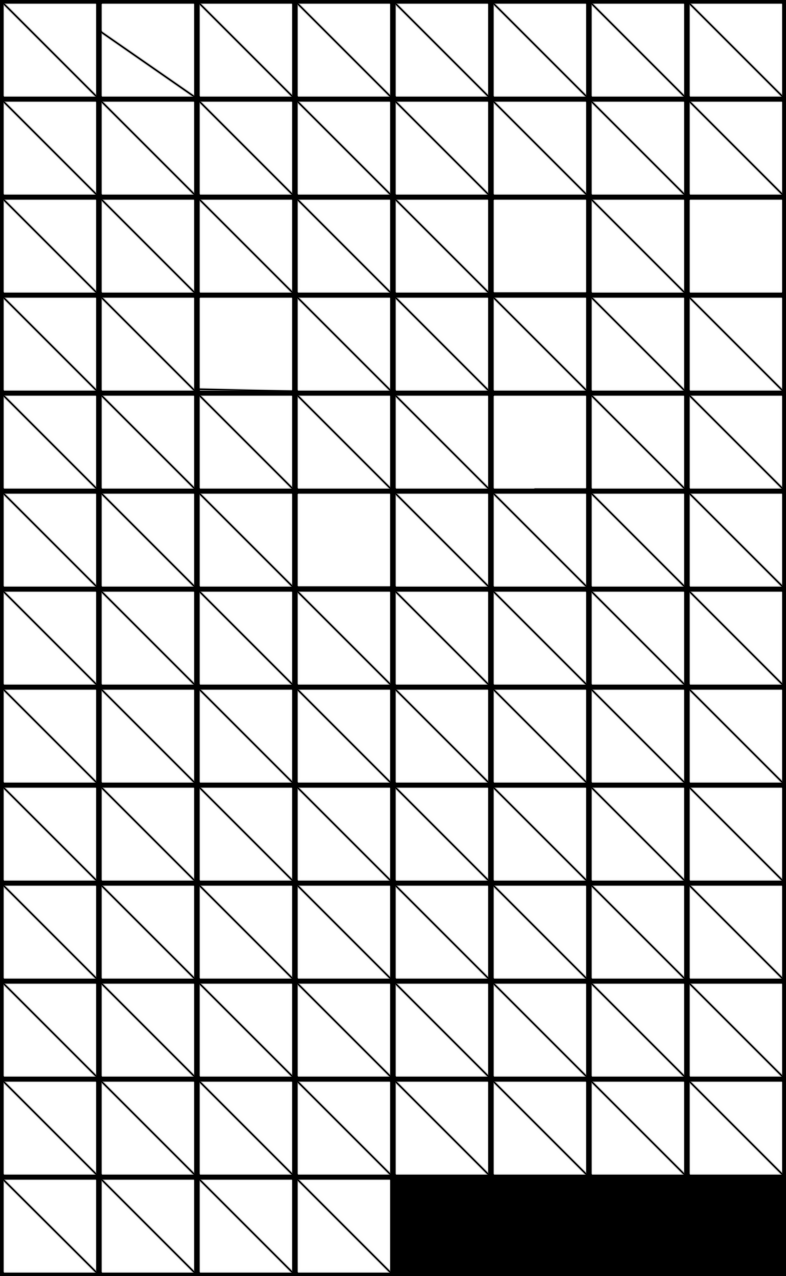} 
        \\
        \bottomrule
    \end{tabular}
\end{table}

\begin{table}[ht!]
    \caption{\textbf{Ablation study II on the number of feature maps extracted from the visual system.} Studying the effect of \textbf{increasing} the number of feature maps (L) extracted from the backbone VGG16 network. To the original set of feature maps, $fmaps$, we add the following extra layers: \texttt{reluX\_*} indicates that the other feature maps from the $X^{th}$ block of convolutions in the VGG16 feature extraction network are being used to compute the perceptual loss.}
    \label{tab:feature-maps-ablation2}
    \centering
    \begin{tabular}{cccc}
        \toprule
        & \verb|relu5_*| & \verb|relu5_*|, \verb|relu4_*| & \verb|relu5_*|, \verb|relu4_*|,\\
        & & & \verb|relu3_*|\\
        \midrule
        Original game & $68.4\%\ (\pm 2.0)$ & $69.5\%\ (\pm 1.8)$ & $69.0\% \ (\pm 1.2)$\\
        \includegraphics[trim=490 490 0 686,clip,width=0.22\textwidth]{images/feature-map-ablation/test_samples_seed0_fmablationstudy.png} &
        \includegraphics[trim=490 490 0 686,clip,width=0.22\textwidth]{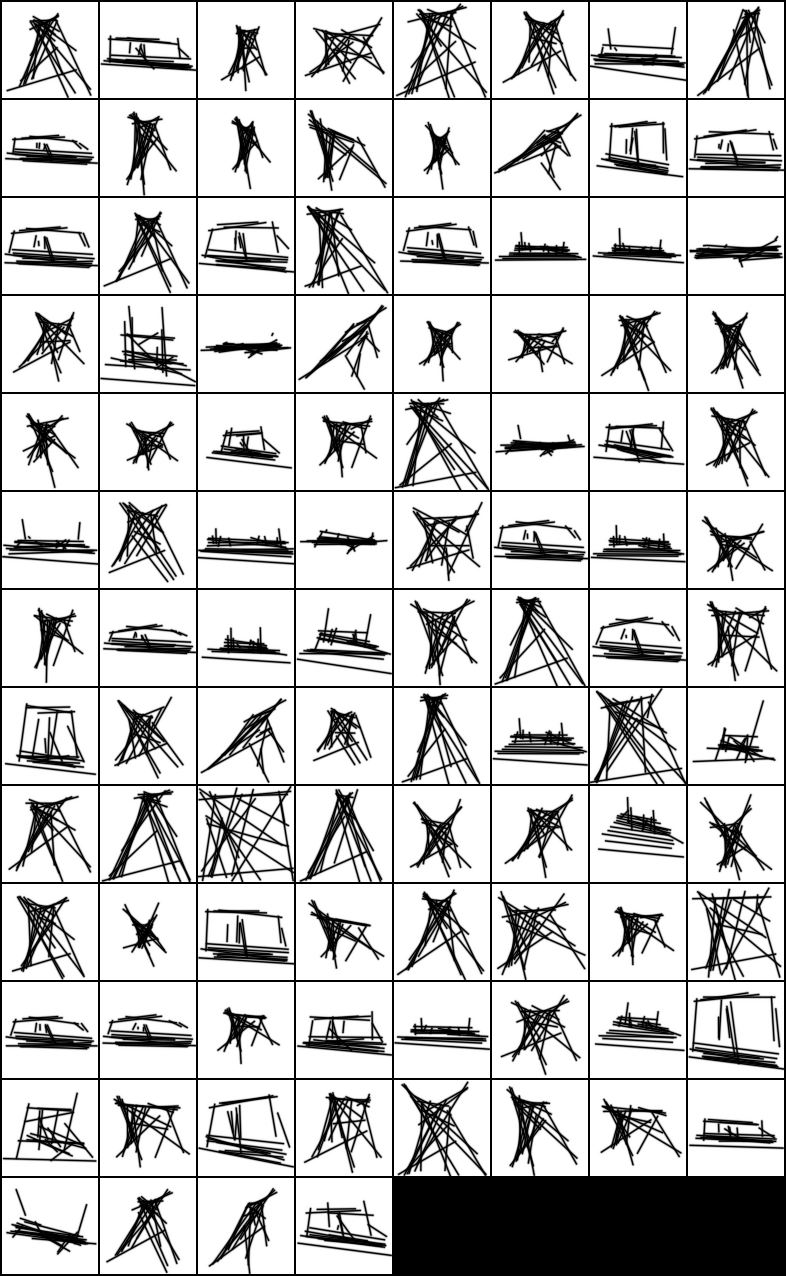} 
        &
        \includegraphics[trim=490 490 0 686,clip,width=0.22\textwidth]{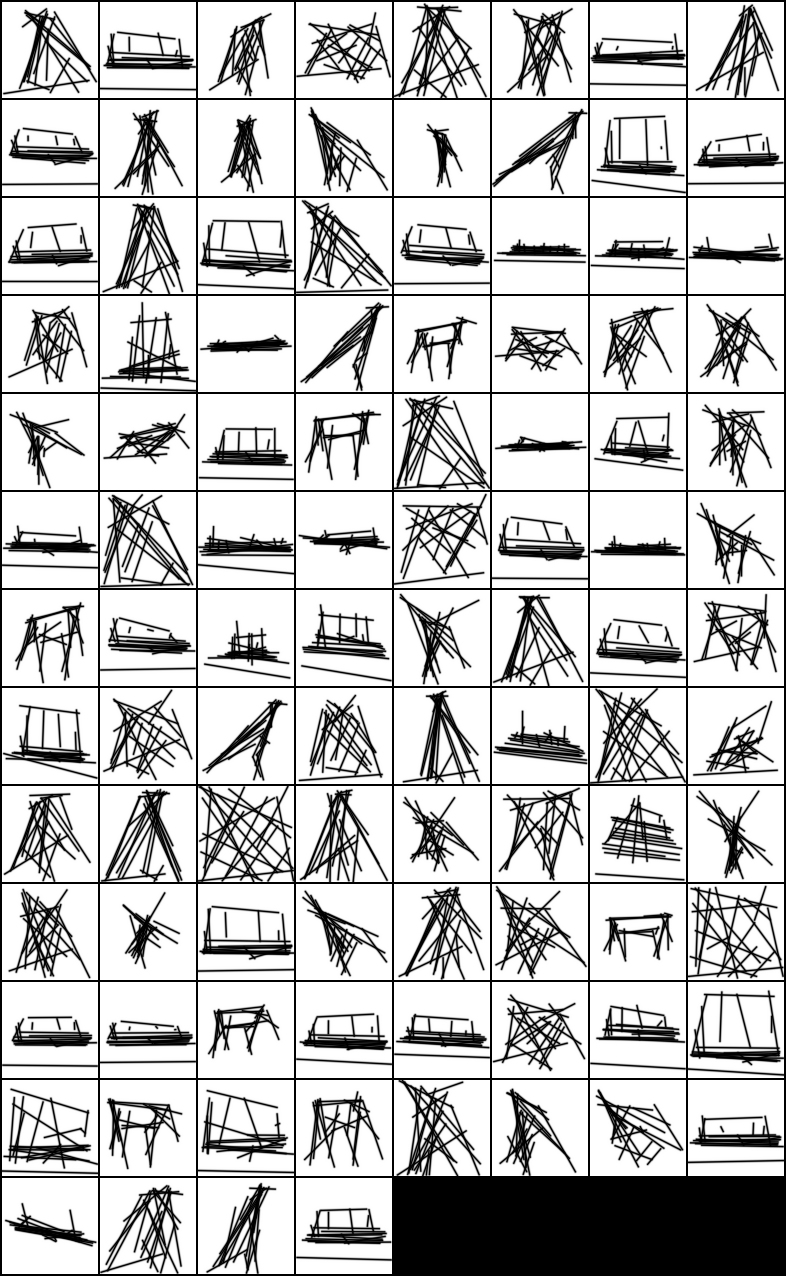} 
        &
        \includegraphics[trim=490 490 0 686,clip,width=0.22\textwidth]{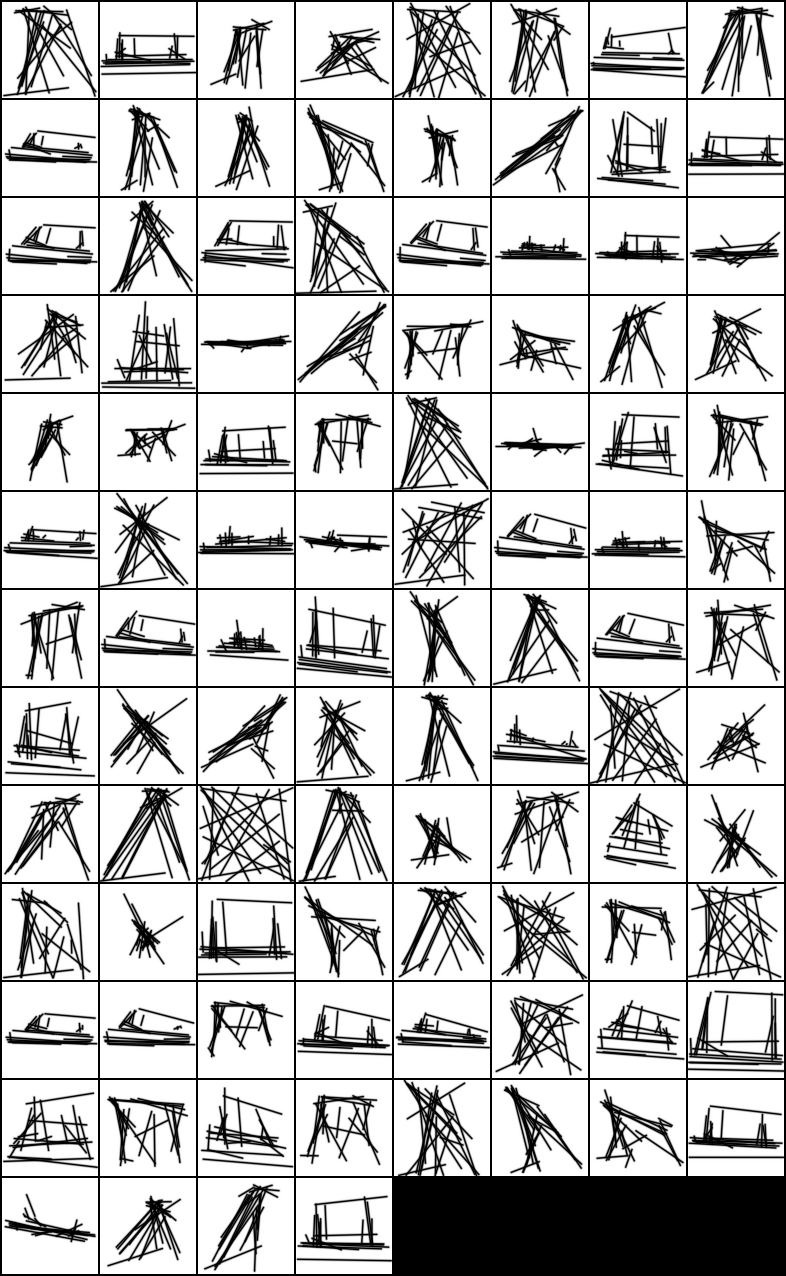} 
        \\
        \bottomrule
    \end{tabular}
\end{table}

\section{What happens if the communication is constrained under an arbitrary, meaningless objective?}

\begin{figure}[h!]
    \centering
    \begingroup
    \setlength{\tabcolsep}{2pt} 
    \renewcommand{\arraystretch}{0} 
    \setlength\extrarowheight{0pt}
    \begin{tabular}{m{0.1\linewidth}m{0.83\linewidth}}
        \includegraphics[width=\linewidth]{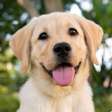} &  \includegraphics[trim=0 115 0 1255,clip,width=\linewidth]{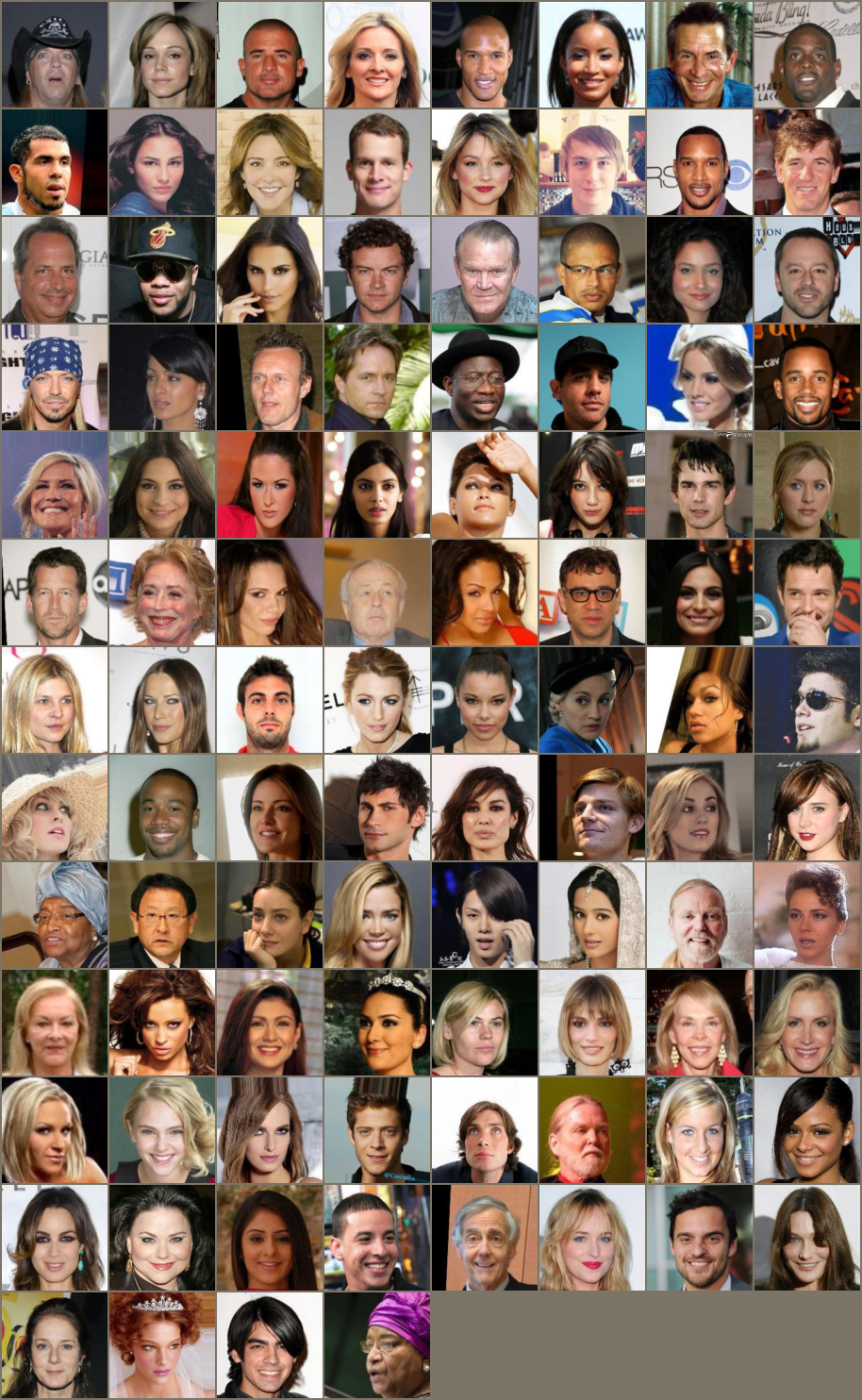}\\
         \begin{flushright} $\lambda=1.0$ (69.2\%) \end{flushright} &  \includegraphics[trim=0 115 0 1255,clip,width=\linewidth]{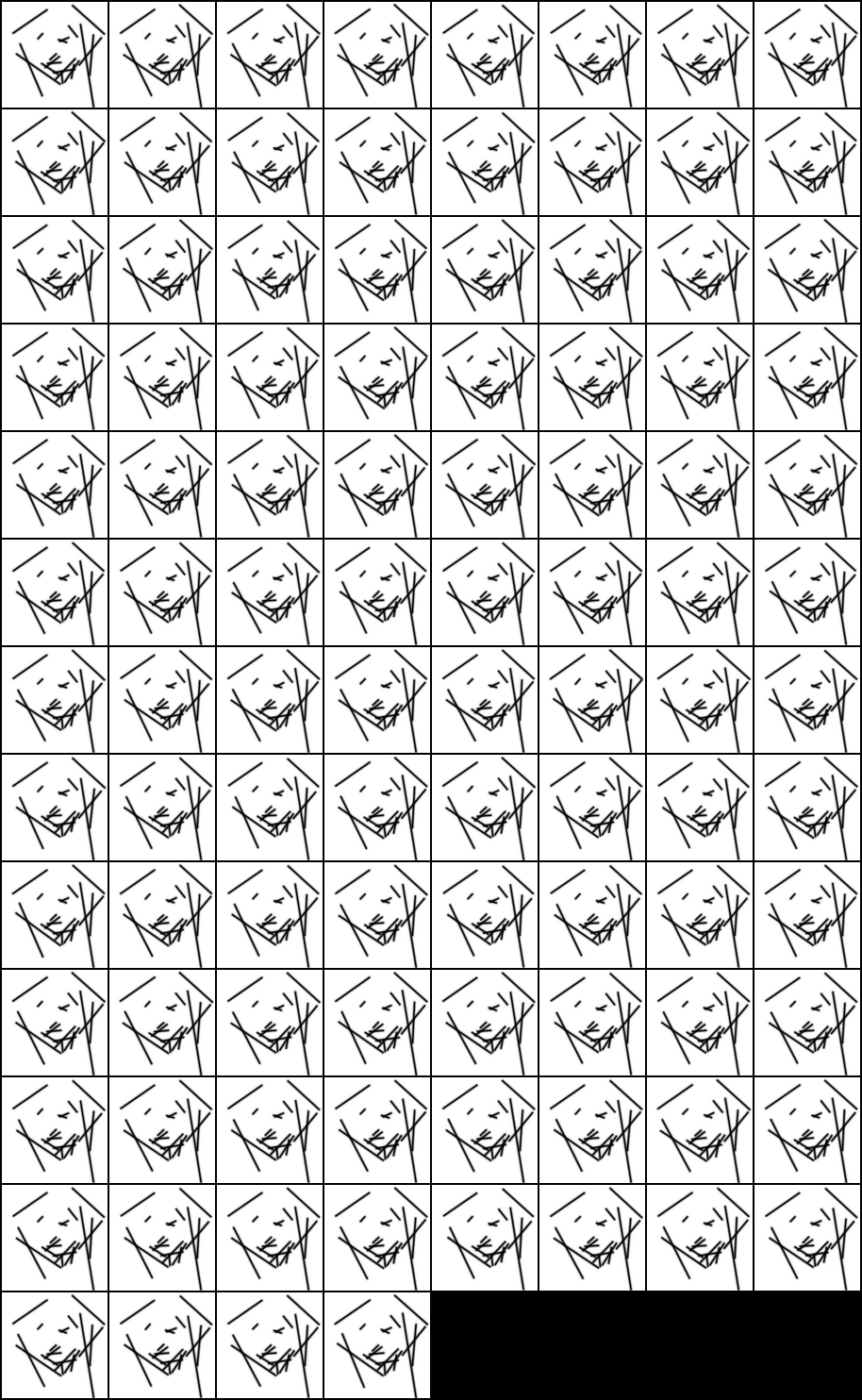}\\
         \begin{flushright} $\lambda=0.01$ (98.9\%) \end{flushright} & \includegraphics[trim=0 115 0 1255,clip,width=\linewidth]{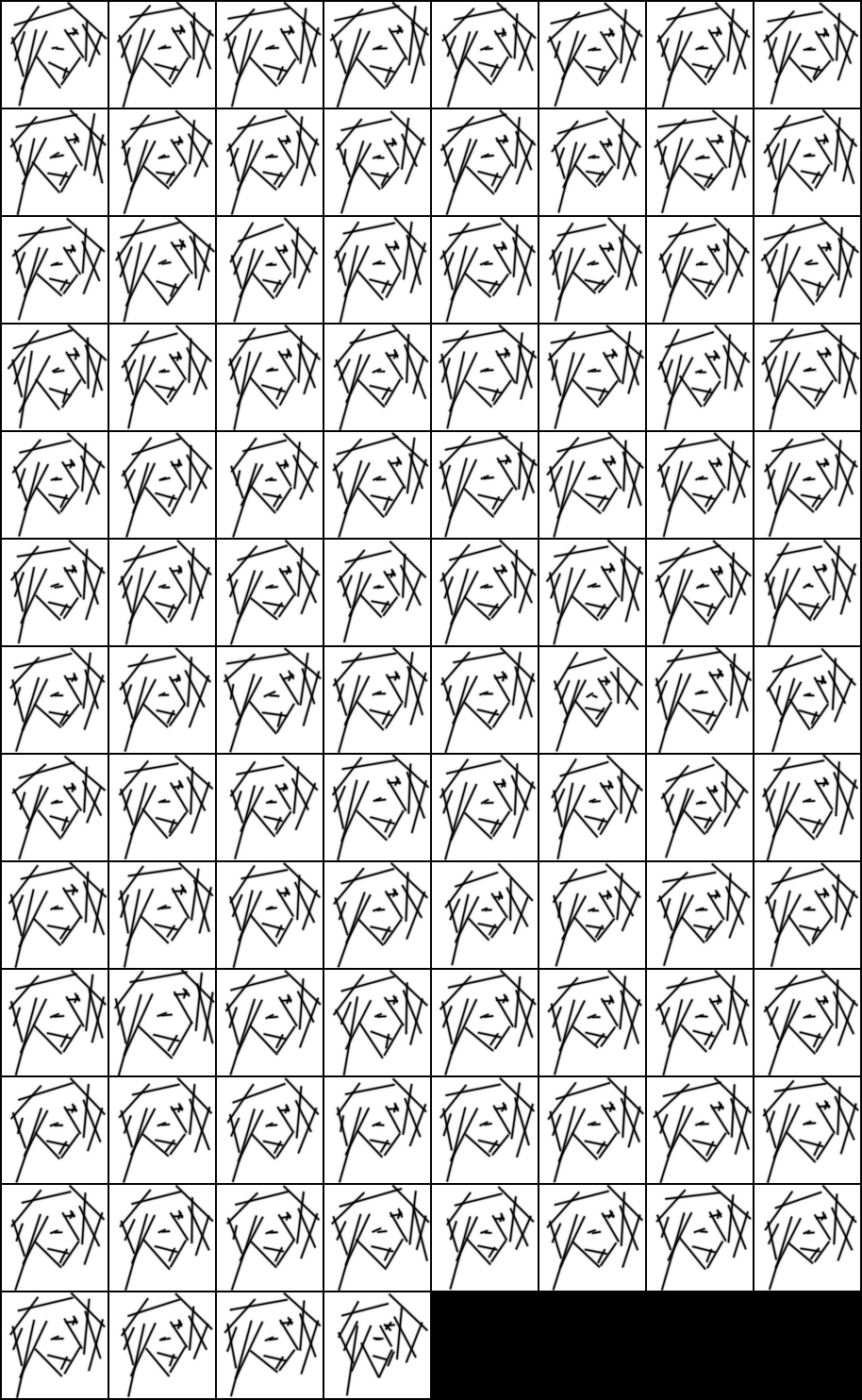}\\
    \end{tabular}
    \endgroup
    \caption{\textbf{Sketches from \textit{original} variant game using the CelebA dataset with an arbitrary objective: the sketches are constrained to look like the image of a dog (fully or partially, by scaling the perceptual loss coefficient $\lambda$).} Results are shown for a model with the visual extraction network pretrained on Stylized-ImageNet.}
    \label{fig:meaningless-objective}
\end{figure}

One might ask what happens to the communication protocol when the perceptual loss is replaced with some meaningless, arbitrary objective. To explore this scenario, we constrain sketches to look like a single image of a dog (shown in the top left of \cref{fig:meaningless-objective}) and train agents to draw in order to communicate about CelebA images. As one might expect, the artificial agents can still establish a successful communication strategy about the correct target even when constrained to draw dog-like sketches. \Cref{fig:meaningless-objective} show results for models trained with such an additional objective, fully or partially, by scaling $\lambda$ in $l=l_{game} + \lambda l_{arbitrary}$. These results show that it matters what the perceptual loss is: if it constrains sketches to look like the corresponding photographs, a human receiver might have a chance at recognising the person, but with such an arbitrary objective, humans stand no chance at understanding which image the sender agent tries to communicate about. Agents' communication success rate is also impacted (compared to the model with Stylized weights trained with our $\perceploss$ and $\lambda=1$, results shown in \cref{fig:celeba} of the paper).

\section{What happens when injecting Out-of-Distribution images?}
To further investigate the emergent visual communication protocol, we test a pair of agents pretrained in the proposed framework on out-of-distribution images. More specifically, we evaluate a pair of agents, previously trained in the \textit{original} setting on CelebA dataset, on games played with STL-10 images. Agents with ImageNet-pretrained visual systems, achieve a test communication rate on STL-10 of $15.8\%$. Similarly, agents initialised with Stylized-ImageNet weights achieve $30\%$ test recognition accuracy. It is worth noting that even if these results are significantly lower, they are still better than random chance, particularly with the stylized imagenet weights, where the sketches have considerably more diversity (but still all look like faces rather than the objects in the images).

\begin{figure}[tb]
    \centering
    \begingroup
    \setlength{\tabcolsep}{2pt} 
    \renewcommand{\arraystretch}{0} 
    \setlength\extrarowheight{0pt}
    \begin{tabular}{m{0.1\linewidth}m{0.83\linewidth}}
         &  \includegraphics[trim=0 115 0 1255,clip,width=\linewidth]{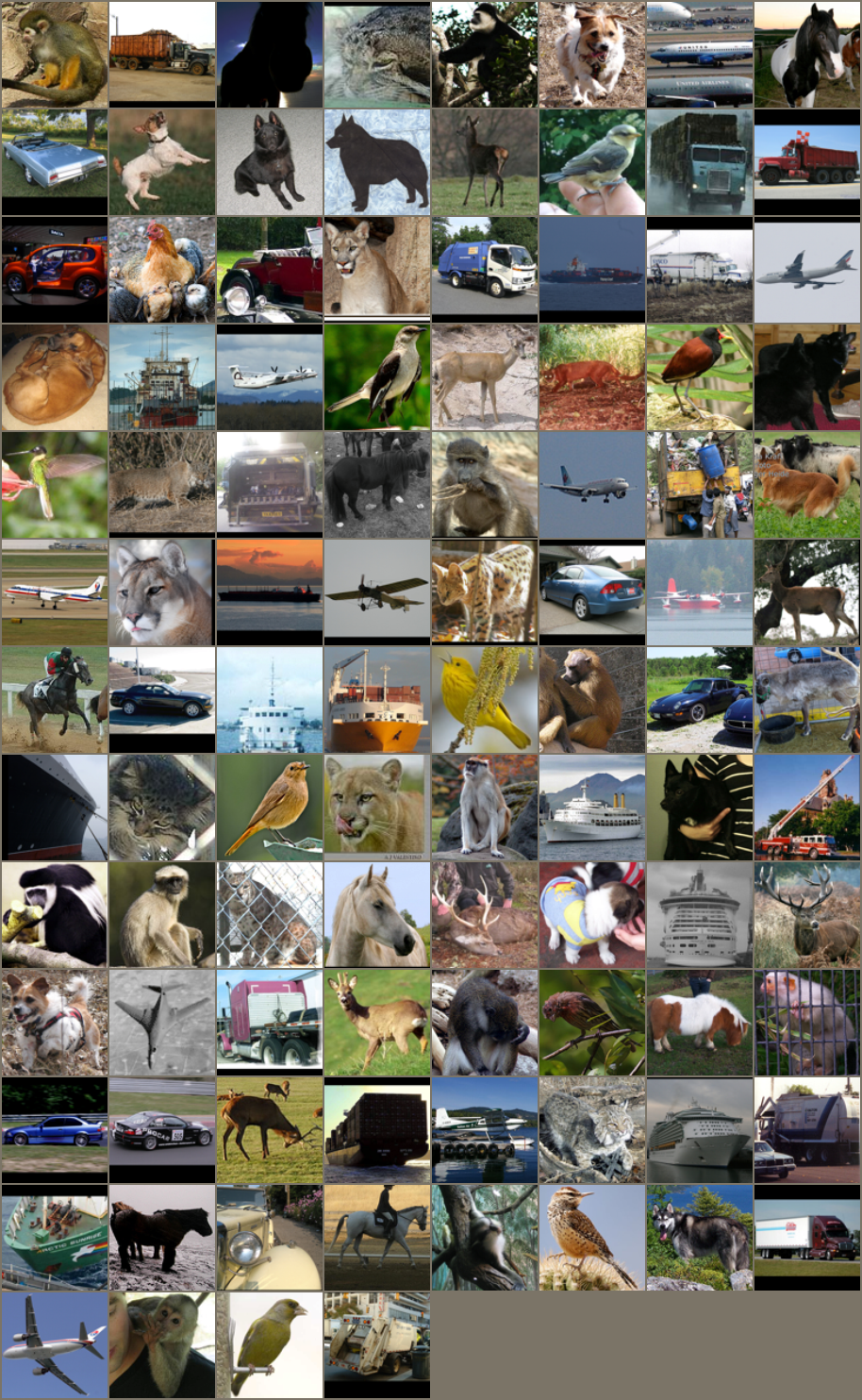}\\
         \begin{flushright} ImageNet weights $15.8\%$ \end{flushright} &  \includegraphics[trim=0 115 0 1255,clip,width=\linewidth]{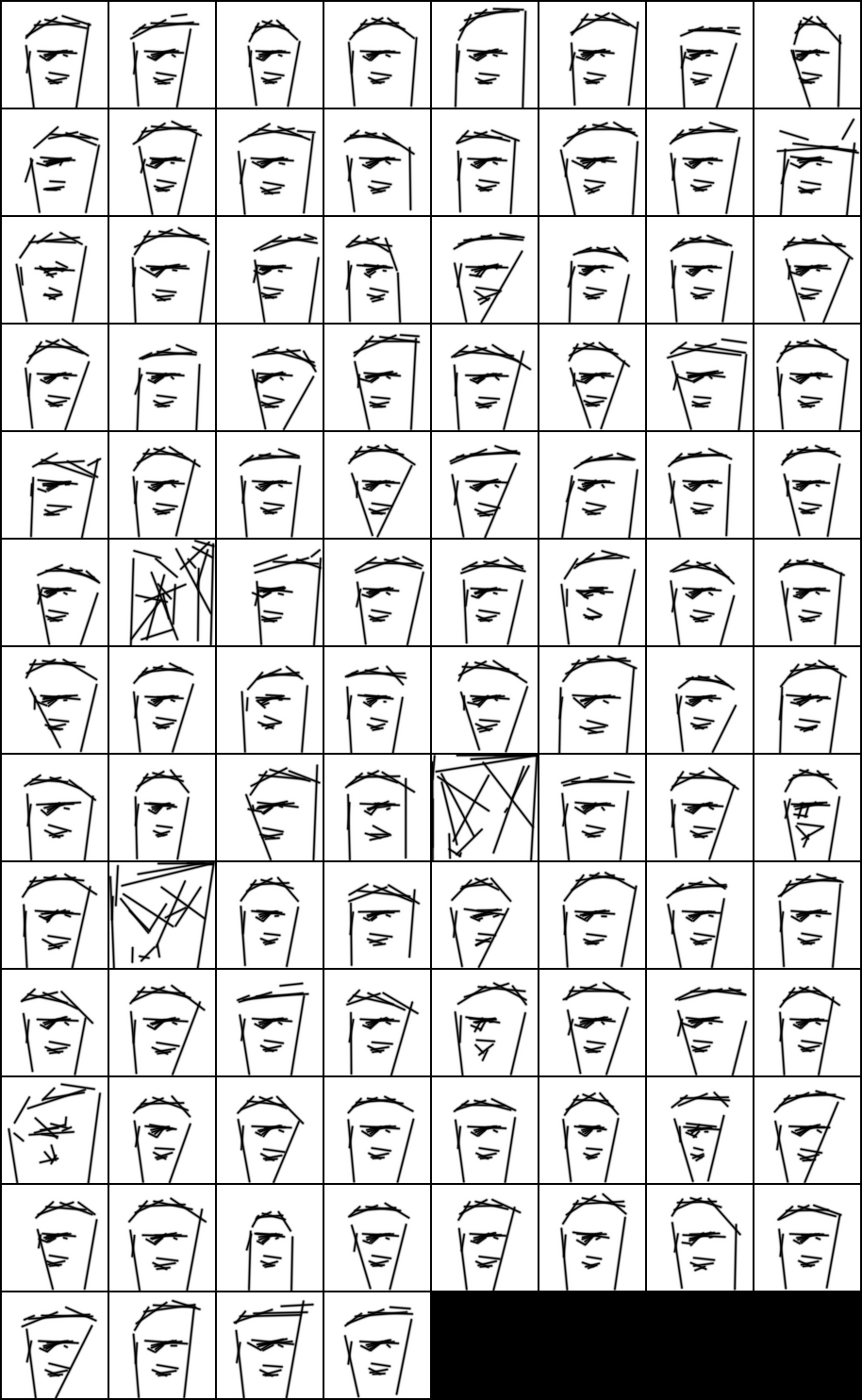}\\
         \begin{flushright} Stylized weights $30\%$ \end{flushright} & \includegraphics[trim=0 115 0 1255,clip,width=\linewidth]{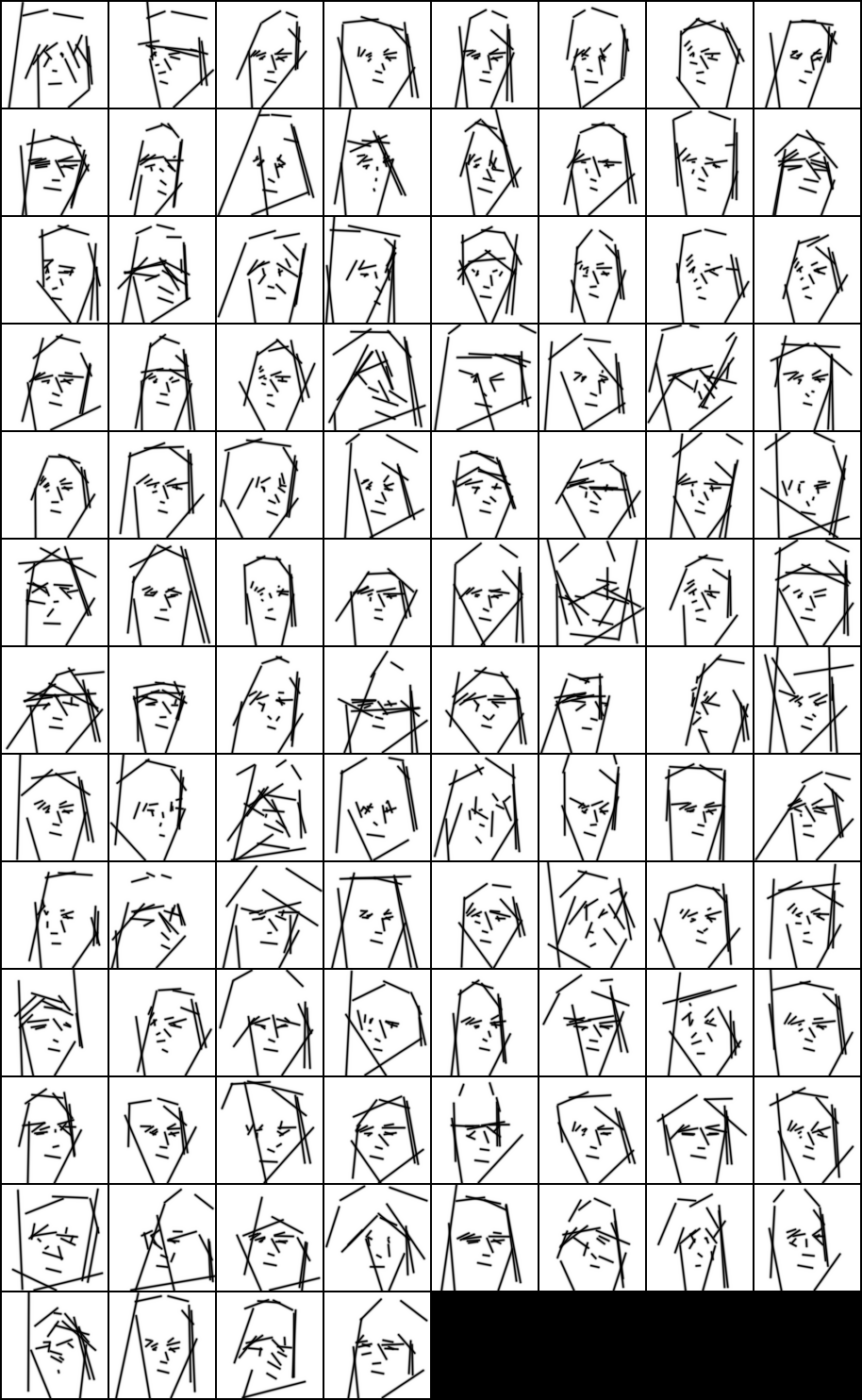}\\
    \end{tabular}
    \endgroup
    \caption{\textbf{Sketching agents, previously trained on CelebA (original game) tested on STL-10 test images.} We compare models with visual systems pretrained on ImageNet and Stylized-ImageNet.}
    \label{fig:ood-images-celebA-stl10}
\end{figure}

A similar experiment is performed with models pretrained on STL-10, with either just the $\gameloss$ or with the additional $\perceploss$. When testing these on Caltech-101 test data, the communication success drops to $22.2\%$ and $26.7\%$ respectively. It is interesting that the perceptual loss helps improve generalisability in this case.

\begin{figure}[tb]
    \centering
    \begingroup
    \setlength{\tabcolsep}{2pt} 
    \renewcommand{\arraystretch}{0} 
    \setlength\extrarowheight{0pt}
    \begin{tabular}{m{0.1\linewidth}m{0.83\linewidth}}
         &  \includegraphics[trim=0 200 0 980,clip,width=\linewidth]{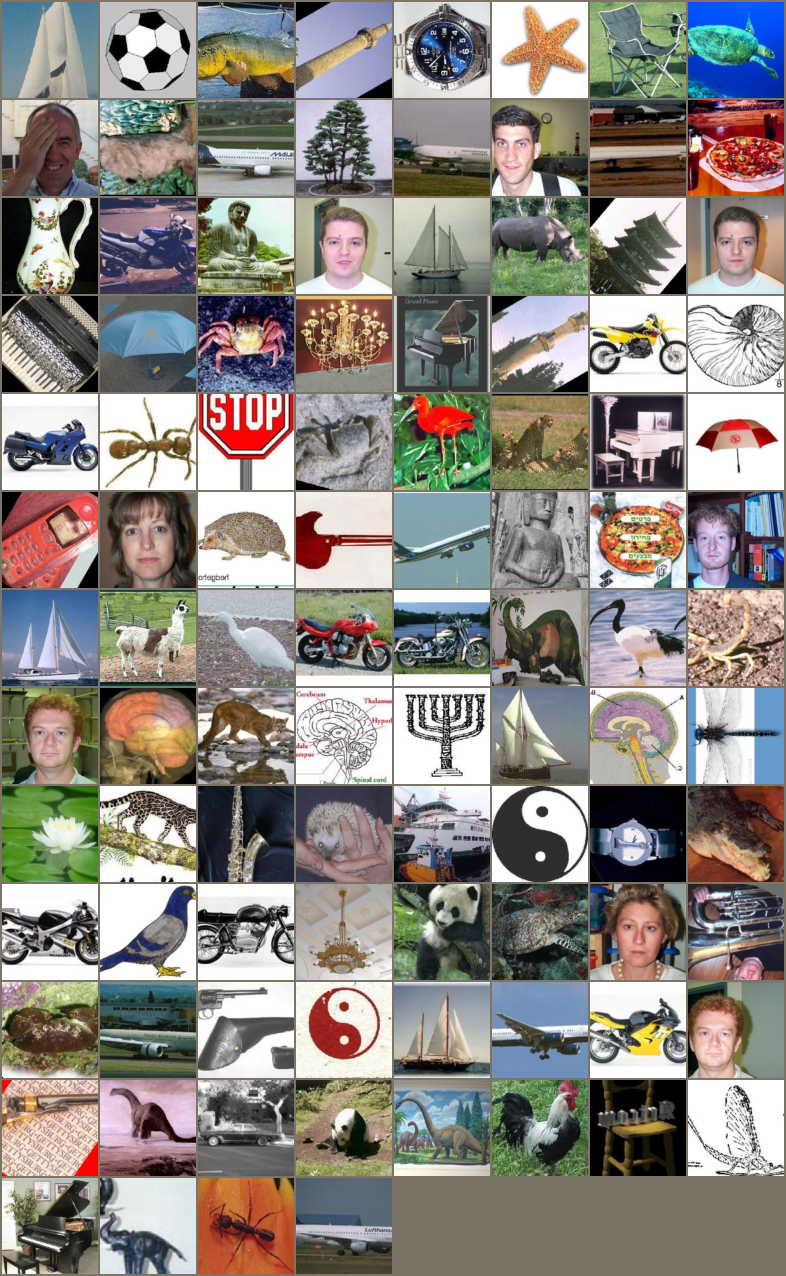}\\
         \begin{flushright} $\gameloss$ $22.2\%$ \end{flushright} &  \includegraphics[trim=0 200 0 980,clip,width=\linewidth]{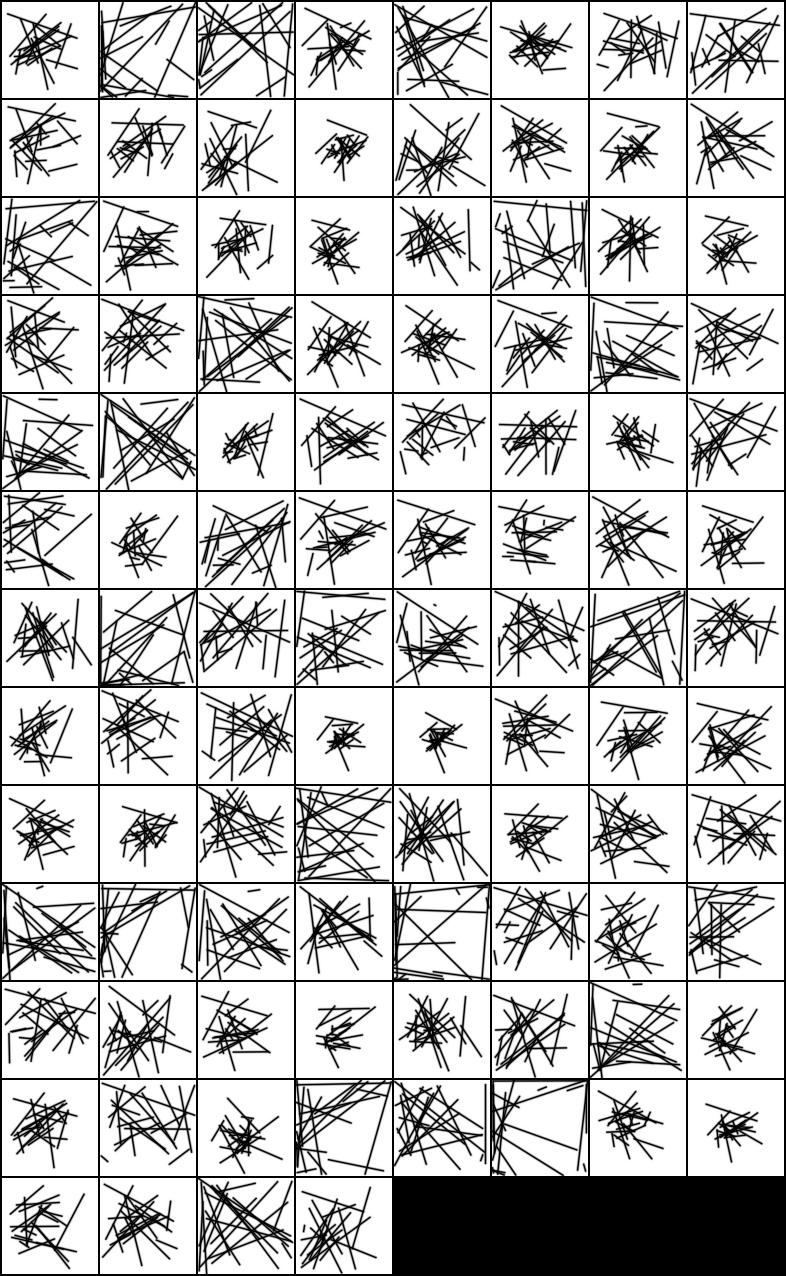}\\
         \begin{flushright} $+ \perceploss$ $26.7\%$ \end{flushright} & \includegraphics[trim=0 198 0 980,clip,width=\linewidth]{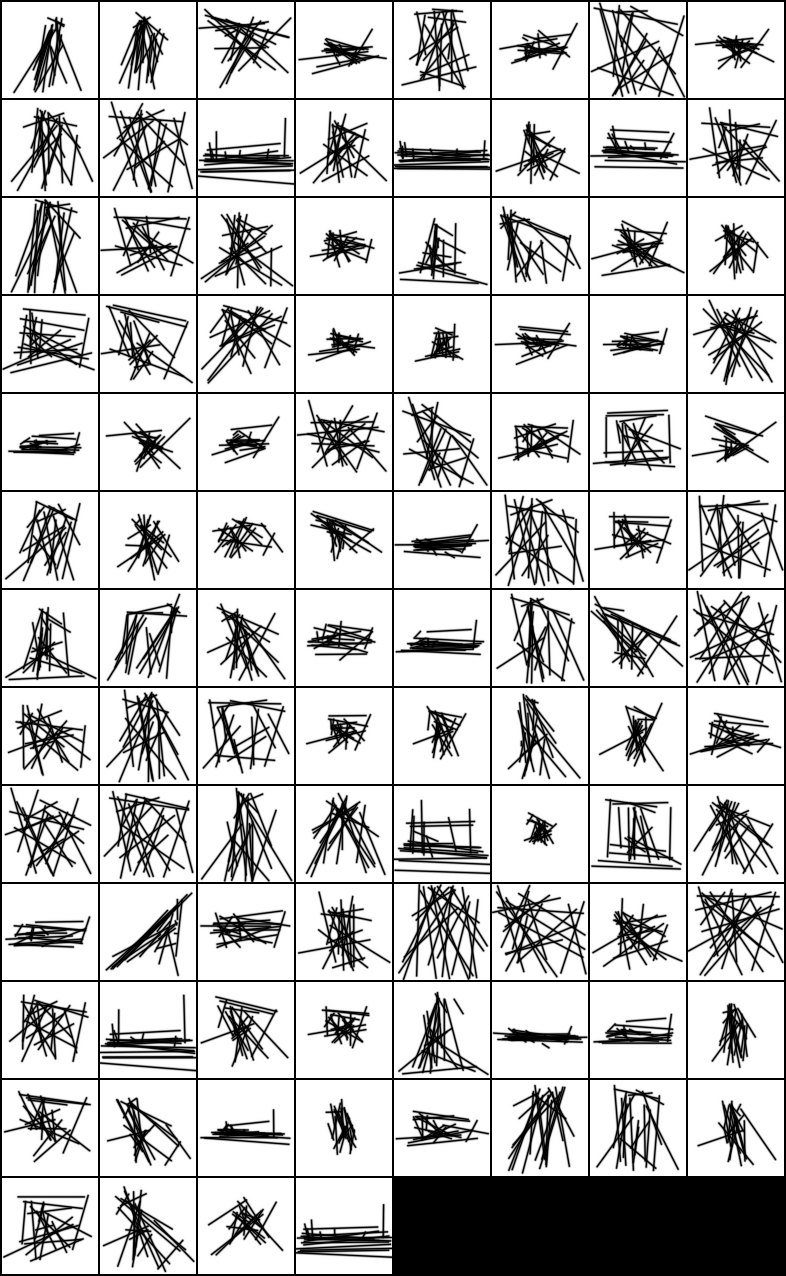}\\
    \end{tabular}
    \endgroup
    \caption{\textbf{Sketching agents, previously trained on STL-10 (original game) tested on Caltech-101 test set.} We compare models pretrained with $\gameloss$ only with those that also use $\perceploss$.}
    \label{fig:ood-images-stl10-caltech}
\end{figure}

\section{Sketching under different game setups} \label{app-games}

Table~\ref{tab:app-diff-games-sketches} illustrates more examples of sketches drawn under different game configurations, as discussed in \cref{subsec:oo-games} of the paper. Clearly, some classes are better represented and more interpretable to a human than others. Overall, the object-oriented game setups, especially \textit{OO-game different}, push the visual communication channel towards a more class-specific representation.

\begin{table}[t]
\caption{\textbf{More example sketches produced by the agents in the three different game setups using the $\gameloss + \perceploss$ loss.} Examples are from STL-10.}
    \label{tab:app-diff-games-sketches}
    \centering
    \begin{tabular}{cc}
        \toprule
        \multicolumn{2}{c}{Original game: $69.57\%\ (\pm 2.6)$}\\
        \includegraphics[width=0.48\textwidth]{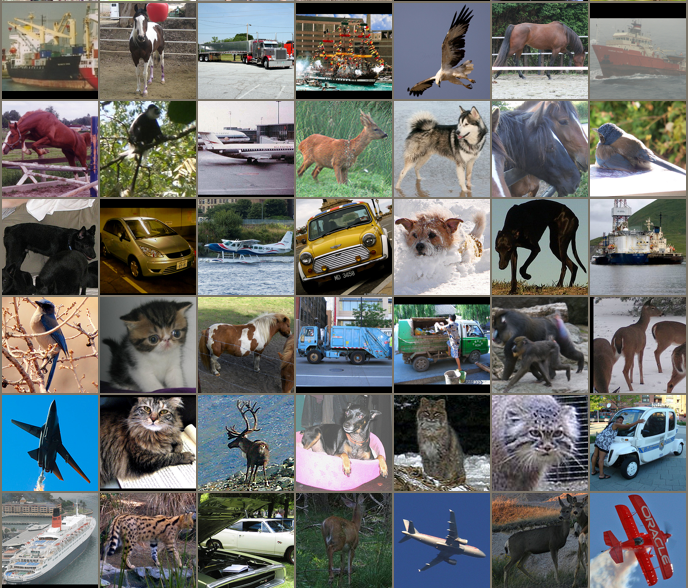} &
        \includegraphics[width=0.48\textwidth]{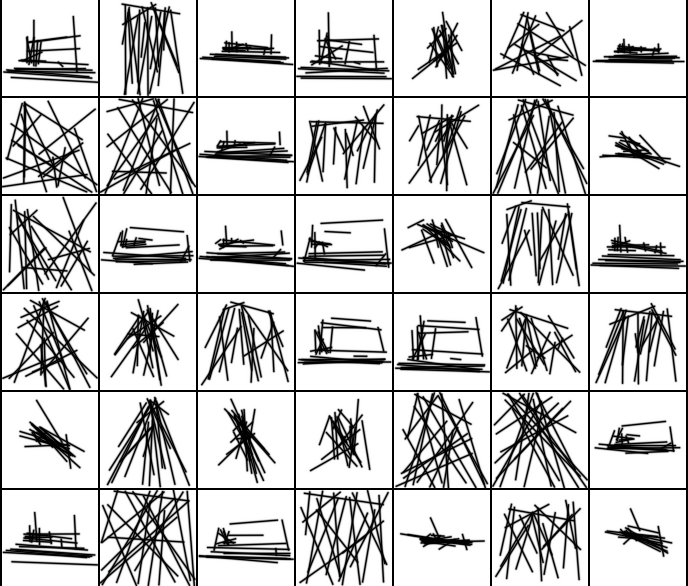}\\
        \midrule
        \multicolumn{2}{c}{OO-game same: $96.04\%\ (\pm 0.5)$}\\
        \includegraphics[width=0.48\textwidth]{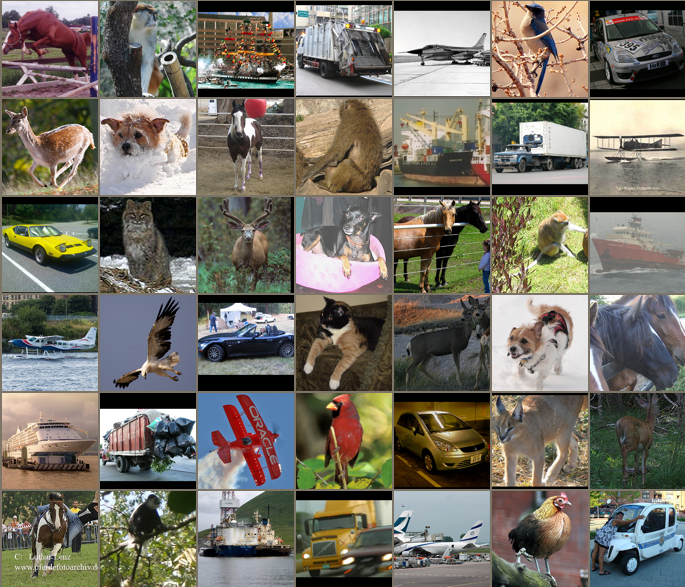} &
        \includegraphics[width=0.48\textwidth]{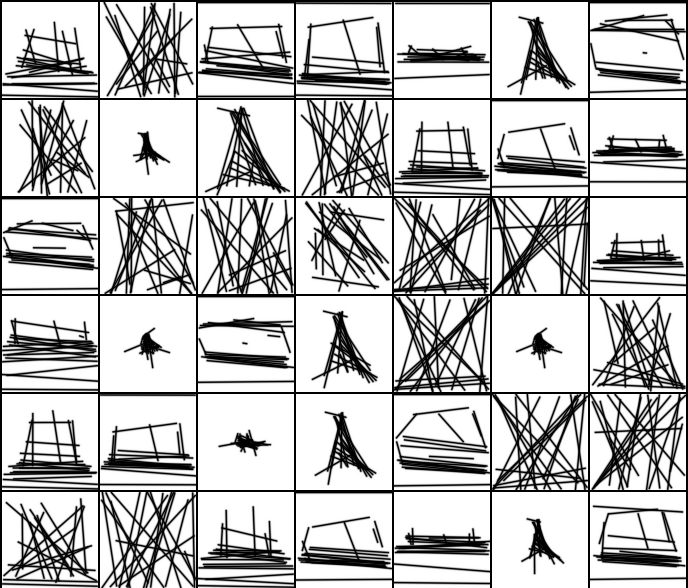}\\
        \midrule
        \multicolumn{2}{c}{OO-game different: $81.09\%\ (\pm 0.6)$}\\
        \includegraphics[width=0.48\textwidth]{images/STL-10-games/oo-samples6x7.png} &
        \includegraphics[width=0.48\textwidth]{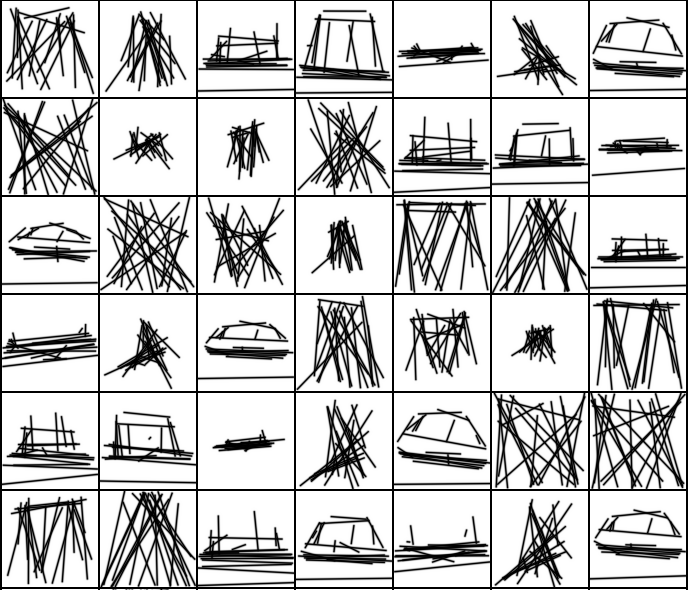}\\
        \bottomrule
    \end{tabular}
\end{table}

\begin{figure}[hp!]
    \centering
    \includegraphics[width=\textwidth]{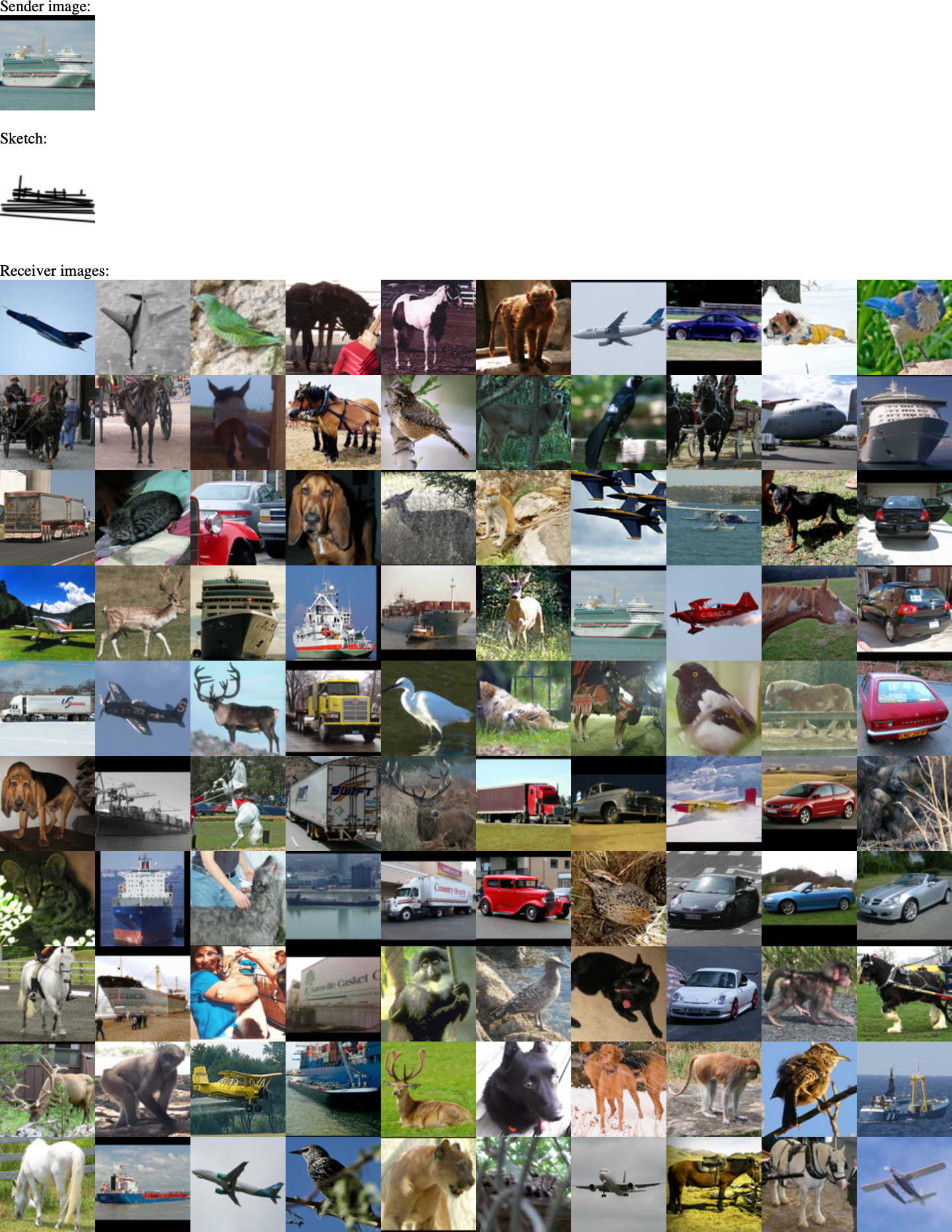}
    
  \caption{\textbf{Example of full reference game - \textit{original} setting with 99 distractors.}}\label{fig:ex-fullgame-orig}
\end{figure}

\begin{figure}[ht!]
    \centering
    \includegraphics[width=0.95\textwidth]{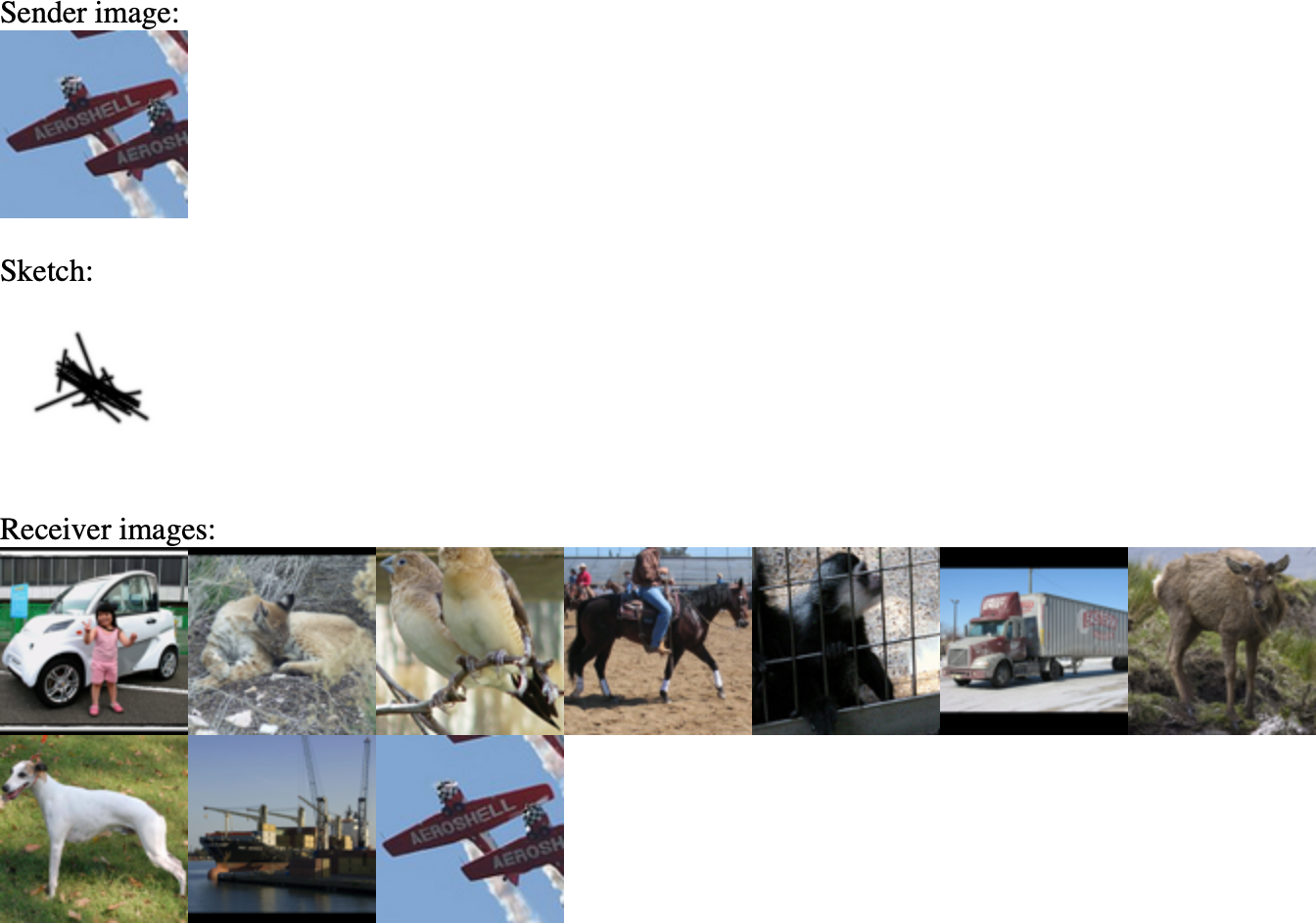}
    
  \caption{\textbf{Example of full reference game - \textit{object-oriented same} setting} in which the sender's target is part of the set of images shown to the receiver.}\label{fig:ex-fullgame-same}
\end{figure}

\begin{figure}[hb!]
    \captionsetup{type=figure}
    \centering
    \includegraphics[width=0.95\textwidth]{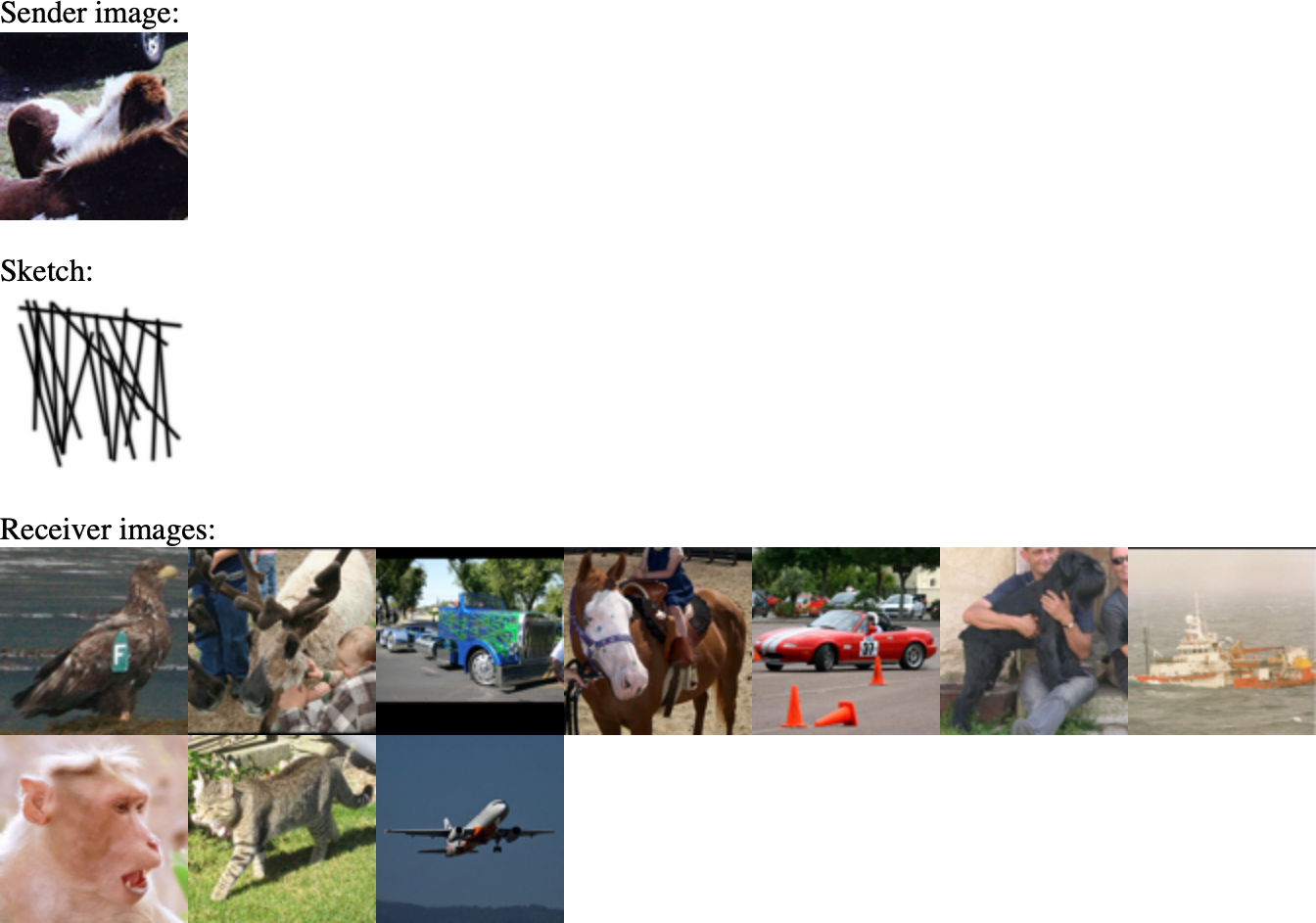}
    
  \caption{\textbf{Example of full reference game - \textit{object-oriented different} setting} in which the receiver's target is a different photograph that belongs to the same class as the sender's image.}\label{fig:ex-fullgame-diff}
\end{figure}

Further, in \cref{fig:ex-fullgame-orig}, we provide an example of the full reference games to help the reader understand how difficult the original game setting, with 99 distractors, would be to play for a human receiver. This should shed some light on how ``interpretable'' the communication is in the full context given to the receiver agent, which may contain many perceptually similar distractors in the original setting. The game in either of the object-oriented settings shown in \cref{fig:ex-fullgame-same,fig:ex-fullgame-diff}, played on STL10 classes, seems much more feasible to a human receiver.

\begin{landscape}
\section{Varying model capacity on Caltech-101}\label{app-modelcapacity}
In \cref{tab:app-modelcapacity} we provide additional examples to illustrate the effect of the model's capacity on the emergent sketches. More instance-specific details are captured by the wide model although the game is played in the \textit{OO-game different setting}.

\begin{table}[b!]
\caption{\textbf{The effect of the model's capacity on its sketches.} Examples from training on $128\times128$ pixel Caltech-101 images, in the \textit{OO-game different} setting.}
    \label{tab:app-modelcapacity}
    \centering
    \begin{tabular}{ccc}
        \toprule
         & Baseline & Wide \\
        \midrule
        & $50.46\%\ (\pm 1.5)$ & $64.99\%\ (\pm 1.5)$ \\
        \includegraphics[width=0.50\textwidth]{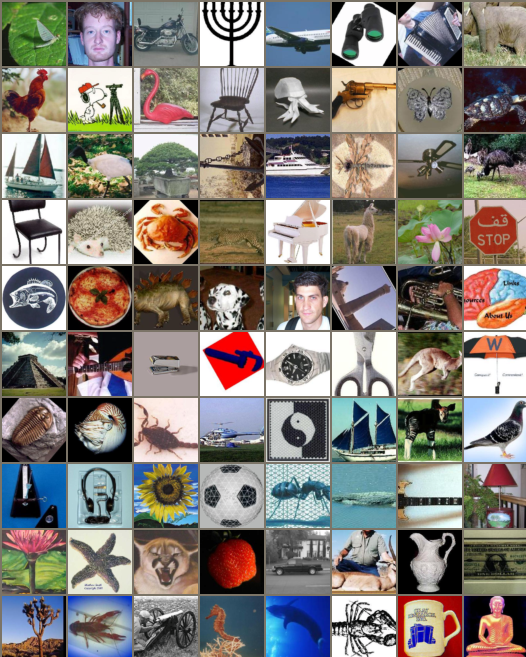} &
        \includegraphics[width=0.50\textwidth]{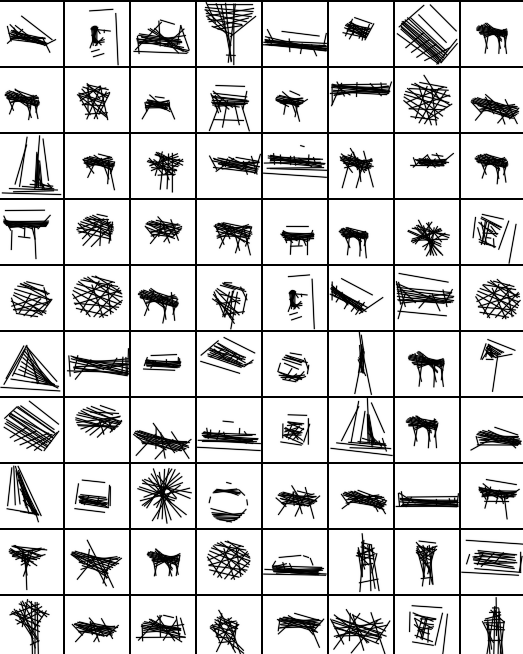} &
        \includegraphics[width=0.50\textwidth]{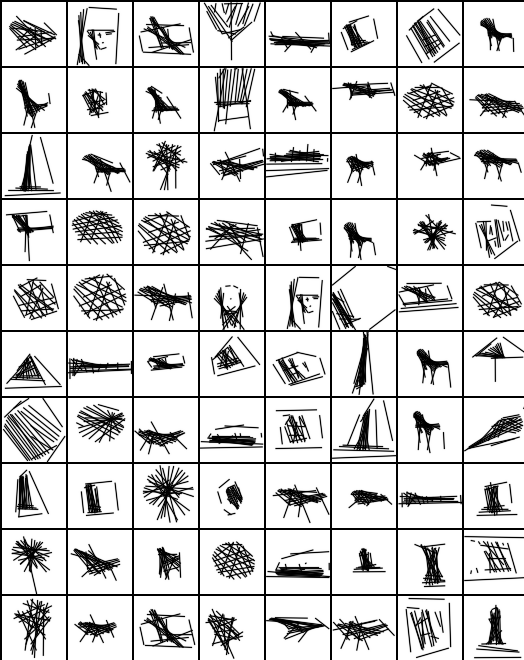} \\
        \bottomrule
    \end{tabular}

\end{table}
\end{landscape}

\begin{landscape}
\section{The effect of pretraining the visual system with texture/shape bias}
\label{app-shape-texture}
\cref{tab:app-shape-texture-bias} shows further examples of the influence the texture/shape bias has on the drawing quality.

\begin{table}[b!]
    \caption{\textbf{The effect of pretraining the VGG16 feature extractor network with a texture (ImageNet) or shape (Stylized-ImageNet) bias.} Examples are from agents trained in the \textit{original} game setup with $128\times128$ Caltech-101 images. Shape-biased sketches are, visually, more similar to the objects they represent.}
    \label{tab:app-shape-texture-bias}
    \centering
    \begin{tabular}{ccc}
        \toprule
         & ImageNet weights & Stylized-ImageNet weights\\
        \midrule
        & $78.46\%\ (\pm 2.0)$ & $77.09\%\ (\pm 1.9)$ \\
        \includegraphics[width=0.50\textwidth]{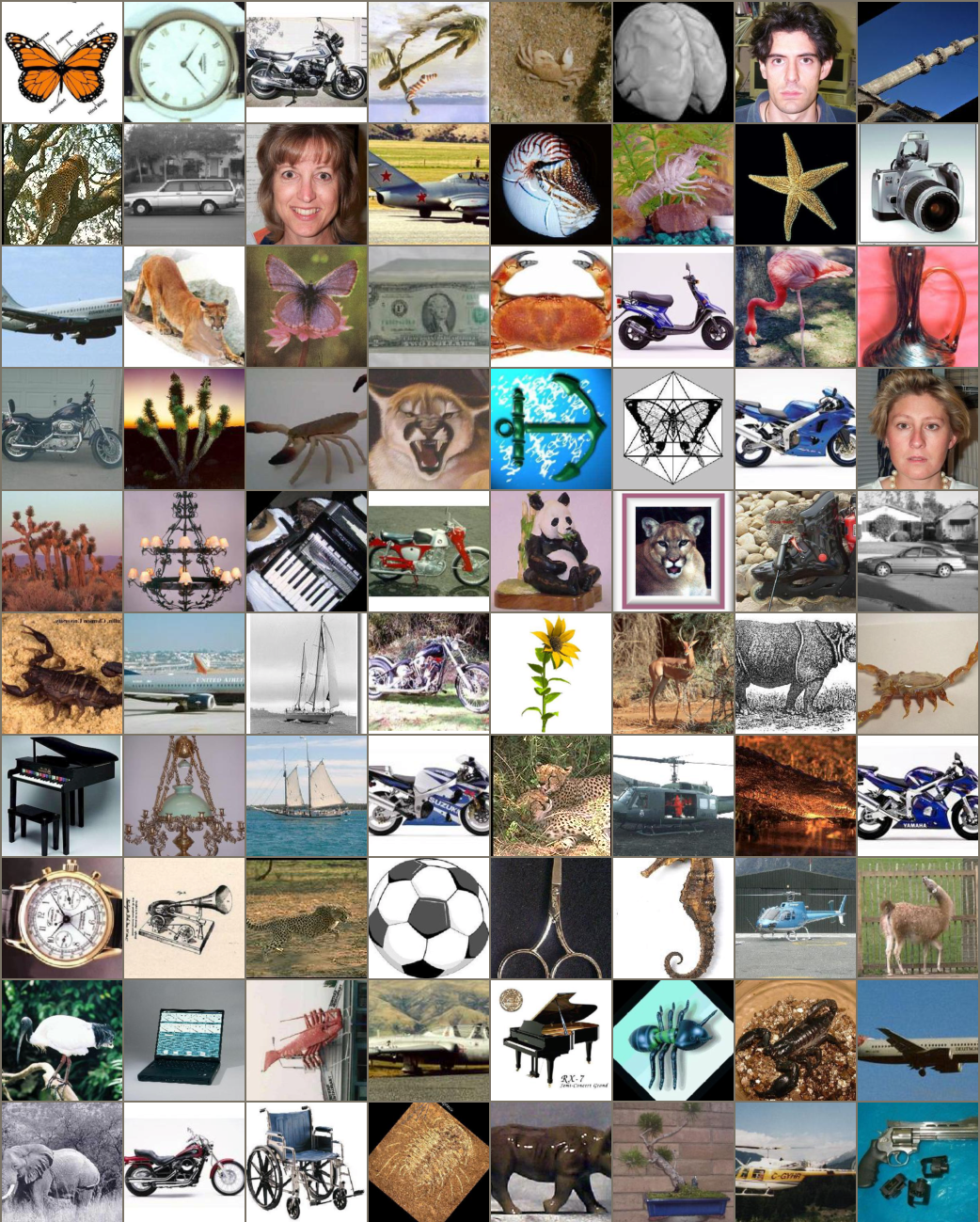} &
        \includegraphics[width=0.50\textwidth]{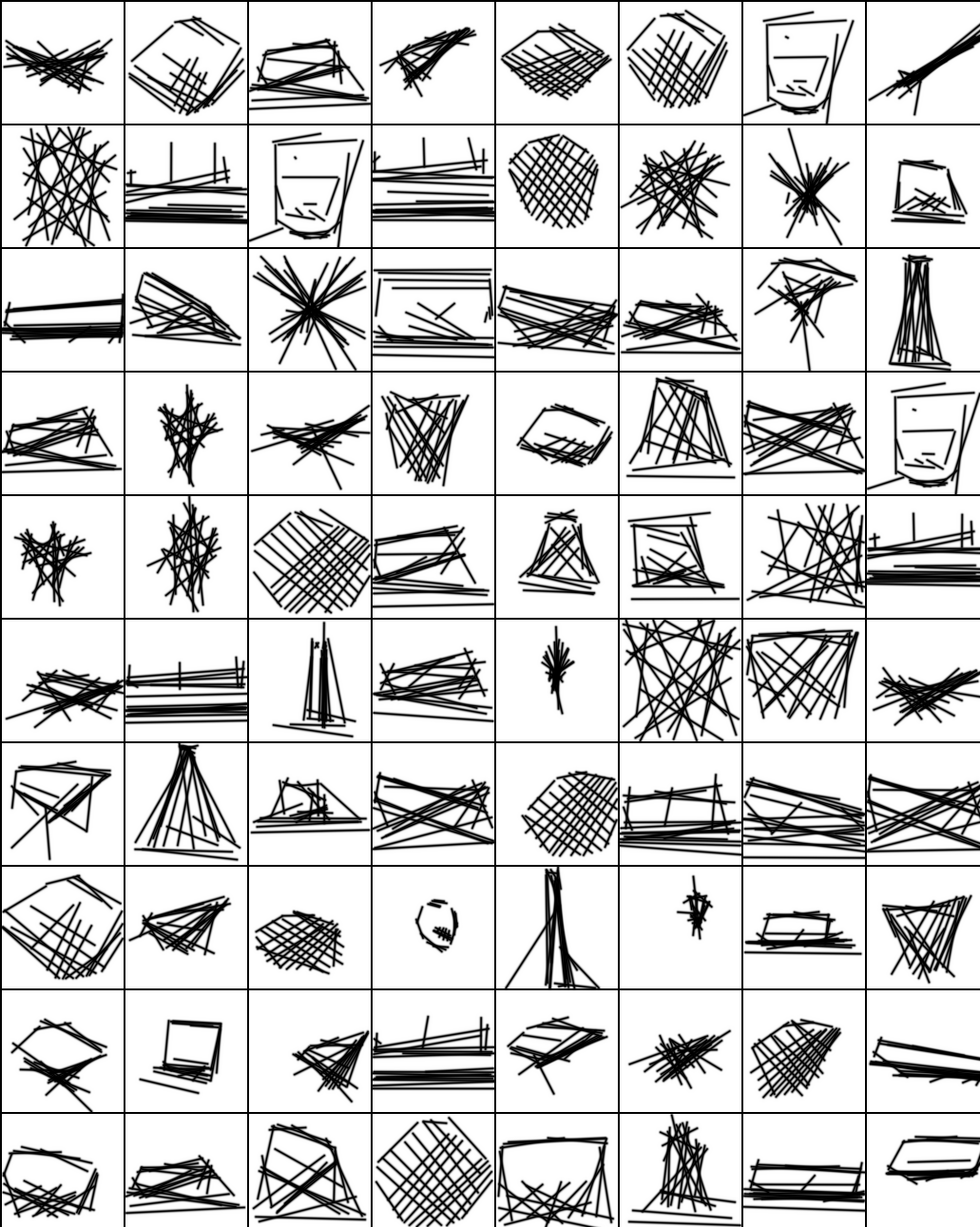} &
        \includegraphics[width=0.50\textwidth]{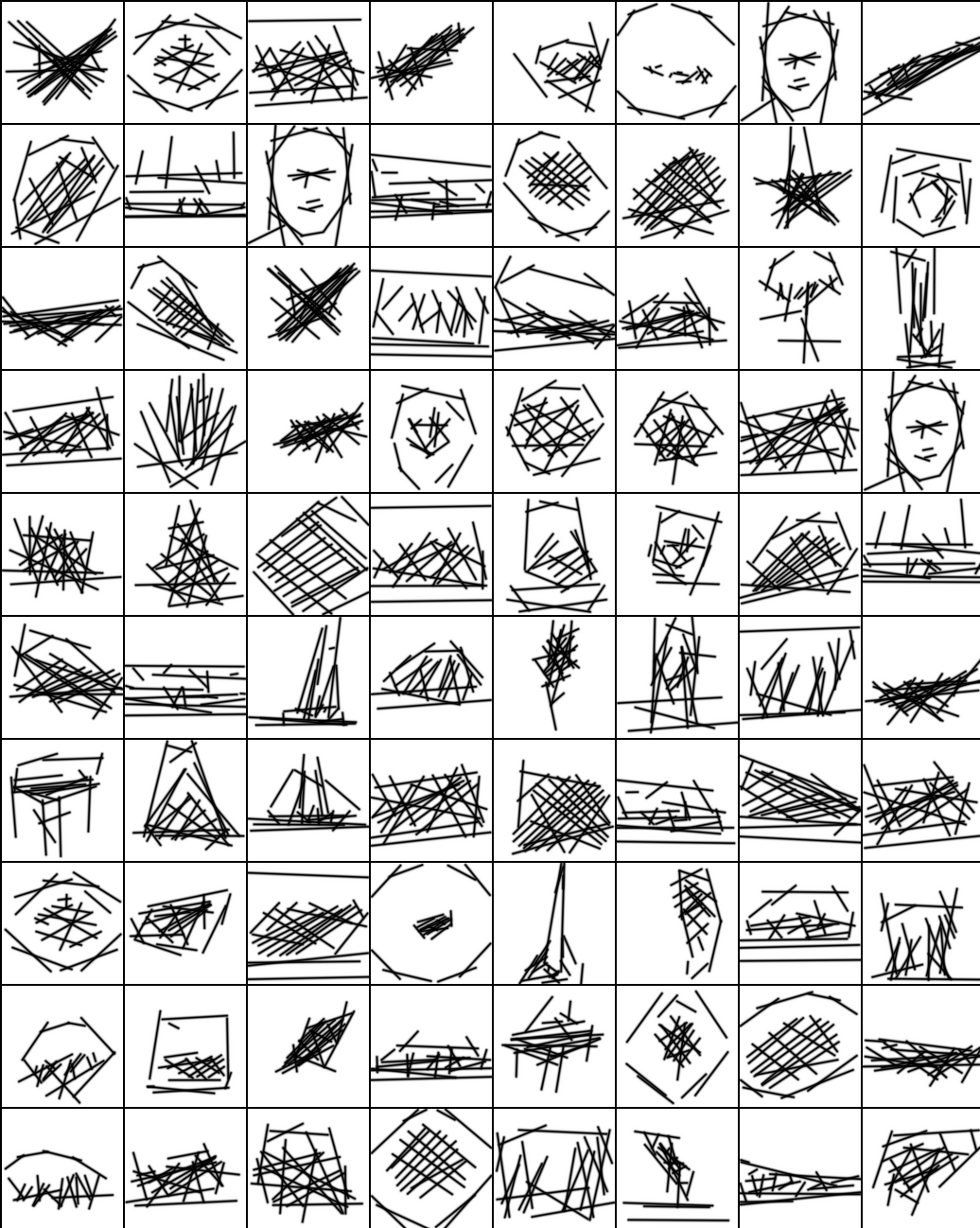} \\
        \bottomrule
    \end{tabular}
\end{table}
\end{landscape}


\section{Do the models learn to pick out salient features?}
\label{app-saliency}
\begin{figure}[b!]
    \centering
    \begingroup
    \setlength{\tabcolsep}{2pt} 
    \renewcommand{\arraystretch}{0} 
    \setlength\extrarowheight{0pt}
    \begin{tabular}{m{0.1\linewidth}m{0.83\linewidth}}
         &  \includegraphics[width=\linewidth]{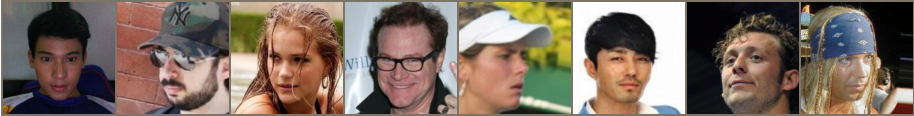}\\
         \begin{flushright} ImageNet weights \end{flushright} &  \includegraphics[width=\linewidth]{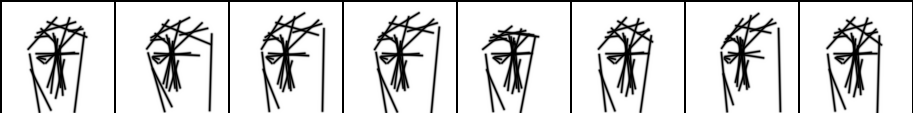}\\
         \begin{flushright} Stylized weights \end{flushright} &  \includegraphics[width=\linewidth]{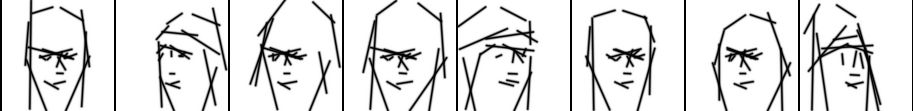}\\
         &  \includegraphics[width=\linewidth]{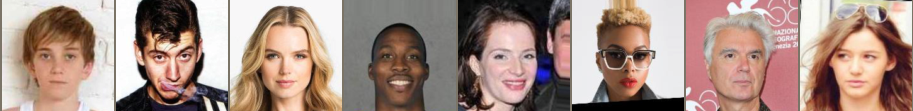}\\
         \begin{flushright} ImageNet weights \end{flushright} &  \includegraphics[width=\linewidth]{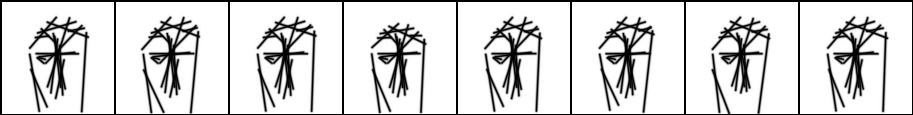}\\
         \begin{flushright} Stylized weights \end{flushright} &  \includegraphics[width=\linewidth]{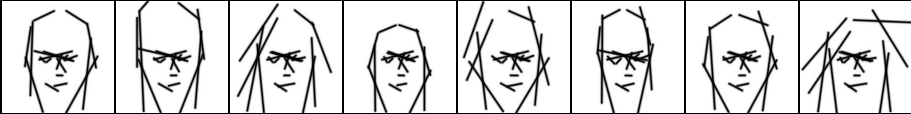}\\
         &  \includegraphics[width=\linewidth]{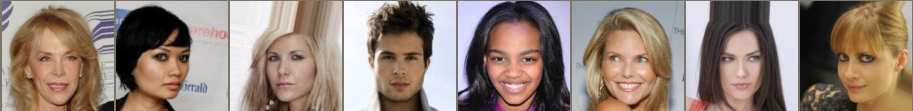}\\
         \begin{flushright} ImageNet weights \end{flushright} &  \includegraphics[width=\linewidth]{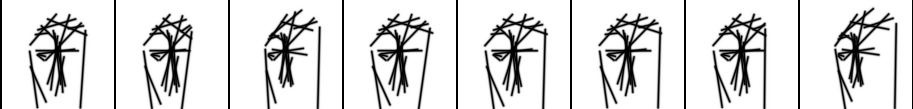}\\
         \begin{flushright} Stylized weights \end{flushright} &  \includegraphics[width=\linewidth]{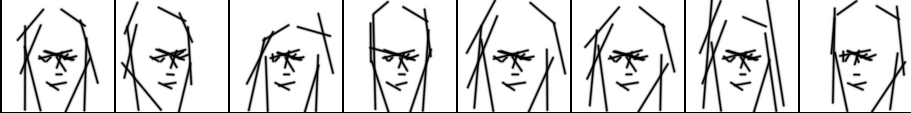}\\
         &  \includegraphics[width=\linewidth]{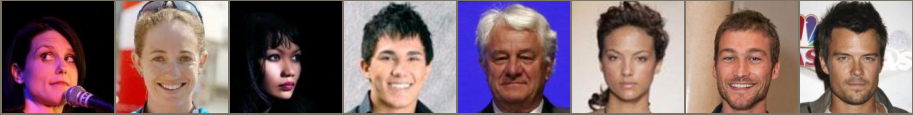}\\
         \begin{flushright} ImageNet weights \end{flushright} &  \includegraphics[width=\linewidth]{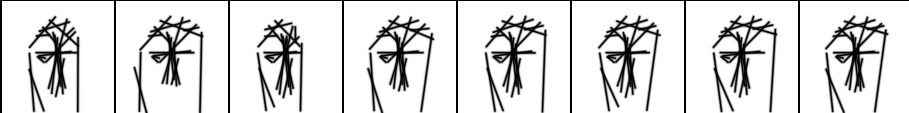}\\
         \begin{flushright} Stylized weights \end{flushright} &  \includegraphics[width=\linewidth]{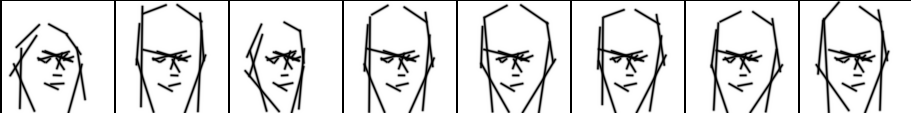}\\
    \end{tabular}
    \endgroup
    \caption{Sketches from \textit{original} variant game using the CelebA dataset, the perceptual loss and different biases from backbone weights. Although both models have near perfect communication success, it is clear the inducing a shape bias helps bring out the most salient and distinctive features.}
    \label{fig:app-celeba}
\end{figure}

\cref{fig:app-celeba} provides additional results of the experiment discussed in \cref{subsec:salient-features}, which looks at the ability of the model, either texture or shape-biased, to capture salient features. It is clear that the shape-biased sketches are visually more correlated with the photos.

\begin{landscape}
\section{How much do sketches differ visually across seeds?}\label{app-sec:overlaid-sketches}
Throughout the paper, the sample sketches are presented from one seed out of the 10 model runs. Here we include an example of an overlay of the 10 seeds, normalised to look like a heatmap so that darker lines represent strokes generated by \textit{more} models. As can be seen in \cref{fig:overlaid-sketches}, the 10 different models trained on Caltech 101 from different seeds are consistent in picking out key features of the input image, but have variation in finer details.

\begin{figure}[b!]
    \centering
    \begingroup
    \setlength{\tabcolsep}{2pt} 
    \renewcommand{\arraystretch}{0} 
    \setlength\extrarowheight{0pt}
    \begin{tabular}{m{0.4\linewidth}m{0.4\linewidth}}
        \includegraphics[width=\linewidth]{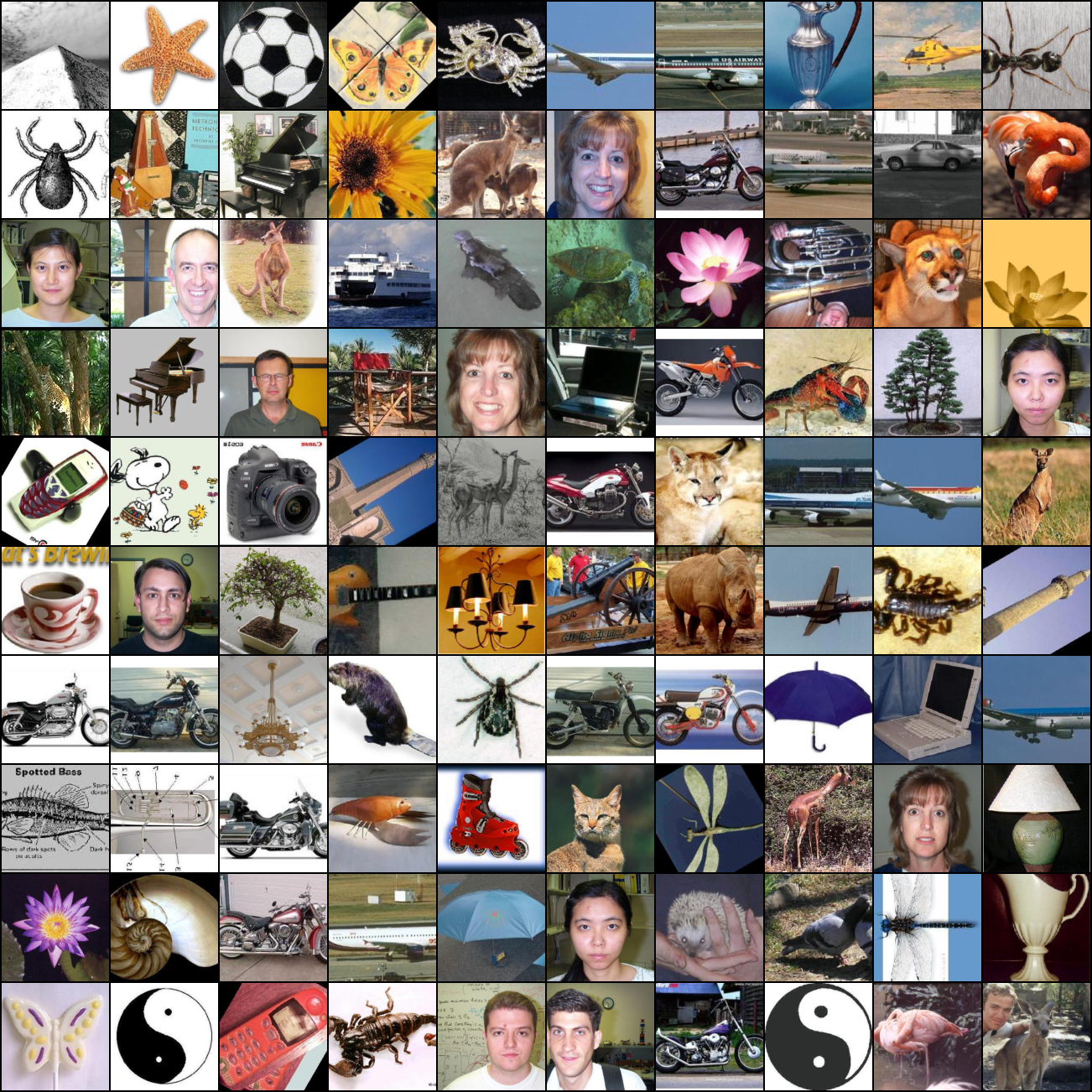} & \includegraphics[width=\linewidth]{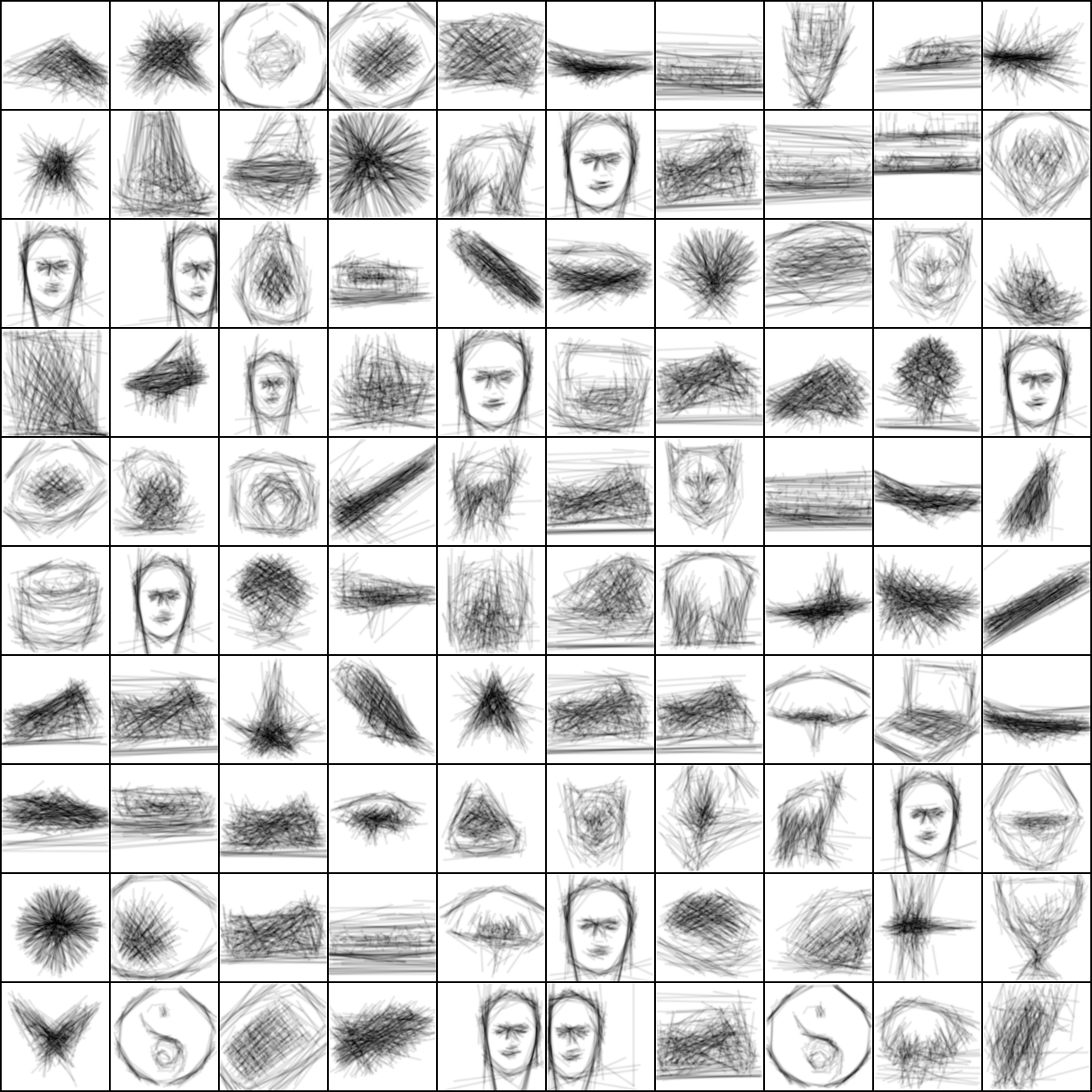}\\
    \end{tabular}
    \endgroup
    \caption{\textbf{An overlay of 10-seeds sketches} drawn by a model trained in the \textit{original} game variant on Caltech101, with Stylized-ImageNet weights.}
    \label{fig:overlaid-sketches}
\end{figure}
\end{landscape}

\section{Human Evaluation Experiment Details} \label{app-sec:humaneval-details}
This section details the setup and results of our human evaluation experiment. As mentioned in \cref{sec:humaneval}, we perform a pilot study that looks at gameplay success when the receiver agent role is played by a human participant. 

\subsection{Experimental Setup}

\begin{figure}[b!]
    \centering
    \includegraphics[width=0.60\textwidth]{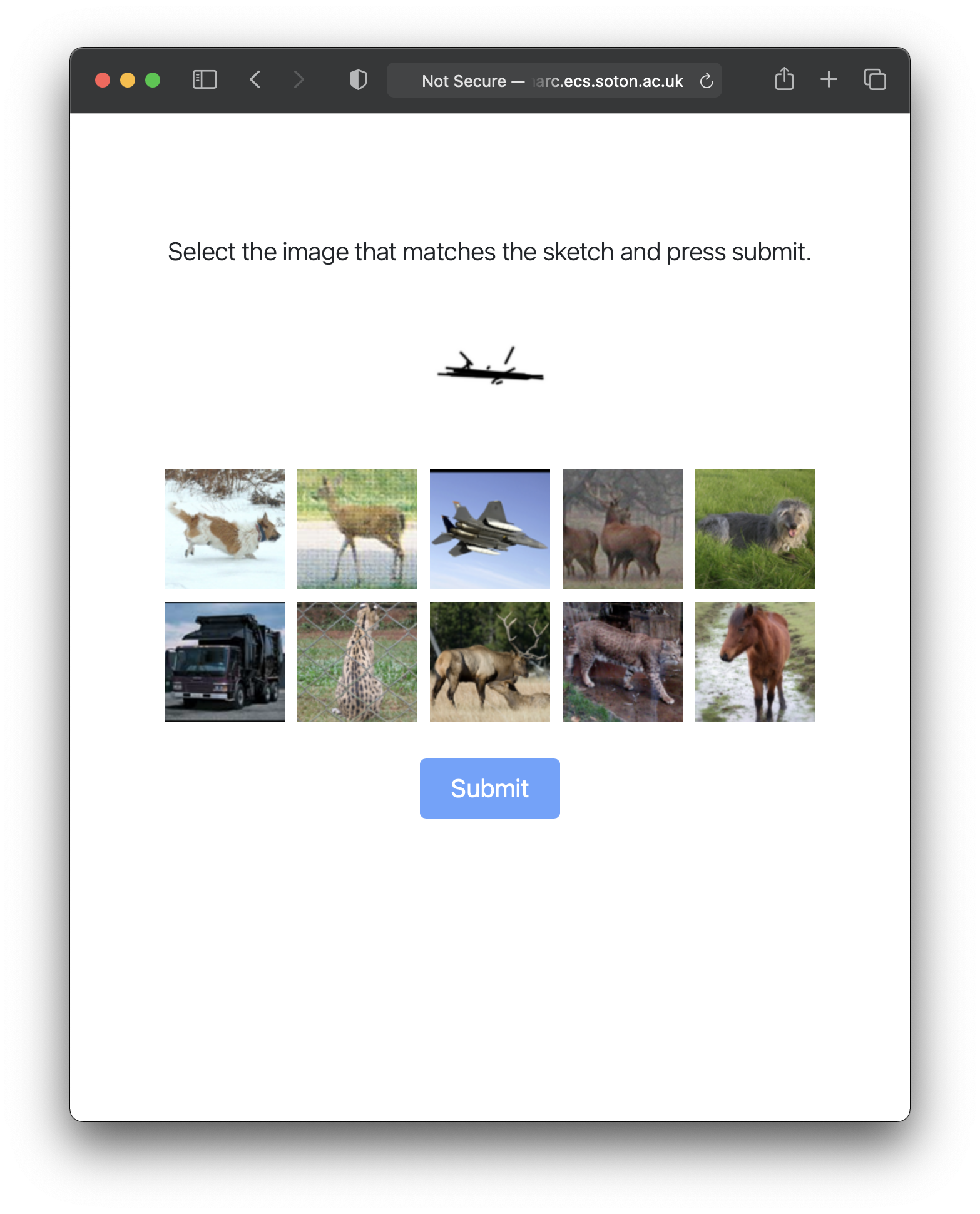}
  \caption{\textbf{Example of a game, original setting - the human participant has to pick from 10 images.}}\label{fig:uiexample}
\end{figure}

\paragraph{The task} 
To reiterate the experimental task, the human participant is shown a sketch (previously generated by a trained Sender agent during model evaluation) and is asked to select by clicking the corresponding target image from a grid of images, as illustrated in \cref{fig:uiexample}.

\paragraph{The data} 
The sketches used in this experiment are generated in five different game configurations, varying game setup, agents' training objective and the number of strokes. For the purpose of this study, the sketches are generated by models trained with the same fixed random seed. Whilst there is inevitably some variation in models from different seeds (see \cref{app-sec:overlaid-sketches}), this is not explored in the human evaluation.

For each game setting, the participant played 30 games, matching a total of 30 sketches to different target image sets. Each human participant played a total of 150 games, and the total amount of data collected in the pilot study corresponds to 1800 games. For the purpose of this study, the games were chosen randomly from all those possible within the STL-10 test dataset. For all game settings used in the human pilot study, we limit the number of distractors to $K=9$.

\paragraph{User interface}
To allow human participants to play the game, a web interface was developed and each participant was provided with a set of 5 unique URLs corresponding to the 5 different game settings. Information on what the different settings involved was not provided to the participants. Each URL took the participant through 30 games and stored their answers in a database. 

An example of such a game is shown in \cref{fig:uiexample}. We do not impose a time limit per game, but record how much time the participants take to make their guess. \Cref{fig:admin-ui} shows our admin interface which summarises the averaged statistics based on the games played in this pilot study. \Cref{fig:uiexamples-withfeedback} shows the interface when feedback is given (see \cref{sec:humaneval-withfeeback}).

\begin{center}
    \captionsetup{type=figure}
    \centering
    \includegraphics[width=0.80\textwidth]{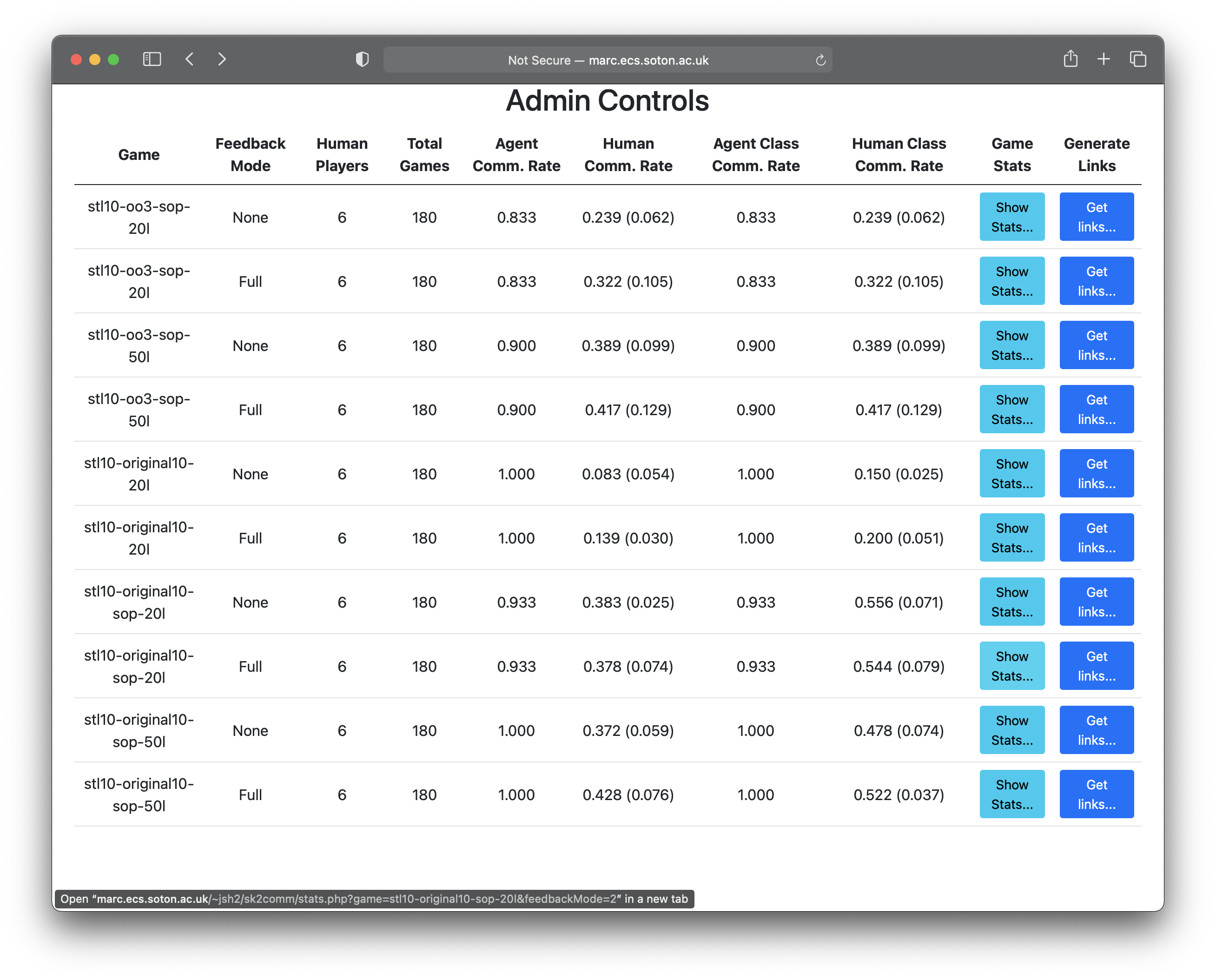}
  \caption{\textbf{The admin interface.}}\label{fig:admin-ui}
\end{center}

\begin{center}
    \captionsetup{type=figure}
    \centering
    \includegraphics[width=0.60\textwidth]{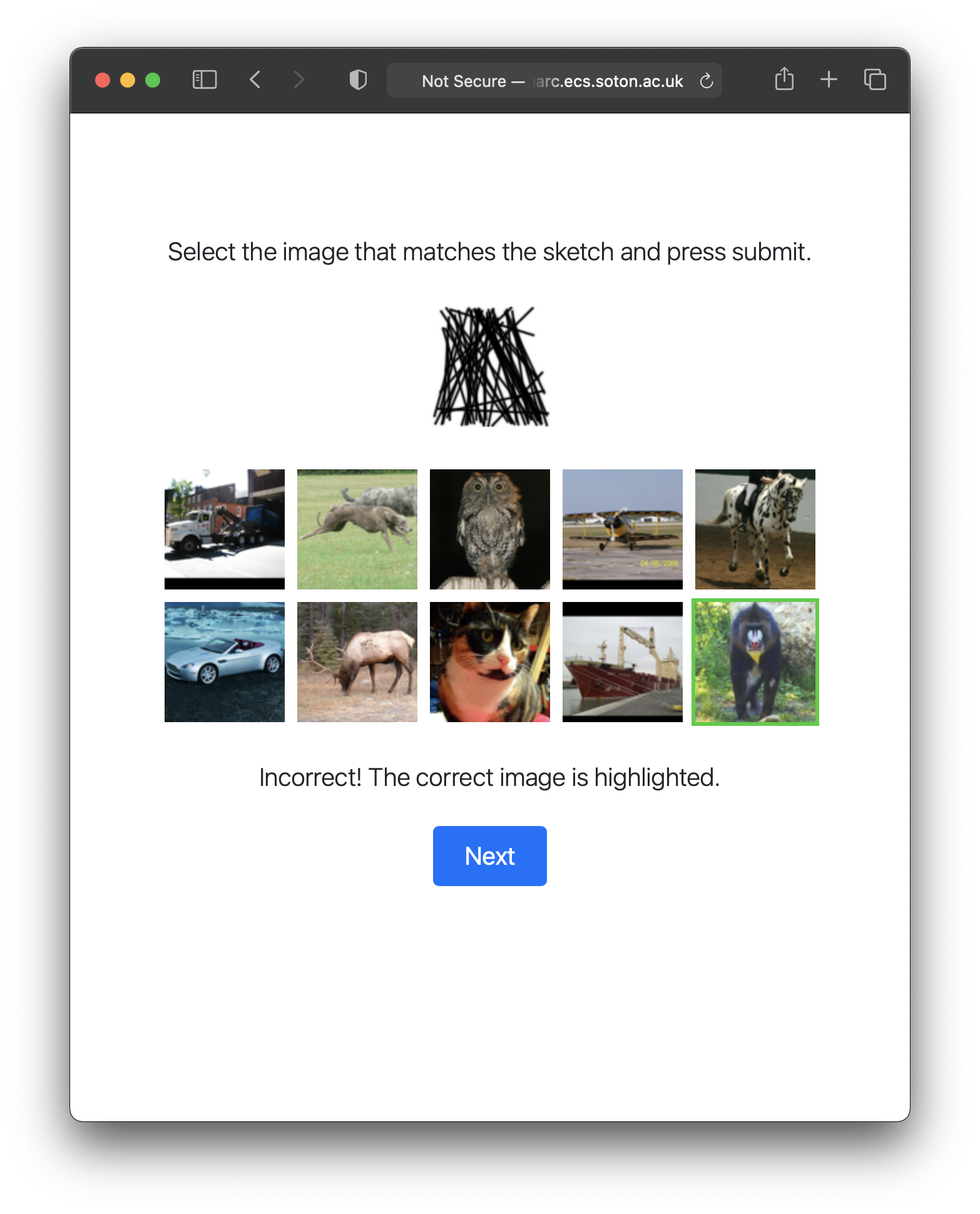}
  \caption{\textbf{Example of the game played with feedback.}}\label{fig:uiexamples-withfeedback}
\end{center}

\paragraph{Participants}
We divide the human evaluation into two disjoint study groups: participants who just play the game with no feedback and, hence, cannot learn during game-play (results are presented in \cref{tab:humaneval-nofeedback}), and a second group which is allowed to learn from feedback. Details about the latter group are discussed in \cref{sec:humaneval-withfeeback}. 

For the purpose of the study, we collect results from 6 participants per group. Overall, the study includes participants aged between 20 to 35 with various professions. Participation in the study does not require any specific skills.

\subsection{Can human participants \textit{learn} to play the game?} \label{sec:humaneval-withfeeback}
The principal pilot study (\cref{sec:humaneval}) is looking at humans' ability to play the game with an agent, with no feedback involved. The human participants will not know what the correct target was or if they guessed correctly. We also pose a slightly different question: Can humans learn to play the game with an agent? For this secondary study, after participants select what they believe to be the target image, they will be told if their selection was correct or not and the correct target will be indicated (as shown in \cref{fig:uiexamples-withfeedback}).

\Cref{tab:humaneval-feedback} summarises the statistics computed over the participants in this secondary study. The participants were tested on the same set of games as the first group, and the same metrics are reported.

\begin{table}[b!]
    \caption{\textbf{Human Evaluation results, learning allowed from feedback.}}
    \label{tab:humaneval-feedback}
    \centering
    \begin{tabular}{llllll}
    \toprule
    &  &  & Agent & Human & Human \\ 
        Game & Loss & Lines & comm. rate & comm. rate & class comm. rate \\ 
        \midrule
        original & $l=l_{game}$ & 20 & $100\%$ & $13.9\%\ (\pm3.0)$ & $20.0\%\ (\pm5.1)$ \\ 
        original & $l=l_{game} + l_{perceptual}$ & 20 & $93.3\%$ & $37.8\%\ (\pm7.4)$ & $54.4\%\ (\pm7.9)$ \\ 
        original & $l=l_{game} + l_{perceptual}$ & 50 & $100\%$ & $42.8\%\ (\pm7.6)$ & $52.2\%\ (\pm3.7)$ \\ 
        oo diff & $l=l_{game} + l_{perceptual}$ & 20 & $83.3\%$ & $32.2\%\ (\pm10.5)$ & $32.2\%\ (\pm10.5)$ \\ 
        oo diff & $l=l_{game} + l_{perceptual}$ & 50 & $90.0\%$ & $41.7\%\ (\pm12.9)$ & $41.7\%\ (\pm12.9)$ \\ \bottomrule
    \end{tabular}
\end{table}

T-tests run between the averaged communication success rates of the same game setting in the group with feedback versus the one without feedback, do not show a statistically significant improvement when participants are allowed to learn from feedback, except for the original game with $l=l_{game}$ only. As we expected, participants in both study groups had the lowest scores in this game across all tested settings: without feedback the averaged $commrate=8.3\% (\pm5.4)$; with feedback, $commrate=13.9\% (\pm3.0)$. The sketches drawn by a Sender agent pretrained without the perceptual loss are not ``constrained'' to resemble the target image, hence they are the least interpretable. However, the two-tailed P-value between the two groups performance in this setting was less than 0.0001 which suggests that feedback can lead to a statistically significant improvement when the sketches are not visually interpretable. Still, this is by far the worst communication scenario. This is also indicated by the amount of time the participants spent on average on this game which is higher than in other settings, the majority taking between 1 minute and 2 minutes 30 seconds per sketch.

In the future, it would perhaps be interesting to explore if humans could learn with feedback if they were to play more games; the 30 games per setting used in this experiment is possibly too little to allow a human player to robustly learn the strategy used by the agent. 

\subsection{Extended discussion of results}
\paragraph{Does the addition of the perceptual loss give statistically significant improvement over games which use only the hinge loss?}
All participants were asked to play the original game with 20 stroke-sketches produced when $l=l_{game}$ and also when $l=l_{game} + l_{perceptual}$. Performing t-tests between the averaged communication rates within each study group, with and without perceptual loss, resulted in P values less than 0.0001, which indicates that the perceptual loss leads to a statistically significant improvement in humans’ ability to play the game with the agent.

\paragraph{Does the number of strokes influence human performance?}
We tested the original game and the object-oriented game setup, each with 20 and 50 strokes. The results indicate that in both settings, a higher number of strokes leads to better communication. However, in the group without feedback (\cref{tab:humaneval-nofeedback}), the mean communication rate was similar for the original setting with 20 strokes and with 50 strokes. The same game setting tested by people with feedback, however, showed a small increase in overall communication success. One should take into account that in this game setting, the human player might have to choose between more images from the same class. For an artificial agent, this can be an easy task. However, we might envisage a scenario in which other characteristics of a drawing would be included, such as colour, which might help the human differentiate between multiple instances from the same class. For example, think of 3 different species of birds, which could all be represented by some very general sketch, but could become distinctive if the colour were to be included. In the object-oriented game setting, the gap between 20-stroke and 50-stroke games is a bit more significant for both study groups.

\paragraph{Are humans better at determining the broader class of a sketch than at recognising the specific instance?} 
In the original game setting, it is possible to encounter distractor images from the same class as the target. In addition to the communication rate measure, which shows the overall success of an agent (human in this case) selecting the correct target image, we also compute the class communication rate, which calculates the overall success of an agent selecting an image from the same class as the true target. T-tests run between human communication rate and human class communication rate in the original game settings showed a statistically significant difference in both study groups. Humans are significantly better at understanding the broad class than they are at determining a specific instance based on the sketch in the games where there are multiple targets of the same class. This effect is possibly weakened by an increase in the number of strokes, however, as evidenced by a consistent lowering of statistical significance.




\end{document}